\documentclass{article}

\usepackage{microtype}
\usepackage{graphicx}
\usepackage{subcaption}
\usepackage{booktabs} 
\usepackage{wrapfig}

\usepackage{hyperref}

\newcommand{\nd}[1]{}
\newcommand{\csb}[1]{}
\newcommand{\bh}[1]{}
\newcommand{\out}[1]{}

\usepackage[preprint]{icml2026}

\usepackage{amsmath}
\usepackage{amssymb}
\usepackage{mathtools}
\usepackage{amsthm}
\usepackage{listings}

\usepackage[capitalize,noabbrev]{cleveref}

\theoremstyle{plain}

\theoremstyle{definition}

\theoremstyle{remark}

\usepackage[textsize=tiny]{todonotes}

\begin{document}

\twocolumn[
  \icmltitle{Small Models, Smarter Learning: The Power of Joint Task Training}



  \icmlsetsymbol{equal}{*}

  \begin{icmlauthorlist}
    \icmlauthor{Csaba Both}{yyy,equal}
    \icmlauthor{Benjamin Hoover}{xxx,comp}
    \icmlauthor{Hendrik Strobelt}{comp}
    \icmlauthor{Dmitry Krotov}{comp}
    \icmlauthor{Daniel Karl I. Weidele}{comp}
    \icmlauthor{Mauro Martino}{comp}
    \icmlauthor{Nima Dehmamy}{comp,equal}
  \end{icmlauthorlist}

  \icmlaffiliation{yyy}{Northeastern University}
  \icmlaffiliation{comp}{IBM Research}
  \icmlaffiliation{xxx}{Georgia Tech}

  \icmlcorrespondingauthor{Csaba Both}{both.c@northeastern.edu}
  \icmlcorrespondingauthor{Nima Dehmamy}{Nima.Dehmamy@ibm.com}

  \icmlkeywords{Machine Learning, ICML}

  \vskip 0.3in
]



\printAffiliationsAndNotice{\icmlEqualContribution}  

\begin{abstract}

    Multi-task learning improves generalization, but when does it reduce the model capacity required to learn?
    We provide a systematic study of how joint training affects the \emph{learning transition}---the minimum model size at which a task can be learned---using nested arithmetic (ListOps) and permutation groups as controlled testbeds.
    Certain task pairings dramatically reduce model size requirements: combining easy operations (MAX, MIN, PROD) with hard ones (modular addition, permutation products) enables learning with 2--7$\times$ fewer parameters.
    Crucially, we also identify when synergies \emph{fail}: pairing structurally similar hard tasks (e.g., ADD with alternating-sign NADD) provides no benefit, nor does pairing tasks lacking shared computational primitives.
    PCA of learned embeddings reveals that successful joint training induces structured number representations (ordering, parity, modular structure) absent in single-task models.
    Transfer experiments confirm these representations are causal: models pretrained on easy tasks learn addition at 7$\times$ smaller sizes.
    Our results establish that task compatibility---not mere diversity---determines whether joint training reduces capacity requirements, providing quantitative guidance for curriculum design. 
\end{abstract}

\section{Introduction}
Scaling laws predict language model performance as a function of model size \cite{hoffmann2022training}, compute \cite{muennighoff2023scaling}, and dataset size \cite{hestness2017deep}, but treat training data as homogeneous and do not explain how task composition shapes learning efficiency.
Prior work has shown that multi-task learning improves generalization \cite{caruana1997multitask, ruder2017overview} and that joint arithmetic training can benefit individual tasks \cite{lee2024teaching, abedsoltan2025task}.
However, a systematic understanding of \emph{when} joint training helps versus fails---and by how much---remains lacking.

\begin{figure*}[t!]
    \centering
    \includegraphics[width=\textwidth]{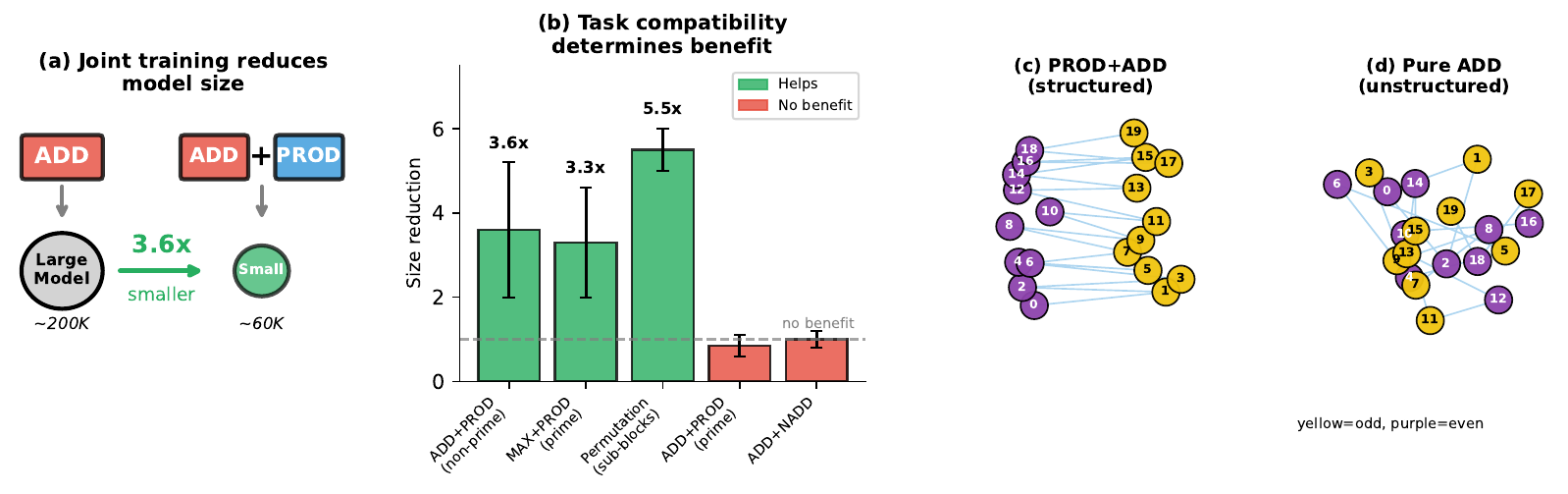}
    \caption{\textbf{Summary of main findings.} (a) Joint training on ADD+PROD reduces minimum model size by 3.6$\times$ compared to ADD alone. (b) Task compatibility determines benefit: easy+hard pairings help (green), while hard+hard pairings provide no benefit (red). (c,d) PCA of learned embeddings: joint training (PROD+ADD) induces structured representations with clear parity separation, while single-task training (pure ADD) yields unstructured embeddings.}
    \label{fig:summary}
\end{figure*}

We address this gap by providing the first quantitative characterization of how task composition affects the \emph{learning transition}: the minimum model size at which a task can be learned.
Using nested arithmetic operations (ListOps) and permutation groups as controlled testbeds, we systematically vary task pairings and measure changes in transition points.
Our key finding is that task compatibility, not mere task diversity, determines whether joint training reduces capacity requirements.
Specifically, we identify task pairings that reduce model size requirements by 2--7$\times$, as well as pairings that provide no benefit despite superficial similarity (Fig.~\ref{fig:summary}).

Mathematical tasks provide an ideal testbed for this investigation: they allow precise control over task difficulty, quantitative evaluation of accuracy, and systematic combination of operations \cite{nangia2018listops, nye2021show, dziri2023faith}.
We use the ListOps dataset \cite{nangia2018listops}, which consists of nested operations such as \texttt{MAX(2,3,ADD(4,5))}, extended with modular arithmetic and permutation groups.

\paragraph{Summary of contributions.}
We study how small transformer models learn nested operations---maximum (MAX), minimum (MIN), median (MED), modular addition (ADD), modular product (PROD), and alternating addition-subtraction (NADD: $\sum_{i} (-1)^i x_i$)---both in isolation and jointly.
We further extend to permutation groups, where products can be decomposed into sub-block operations.
Our main findings are:
\begin{enumerate}
    \item \textbf{Task difficulty hierarchy:}
    MAX/MIN are easiest; MED and PROD (non-prime modulus) are intermediate; ADD, NADD, and PROD (prime modulus) are hardest, requiring an order of magnitude more parameters.

    \item \textbf{Synergies that work:}
    Pairing easy tasks (MAX, MIN, MED, PROD) with hard tasks (ADD) reduces model size requirements by 2--7$\times$.
    Pretraining on MAX+MED then switching to ADD achieves similar reductions.

    \item \textbf{Synergies that fail:}
    Pairing structurally similar hard tasks (ADD+NADD) provides no benefit.
    Nor does pairing ADD with PROD when the modulus is prime.
    This indicates that task synergy depends on shared computational primitives, not superficial similarity.

    \item \textbf{Mechanistic evidence:}
    PCA of learned embeddings shows that successful joint training induces structured number representations (ordering, parity, divisibility) absent in single-task models.

    \item \textbf{Generalization beyond arithmetic:}
    The same effect appears in permutation groups: joint training with sub-block operations reduces model size by 5--6$\times$.
\end{enumerate}

These results demonstrate that joint training can guide models toward more efficient solutions, revealing structure in scaling laws that depends critically on task composition.
Understanding these synergies has practical implications: if certain task combinations make models ``smarter'' by inducing better representations, this could inform curriculum design for pretraining---a form of pre-pretraining that reduces the resources needed for downstream tasks.

\section{Related Work}

\textbf{Multi-task learning.}
Multi-task learning improves generalization by sharing representations across tasks \cite{caruana1997multitask, ruder2017overview}.
Recent work has shown benefits for arithmetic: \citet{lee2024teaching} demonstrated that joint training on multiple arithmetic operations improves each task through shared number representations, and \citet{abedsoltan2025task} showed that compositional joint training leads to sudden performance jumps.
However, these studies focused on task combinations that help, without systematically testing which pairings \emph{fail}.
Our work fills this gap: we identify specific task pairings that help (easy+hard) versus those that fail (hard+hard), and quantify the effect on minimum model size---showing reductions of 2--7$\times$, not just accuracy improvements.

\textbf{Curriculum learning.}
Presenting tasks in order of difficulty improves learning \cite{bengio2009curriculum, yin2024mumath, kim2024strategic}.
Our transfer experiments (Section~\ref{sec:transfer}) relate to curriculum learning, showing that pretraining on easy tasks enables learning hard tasks at smaller model sizes.
However, our focus is on simultaneous joint training rather than sequential curricula.

\textbf{Grokking and mechanistic interpretability.}
Work on grokking \cite{power2022grokking, liu2022towards} and mechanistic interpretability \cite{nanda2023progress} has shown that models solving modular arithmetic develop structured internal representations.
We build on these insights by using embedding analysis to understand \emph{why} joint training helps: successful task pairings induce structured embeddings (ordering, parity) that single-task training does not.
Our setup differs from grokking studies in that we vary model size at fixed dataset size, rather than observing delayed generalization at fixed model size.

We provide the first systematic map of task compatibility for arithmetic operations, extending to permutation groups and confirming robustness to architectural choices (Appendix~\ref{app:ablations}).

\section{Methodology}
We train small-scale transformer models and analyze both their performance and internal representations. 
This section outlines our methodology, detailing the dataset, model architecture, and evaluation protocols.
\paragraph{Choice of task.}
ListOps consists of nested mathematical equations involving operations. 
Our setup uses MAX, MIN, MED (median), ADD (addition modulo $n$), PROD (product modulo $n$) and NADD (alternating
addition and subtraction modulo $n$) applied to numbers ($0$ to $n-1$).
We conduct experiments using $n=10$ to $n=226$ (Appendix \ref{ap:listops-notation}).
ListOps provides procedural generation with controlled difficulty, exact evaluation, adjustable complexity via nesting depth, and systematic investigation of inter-task relationships.

\paragraph{Dataset description.}
We use a simplified functional notation of the form $f(x,y,\ldots)$.  
For example, \verb|max_10(3,min_10(7,4,9))=4<eos>|, where \verb|<eos>| denotes the end token. 
The subscript in the function name indicates the modulo operation e.g.,  
\verb|max_10| represents the \verb|max| function modulo $10$.  


\paragraph{Tokenization and CoT.}
We use word-level tokenization: each operation (e.g., \verb|add_20|), each number (e.g., \verb|13|), and each symbol (\verb|(|, \verb|)|, \verb|=|, \verb|<eos>|) is a single token (Appendix \ref{ap:token}).
Since directly solving nested expressions proves challenging (Fig.~\ref{fig:cot}), we use chain-of-thought (CoT) that decomposes expressions into intermediate steps (Appendix \ref{ap:cot}):
\verb|add_10(9 add_10(3 4))>add_10(9 7)>6=6<eos>|, where \verb|>| indicates solving the innermost operation.
This follows the scratchpad approach of \citet{lee2024teaching}. 
Task difficulty increases with the number of operands and nesting depth (Appendix Fig. \ref{fig:transition_point_length}).

\paragraph{Train-Test split.} 
The experiments were conducted on data generated with a maximum nesting depth of 2 and up to 3 operands per operation. 
Equations are constructed from digit sequences forming duplets and triplets (e.g., "1,0", "0,0,1"). 
Since the CoT format can produce many intermediate patterns within a single equation, we implement a rigorous train-test split to prevent data leakage: we designate 100 of the 1000 possible triplets (e.g. "749") as an exclusion set. 
Training equations never contain any excluded triplet, while test equations each contain at least one excluded triplet, ensuring that test set patterns are novel.

\paragraph{Permutation groups.}
To test generality beyond arithmetic, we extend ListOps to permutation groups.
By Cayley's theorem, every finite group is isomorphic to a permutation subgroup, making this a natural generalization.
We construct $6 \times 6$ block-diagonal permutation matrices (two $3 \times 3$ blocks) and define three operations: \verb|OP| (full matrix, 36 elements), \verb|OP_TOP| and \verb|OP_BOTTOM| (individual blocks, 6 elements each).
See Fig.~\ref{fig:permutation}a and Appendix Figs.~\ref{fig:perm_op}--\ref{fig:perm_elements} for details.


\paragraph{Performance evaluation.}
We randomly select 1000 
equations from the held-out test set.
The model is prompted to generate solutions character-by-character, starting from the equation prompt (e.g., \verb|add_10(3,min_10(7,4,9))>|). 
We use the output to produce two metrics:
\textbf{Loss:} using cross-entropy computed for every character of the output. 
\textbf{Accuracy:} based on the number of correct answers evaluated using \textit{only the final answer}, which we define as the first character after the first `$=$' symbol.  

\begin{figure*}[t!]
    \centering
    \includegraphics[width=\linewidth]{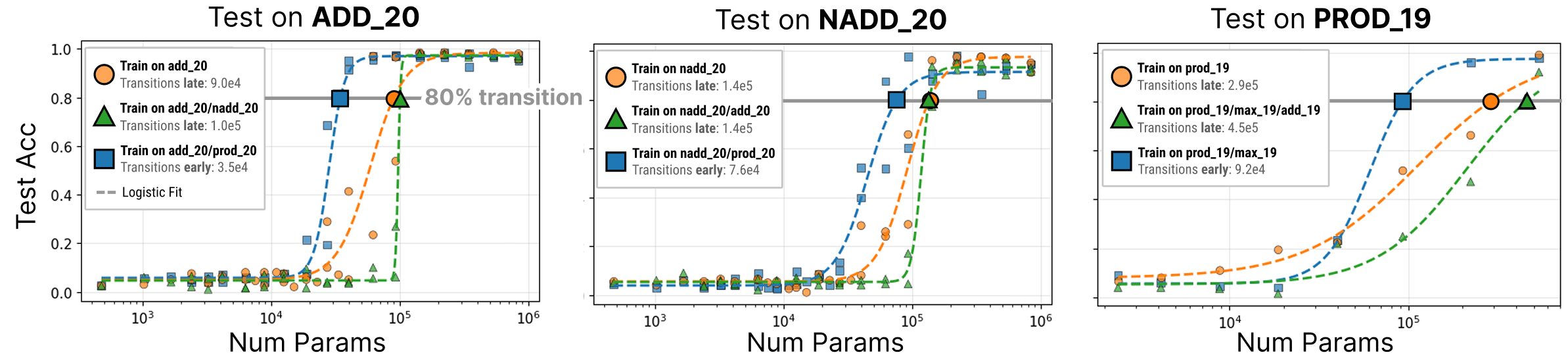}
    \caption{
    \textbf{Joint training with compatible tasks reduces the model size needed to learn.}
    Each panel shows accuracy vs.\ model size for operations trained in \textcolor[HTML]{FFA556}{\textbf{isolation}} ({\color[HTML]{FFA556}$\bullet$}), with an \textcolor[HTML]{1F77B4}{\textbf{easier task}} ({\color[HTML]{1F77B4}$\blacksquare$}), or with a \textcolor[HTML]{2CA02C}{\textbf{harder task}} ({\color[HTML]{2CA02C}$\blacktriangle$}).
    \textbf{Left:} ADD benefits from PROD ($2.5\times$ smaller models) but not from NADD.
    \textbf{Center:} NADD benefits from PROD but not from ADD.
    \textbf{Right:} PROD benefits from MAX ($2.6\times$ smaller) but not from ADD+MAX.
    Dashed lines show logistic fits; legends report the 80\% accuracy transition point.
    }
    \label{fig:joint-training}
\end{figure*}

\paragraph{Model architecture.}
For our experiments, we employed a series of tiny GPT models, inspired by the nanoGPT architecture \citep{karpathy2022nanogpt}. 
Unlike the standard sequential transformer, we implemented a recurrent variant in which a single transformer block was iteratively applied by feeding back its output as the next input (see the architecture Appendix Fig. \ref{fig:rec-gpt_sketch}). 
This recurrent design improved learning efficiency and yielded more structured embedding patterns (see results from traditional deep transformer models in the Appendix \ref{resutls_deep_transformer}). 
Each model in our study uses a single attention head. 
We set the feedforward hidden dimension to four times the embedding dimension, 
a common practice in transformer architectures, providing sufficient complexity in the feedforward networks while keeping the model size constrained. 
By varying the embedding size we create a range of models with different parameter counts.
Ablation studies with 1--4 attention heads show no qualitative change in results (Appendix~\ref{app:ablations}), and experiments with standard deep transformers (2--4 layers) yield similar patterns, though with less structured embeddings (Appendix~\ref{resutls_deep_transformer}).

\paragraph{Learning transition.}
We define the \emph{learning transition} as the minimum model size at which a task can be learned.
Operationally, we fit a logistic function to accuracy vs.\ model size and define the transition point as the model size reaching 80\% of the asymptotic accuracy.
We chose 80\% rather than 50\% because the latter corresponds to models that have not yet reliably mastered the task; 80\% ensures the model has learned a generalizable solution rather than partially memorizing patterns.
The transition curves are remarkably consistent across runs for the recurrent architecture with fixed dataset size.

\paragraph{Dataset scaling.}
When varying the modulo $n$, we scale the dataset size linearly with $n$.
This choice is motivated by the observation that smaller moduli (e.g., $n=10$) can be learned with relatively little data, while larger moduli (e.g., $n=113$) require substantially more examples to avoid underfitting.
We found that linear scaling produces transition points that grow slowly (often sub-linearly) with $n$ for most task combinations (Fig.~\ref{fig:scaling}), suggesting that data complexity requirements increase more slowly than linearly.
Importantly, the \emph{relative} benefits of joint training---the 2--7$\times$ reduction in model size---are consistent across moduli, indicating that our findings are robust to this design choice.

\section{Results}

\begin{figure*}[t]
    \centering
    \includegraphics[width=0.68\linewidth]{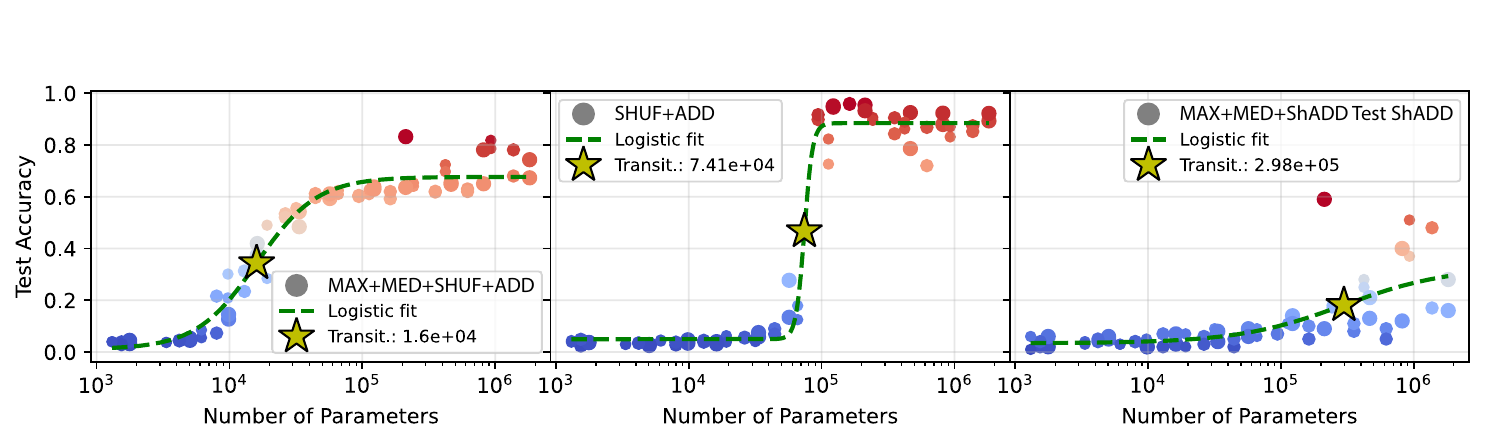}
    \includegraphics[width=0.26\linewidth]{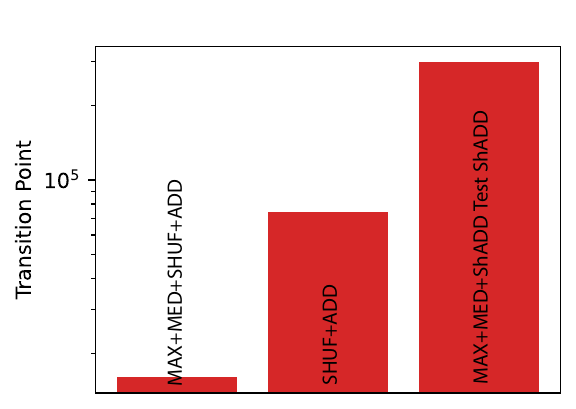}
    \caption{\textbf{Shuffled ADD: joint training fails when number structure is destroyed.}
    \textbf{Left:} Learning curves for various tasks including Shuffled ADD (randomly permuted addition table).
    \textbf{Right:} Transition points---Shuffled ADD is nearly as hard as pure ADD.
    Unlike regular ADD, joint training with MAX+MED \emph{hurts} Shuffled ADD: the mixed model reaches only $\sim$20\% accuracy (third panel).
    This confirms that MAX+MED induces number-like embeddings incompatible with the randomized table.
    }
    \label{fig:transition-shuffled-sum-26-40k-main}
\end{figure*}

To understand how language models acquire mathematical abilities, we focus on accuracy as the primary metric for observing the emergence of learning, following the approach of the emergent abilities literature \cite{wei2022emergent}. 
While some subsequent studies have questioned the concept of emergence by examining other metrics \citep{schaeffer2023emergent}, we argue that this critique overlooks similar patterns observed in physics: 
in a phase transition, only some quantities may exhibit discontinuous changes. 
A metric that can capture the emergence of the new phase is called the ``order parameter''. 
Hence, we choose to use accuracy as our order parameter. 

Experiments use a recurrent transformer with embedding dimension $n_{embed}$ ranging from 8 to 362, and 4 recurrence steps (we have found that using 4 recurrence steps yields the best performance). 
Similar to the emergence literature, we observe the total number of parameters to be the strongest indicator of accuracy, rather than $n_{embed}$ or depth. 
Our experimental design focuses on identifying the minimum model size sufficient for learning each task or task combination—once this threshold is exceeded, larger models reliably reach high accuracy, making further scaling unnecessary for our analysis. 
We primarily present results for modulo 20, with similar patterns observed for other modulos (Appendix \ref{ap:mod10} - mod $10$ results, \ref{ap:mod26} - mod $26$ results). 
Experiments with higher moduli reveal a basic scaling relationship: larger moduli require proportionally larger models and more training data to achieve comparable performance.

We find that joint training  on compositional tasks substantially improves performance and reduces the parameter requirements for learning individual operations. 
Training on multiple arithmetic tasks helps models learn the basic building blocks needed for solving more complex problems. 
This shared representation not only accelerates learning but also produces abrupt transitions in performance once the necessary constituent skills are mastered \cite{lubana2024percolation, okawa2023compositional}. 
However, joint training is not universally beneficial, and task compatibility is more nuanced than surface-level similarity suggests.
Our experiments reveal that effective transfer requires overlap in underlying computational primitives and mathematical structures, not merely similar algorithms or comparable difficulty.
Pairing hard problems together often increases required model size when tasks lack shared low-level building blocks, demonstrating that joint training can actively interfere with learning rather than facilitate it.
These findings extend previous work on compositional generalization \cite{lee2024teaching} by systematically characterizing when multi-task learning helps or hinders algorithmic task performance, showing that compatibility depends on deep structural properties rather than intuitive task similarity.

\paragraph{Transition to learning.}
Figure \ref{fig:joint-training} shows learning transitions for single and joint operations. 
We find that joint training enables models to solve more difficult tasks with fewer parameters and less data.  
For example, ADD and NADD are the most demanding tasks when trained in isolation, yet when combined with PROD, models that are $\times$2.5 smaller are able to reach 100\% accuracy on the task. 
Similarly, while PROD on prime moduli is challenging on its own, pairing it with MAX reduces the required model size by a factor of $\times$2.6. 
In contrast, pairing ADD with NADD does not decrease the required model size, nor does pairing PROD-prime with ADD or NADD.

These results challenge the assumption that harder tasks always demand larger models.
Joint training helps only when tasks share the right structure: ADD and NADD use similar algorithms but require different representations, so pairing them fails.
Meanwhile, PROD with non-prime modulus helps ADD because divisibility patterns create shared computational primitives.
The key insight: task synergy requires overlapping low-level building blocks, not superficial similarity.
These patterns extend to combinations of 3--4 tasks: joint training on all combinations of MAX, MIN, MED, and ADD (modulo 26) shows the same grouping of transition points by task compatibility (Appendix Fig.~\ref{fig:transition26}).

\paragraph{Testing the number properties hypothesis.}
We hypothesize that joint training helps because easier tasks reveal number properties (e.g., ordering, parity) that benefit harder tasks.
To test this, we designed an experiment that destroys number relationships: we shuffle the addition table so that $A+B=C$ becomes a random mapping with no arithmetic structure (Appendix \ref{shuffled_sum}).
If joint training benefits ADD by revealing number properties, then Shuffled ADD should not benefit from joint training with MAX and MED.

We conducted experiments using a shuffled, symmetric ADD table ($A+B = B+A$) in modulo 26.
The shuffled addition proved more difficult to learn than all other mathematical operations except the original ADD (Fig. \ref{fig:transition-shuffled-sum-26-40k-main}).
Crucially, MAX+MED+Shuffled ADD was \emph{harder} than Shuffled ADD alone: the mixed model reached only 60\% accuracy on Shuffled ADD and required twice as many parameters.
This is the opposite of what we observe with regular ADD, where joint training helps dramatically.

This confirms our hypothesis: MAX and MED induce number-like embeddings with ordering structure, but Shuffled ADD is incompatible with these representations.
PCA analysis shows Shuffled ADD embeddings lack parity separation or ordering (Appendix Fig. \ref{fig:corr-PCA-shuffle-26}).

\subsection{Embedding Analysis: What Number Properties Emerge?}

The Shuffled ADD experiment confirms that number properties are key to joint training benefits.
We now examine \emph{what specific properties} emerge in the learned embeddings.
To do this, we analyze the embedding layer using principal component analysis (PCA).

\paragraph{PCA methodology.}
In our tokenization scheme, each number and operation is a single token.
While embedding layers of different-sized models cannot be directly compared (they have different dimensions), we can compare their $n_{\text{vocab}} \times n_{\text{vocab}}$ cosine similarity matrices.
To reduce noise, we average similarity matrices across models that learned each task (accuracy $>90\%$), then compute PCA on the averaged matrix.
This reveals consistent patterns across model sizes (see Appendices \ref{ap:mod26} and \ref{ap:mod10} for additional results).

\begin{figure*}[t!]
    \centering
    \includegraphics[width=0.95\linewidth]{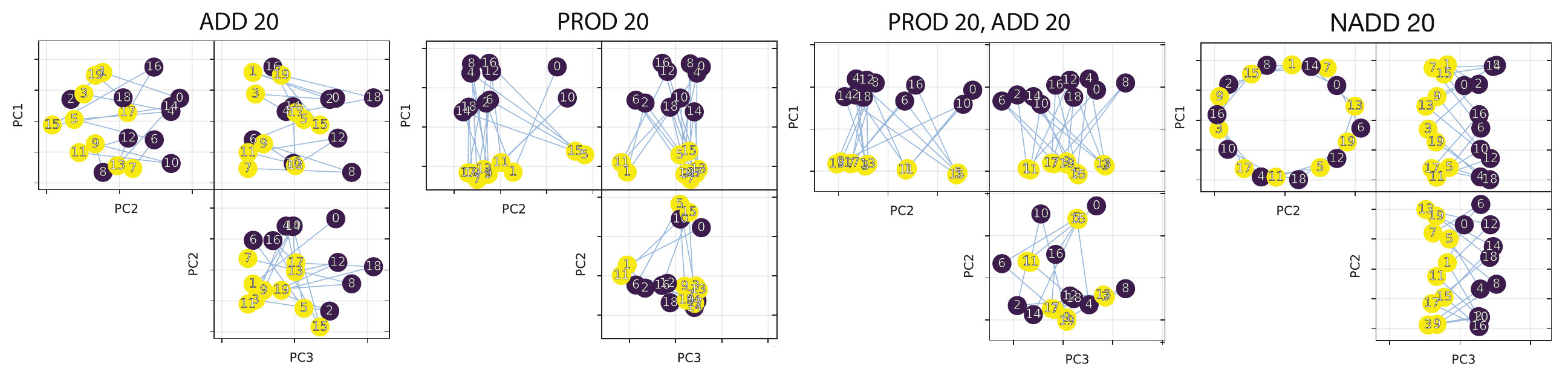}
    \caption{
        \textbf{PCA of embeddings reveals structured number representations in joint training.}
        Top 5 principal components of the cosine similarity matrix, colored by parity (yellow=odd, purple=even).
        \textbf{PROD:} Strong parity separation and ordering structure.
        \textbf{ADD+PROD:} Preserves PROD's structure, explaining why joint training helps ADD.
        \textbf{Pure ADD:} Weak, noisy structure.
        \textbf{NADD:} Highly structured but different from ADD.
        \textbf{ADD+NADD:} Loses structure from both tasks, explaining why this pairing fails.
    }
    \label{fig:joint-PCA-emb}
\end{figure*}

\paragraph{Structured embeddings in joint training.}
Figure \ref{fig:joint-PCA-emb} shows that joint training induces structured embeddings: PROD shows strong parity separation and ordering; ADD+PROD preserves this structure; pure ADD shows weak structure; NADD develops different structure; and ADD+NADD loses structure from both tasks.
This explains why ADD+NADD fails despite algorithmic similarity: incompatible representations.

\begin{figure*}[t!]
    \centering
    \includegraphics[width=0.95\linewidth]{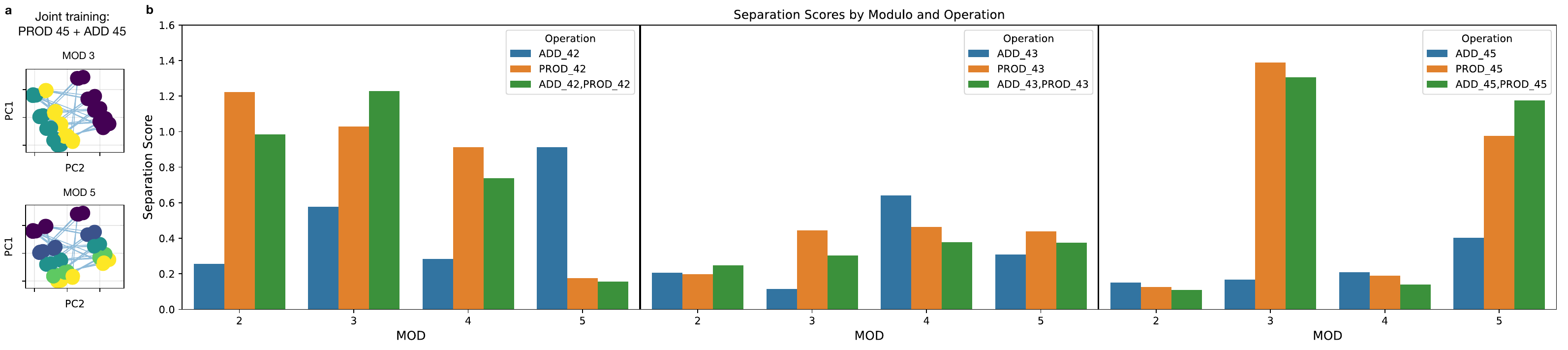}
    \caption{
    \textbf{Embeddings cluster by divisors of the modulus.}
    \textbf{(a)} PCA of PROD-45+ADD-45 embeddings colored by remainder class shows clear mod-3 and mod-5 clustering (divisors of 45).
    \textbf{(b)} Separation score (between-centroid distance / within-group variance) quantifies clustering strength across moduli.
    Non-prime moduli (42, 45) show strong clustering by their divisors; prime modulus (43) shows none.
    ADD+PROD inherits PROD's divisibility structure; pure ADD shows weak clustering.
    This explains why PROD helps ADD only for non-prime moduli.
    }
    \label{fig:sep-emb-main}
\end{figure*}

\paragraph{Quantifying embedding structure: separation score.}
Beyond parity, we observe that embeddings cluster numbers by their remainders modulo divisors of $n$.
To quantify this, we define a \emph{separation score}: for each candidate divisor $d$, we group numbers by their remainder mod $d$, compute centroids for each group in the top PCs, and measure the ratio of between-group distance to within-group standard deviation.
High separation scores indicate strong clustering by that divisor (Fig.~\ref{fig:sep-emb-main}).

The results reveal: PROD embeddings cluster by divisors of $n$ (e.g., mod-2, 3, 6 for $n$=42); prime moduli show no clustering, explaining why PROD-prime is hard; and ADD+PROD inherits PROD's divisor structure while pure ADD shows weak clustering.
This explains why PROD helps ADD only for non-prime moduli: divisibility structure creates transferable primitives.

\subsection{Learning Dynamics: How Structure Emerges}

The embedding analysis shows that joint training produces structured representations.
But \emph{when} does this structure emerge during training?
We examine the learning dynamics to understand how easy tasks scaffold hard tasks.

\begin{figure}
    \centering
    \includegraphics[width=\linewidth]{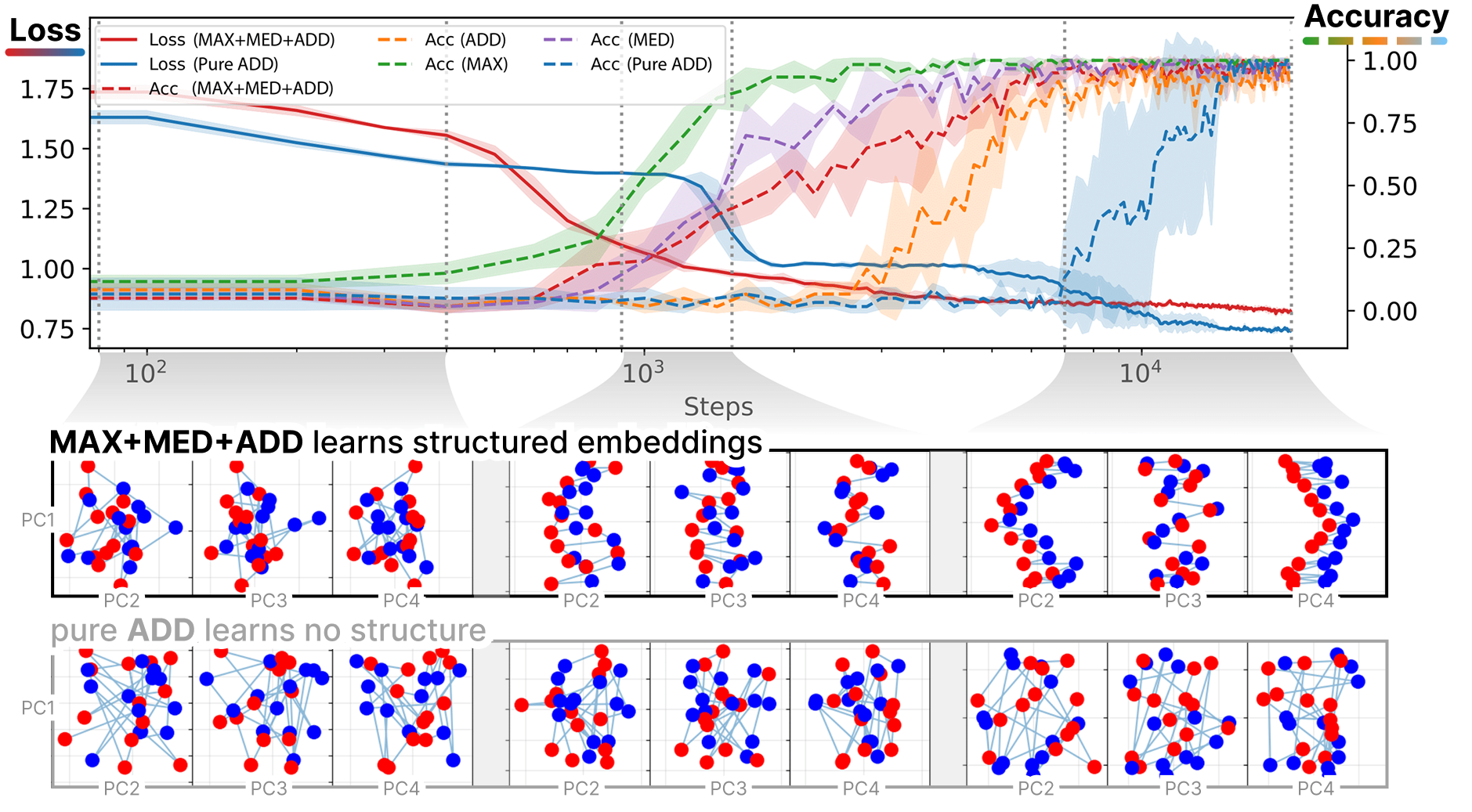}
    \caption{
    \textbf{Joint training improves training speed by encouraging structured embeddings during learning dynamics.} (Top) Two models are trained: one on a joint MAX+MED+ADD dataset and one on a pure ADD dataset (mod 26). We observe that both training loss (solid) and accuracies (dashed) improve \emph{faster} for the joint training model across all evaluated tasks (mean $\pm\sigma$ over 3 runs.). 
    In joint training, MAX learns first, then MED, then ADD learns only after the embeddings become structured.
    (Bottom) PCA snapshots are taken during training (number parity is colored red/blue). Joint training accuracies rise quickly after the embeddings learn structure, which only happens in the joint training model. The pure ADD model shows no embedding structure and exhibits loss plateaus, suggesting it struggles to discover an efficient algorithm.
    }
    \label{fig:evo-loss-acc-pca-26}
\end{figure}
Figure \ref{fig:evo-loss-acc-pca-26} compares pure ADD vs.\ MAX+MED+ADD ($n_{\text{embed}}=96$).
In joint training, MAX learns first, then MED, then ADD---and structured embeddings emerge \emph{before} ADD accuracy rises, suggesting structure enables learning.
Joint training converges in half the steps; pure ADD shows no structure and loss plateaus. 

\subsection{Causal Test: Can Learned Structure Transfer?}\label{sec:transfer}

\begin{figure*}[t!]
    \centering
    \includegraphics[width=\textwidth]{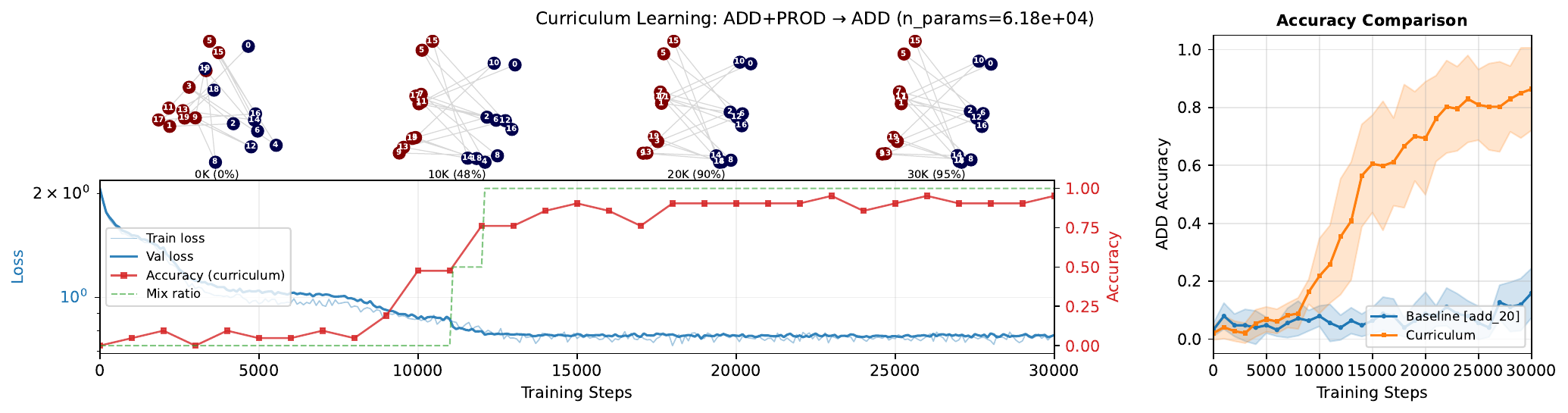}
    \caption{\textbf{Curriculum learning enables transfer of structured representations.}
    \textbf{Left:} PCA of embeddings at 0K, 10K, 20K, and 30K steps (top), colored by parity (red=odd, blue=even).
    Structure emerges during joint ADD+PROD training (mix ratio rises at 10K) and persists as the curriculum shifts to pure ADD.
    Loss and accuracy curves (bottom) show the curriculum transition.
    \textbf{Right:} Baseline ADD stays at $\sim$15\% accuracy; curriculum reaches $>$95\%. Shaded: $\pm 1\sigma$ over runs.
    }
    \label{fig:curriculum}
\end{figure*}

To establish causation, we test whether structure from easy tasks \emph{transfers} to hard tasks.
We design a curriculum (Fig.~\ref{fig:curriculum}): train on ADD+PROD, gradually shift to pure ADD, and compare to a baseline trained only on ADD.

The results are striking: baseline ADD stays at $\sim$15\% accuracy, but curriculum achieves $>$95\%.
Structured embeddings develop during ADD+PROD and persist through the transition.

We also verified this effect with other task combinations: pretraining on MAX+MED followed by transition to pure ADD yields similar results, with models 7$\times$ smaller than the pure ADD transition successfully learning the task (see Appendix~\ref{app:curriculum} for details).
\subsection{Scaling Across Moduli}

\begin{figure}[t]
    \centering
    \includegraphics[width=\linewidth]{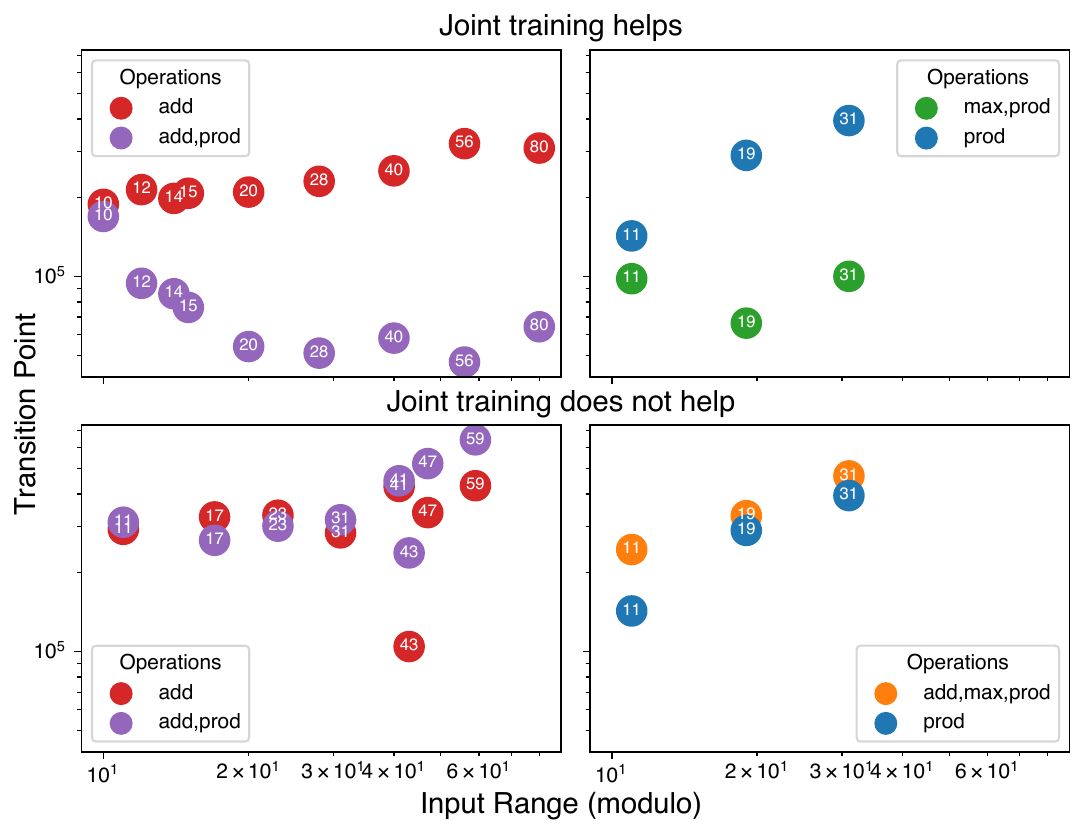}
    \caption{
    \textbf{Joint training helps only with compatible tasks} when scaling across moduli.
    \textbf{Top (helps):}
    ADD+PROD (left) reduces transition points by $3.6\pm1.6\times$ for non-prime moduli, while
    MAX+PROD (right) reduces PROD transition by $3.3\pm1.3\times$ even for prime moduli.
    \textbf{Bottom (no benefit):}
    ADD+PROD (left) provides no benefit for prime moduli (ratio $0.85\pm0.25\times$), while
    ADD+MAX+PROD (right) \emph{increases} PROD-prime transition (ratio $0.77\pm0.13\times$).
    Modulus values are indicated on the x-axes and as numbers on each scattered point.
    }
    \label{fig:scaling}
\end{figure}

Do the benefits (and failures) of joint training persist across different moduli?
We systematically measure transition points for various task combinations across moduli ranging from 10 to 80 (Fig.~\ref{fig:scaling}).
The results cleanly separate into cases where joint training helps versus cases where it does not.

\paragraph{Joint training helps:} ADD+PROD reduces transitions by $3.6\pm1.6\times$ for non-prime moduli; MAX+PROD reduces PROD transitions by $3.3\pm1.3\times$ even for primes (MAX provides ordering structure).

\paragraph{Joint training fails:} ADD+PROD provides no benefit for prime moduli ($0.85\pm0.25\times$); ADD+MAX+PROD actually \emph{hurts} PROD-prime ($0.77\pm0.13\times$).
These patterns hold robustly across moduli (Appendix Fig.~\ref{fig:scaling-full}).

\subsection{Generalization Beyond Arithmetic: Permutation Groups}

To test generality, we extend ListOps to permutation groups with no numerical structure.
We construct $6\times 6$ block-diagonal permutation matrices (two $3\times 3$ blocks) with three operations: \verb|OP| (full matrix, 36 elements), \verb|OP_TOP| and \verb|OP_BOTTOM| (individual blocks, 6 elements).
The sub-block operations decompose the full operation into simpler components---analogous to how MAX helps ADD.

\begin{figure*}[t!]
    \centering
    \includegraphics[width=0.95\linewidth]{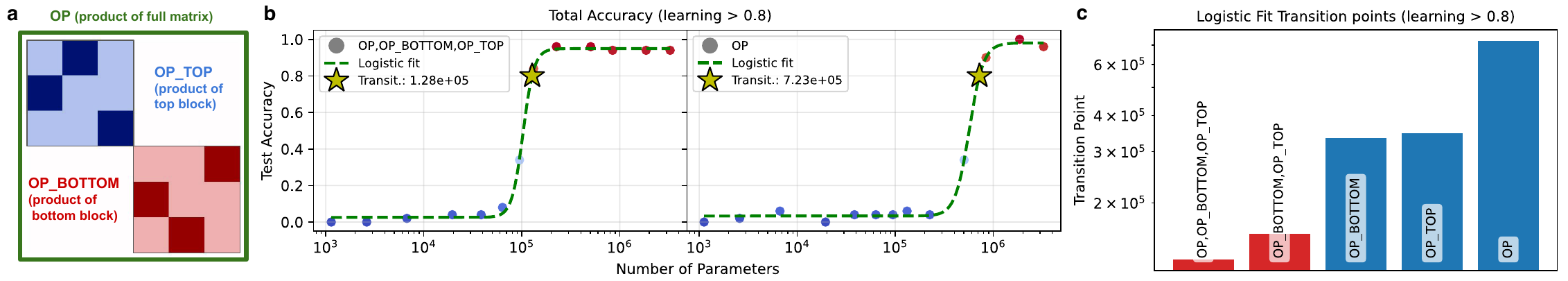}
    \caption{
    \textbf{Joint training benefits generalize to permutation groups.}
    \textbf{(a)} Setup: $6\times 6$ block-diagonal permutation matrices with two $3\times 3$ blocks.
    OP acts on the full matrix (36 elements); OP\_TOP and OP\_BOTTOM act on individual blocks (6 elements each).
    \textbf{(b)} Transition curves: joint training vs.\ pure OP.
    \textbf{(c)} Sub-block operations reduce the OP transition point by $5\text{--}6\times$, confirming that decomposing complex operations into simpler primitives helps learning---even in non-numerical domains.
    }
    \label{fig:permutation}
\end{figure*}

Joint training yields a \textbf{5--6$\times$ reduction} in model size (Fig.~\ref{fig:permutation}b,c).
By Cayley's theorem, every finite group is isomorphic to a permutation subgroup, suggesting these principles apply broadly: decomposing complex operations into simpler primitives helps learning, regardless of numerical structure.

\section{Discussion}

\paragraph{Why does joint training help?}
Our results suggest that joint training benefits coincide with better representation learning: easier tasks like PROD and MAX restrict the embedding space to representations capturing number properties, which then support harder tasks like ADD.
This aligns with grokking literature \cite{power2022grokking, liu2022towards}, where successful generalization coincides with organized internal representations.
The path to learning matters: models developing structured embeddings early require fewer resources. Our transfer experiments show models pretrained on MAX+MED learn ADD at 7$\times$ smaller sizes.

\paragraph{Toward a predictive theory: output distributions.}
We hypothesize that task difficulty relates to the output distribution of $\text{Op}(a,b)=c$: tasks with uniform output distributions (like ADD) offer no patterns to exploit, while tasks with peaked distributions (like PROD at non-prime moduli) allow divide-and-conquer strategies---learning frequent outputs first, then rarer ones.
This explains why PROD difficulty changes drastically with modulus: at primes, the output distribution is nearly uniform; at non-primes, divisibility creates structure.
The parity and modular clustering we observe in embeddings may emerge from such divide-and-conquer: e.g., in ADD$_{20}$, even+even and odd+odd both yield even results, creating exploitable structure.
Our separation score (Fig.~\ref{fig:sep-emb-main}) provides quantitative evidence: tasks inducing higher separation by divisors tend to help harder tasks more.

\paragraph{Beyond arithmetic.}
Our permutation group results suggest these effects extend beyond arithmetic.
By Cayley's theorem, every finite group is isomorphic to a permutation subgroup, so these principles may apply to a large class of algebraic tasks.

\paragraph{Connections to curriculum learning.}
Joint training and curriculum learning both exploit task relationships.
Our transfer experiments show that sequential curricula benefit from the same mechanism: structured representations learned on easy tasks transfer to hard tasks.
This suggests a general principle: pair tasks sharing computational primitives, not merely similar difficulty.

\paragraph{Algorithmic complexity.}
Minimum model size depends not just on task complexity but on the \emph{algorithmic complexity} of the discovered solution---brute-force memorization requires far more resources than efficient computation.
Joint training appears to bias models toward efficient algorithms by constraining the representation space.

\paragraph{Broader applicability.}
Prior work shows training arithmetic with text improves accuracy \cite{lee2024teaching}, and structured patterns help downstream tasks \cite{yin2024scaffolding, zhang2024intelligence}, suggesting these principles may transfer across domains.

\paragraph{Limitations.}
Our synthetic tasks may not directly transfer to natural language with ambiguous outputs, though they suggest principles for curriculum design in pretraining.
We focus on small models; effects at larger scales remain untested.
Our PCA-based analysis captures only linear structure, and while our output distribution hypothesis provides partial predictions, a complete theory of task compatibility remains future work.

\section*{Impact Statement}

This paper presents work whose goal is to advance the field of Machine Learning.
Our findings suggest that strategic task composition can reduce model size requirements by 2--7$\times$, potentially enabling more energy-efficient training and reducing computational costs.
We study synthetic tasks and do not foresee direct negative societal consequences from this basic research.



\bibliography{bib}
\bibliographystyle{icml2026}

\newpage
\appendix
\onecolumn

\nd{
What are the key questions definng this project? 
What answers do we find? 
What are the most exciting points about the results? 

We wanted to understand how emergent abilities are related to the task complexity. 
We needed to find a way to estimate complexity quantitatively. 
We also wanted exact evaluation, so we chose listops. 

So how do we want to start the paper? 
Currently, we talk about scaling laws etc, which is good. 
But I'm wondering if there is a more direct

}

\nd{Flow: 
We want to work our way toward two major findings: 
1) joint training leads to learning number properties and more efficient algorithms. 2) complexity plot. 

Findings for number learning
\begin{enumerate}
    \item Sum alone seems to not learn numbers generally (check how often it does using cos corr) 
    \item PCA of embedding reveals all3 learns strong number features 
    \item weights: attention seems more important in all3 than in sum alone. 
    We observe significant differences. 
\end{enumerate}

PCA findings:
\begin{enumerate}
    \item max: sequentiality of numbers  (and med?) 
    \item max: arc, not clear why but maybe longer-range relations, or simply next eigenvec of such a line graph? 
    \item sum: alone usually has no pattern of correltion
    \item all3: 1) sequentiality, 
    \item all3: 2) more pronounced arc, almost a circle, akin to Whittaker-Shannon number rep. (write subsection for it) 
    \item all3: 3) odd-even, very curious as not necessary for any of the 3 ops. (check if binary number rep show this)
    \item Do consistency check among PCA (how?)
\end{enumerate}

Findings from the complexity plot: 
\begin{enumerate}
    \item joint training learns significantly earlier 
    \item the system seems to learn number properties 
    \item using KC of algorithms using symbols only and using number properties, in the right cases, yields a power law pattern for KC vs num params. 
\end{enumerate}

We should mention the hysteresis an its curious absence in all3. 
So maybe we should use a different measure for the phase transition and plotting. 
The average over params is misleading, as it conceals the hysteresis.  
Some of the analysis needs to be updated, including complexity plot and transitions. 
}


\section{Related works}

\textbf{Scaling laws} for language models characterize how validation loss scales with model size (number of parameters) \cite{hoffmann2022training}, compute (FLOPs) \cite{muennighoff2023scaling}, dataset size (number of examples) \cite{hestness2017deep}, and information capacity (in bits) \cite{allen2024physics}, offering a foundation for model design. However, this perspective overlooks critical aspects of learning. Our results show that the composition of the training data—specifically, the mix of tasks—can significantly alter what and how a model learns. Even models with identical sizes can follow different learning trajectories depending on the training setup, suggesting that task structure shapes the internal algorithms that emerge during training.

\textbf{Joint training}
Joint training on compositional tasks allows models to acquire the foundational primitives necessary for solving complex operations, often leading to abrupt performance improvements once all constituent skills are learned \cite{lubana2024percolation, okawa2023compositional}. Similarly, training on multiple arithmetic tasks has been shown to improve accuracy on individual operations, highlighting the benefits of shared representations for compositional generalization \cite{lee2024teaching}.

\textbf{Number representation} 
Neural networks trained on modular addition tasks develop structured embedding spaces that reflect underlying arithmetic operations. Mechanistic analyses have shown that small models can exhibit ordered number patterns and implement distinct algorithmic strategies, depending on initialization and hyperparameters \cite{zhong2023clock}. Some models converge to known solutions such as the Clock algorithm, while others discover novel procedures like the Pizza algorithm, illustrating the algorithmic diversity that can emerge from fixed training data. Periodic structures in the embeddings can be characterized via Fourier analysis, offering additional interpretability \cite{nanda2023progress}. These behaviors have also been linked to grokking dynamics, where models abruptly generalize after extended training, accompanied by the emergence of structured embedding patterns \cite{power2022grokking, liu2022towards}.

In general language models performs poorly on symbolic mathematical tasks \cite{frieder2023mathematical,  dziri2023faith, dave2024investigating} such as the ListOps dataset \cite{nangia2018listops} used in this study. 
Models often struggle with generalization, tending to memorize tasks rather than simulate the underlying algorithms. 
While mathematical tasks prove challenging for large language models to learn, they provide a controllable playground to test how models learn different tasks and to evaluate their accuracy quantitatively. 
Furthermore it allows us to tune the task difficulty by combining different operations. 


Here, we investigate how language models acquire arithmetic skills by training small models on the ListOps dataset, which enables explicit evaluation through structured mathematical expressions. 
Adopting a bottom-up approach, we find that models learn to solve these tasks once the number of trainable parameters surpasses a critical threshold. 
This threshold shifts with the difficulty of the target operation (MAX = MIN $<$ MED $<$ ADD), indicating a dependency on task difficulty. Surprisingly, joint training on multiple operations often facilitates learning, outperforming models trained on individual operations (e.g. MAX+MED+SUM $<$ ADD). 
This suggests that task diversity can ease optimization by promoting shared representations.

\clearpage
\section*{Supplemental Material}

\section{Additional plots}

\subsection{Modulo 20\label{ap:mod20} }

\begin{figure}[!htbp]
    \centering
    \includegraphics[width=.8\linewidth]{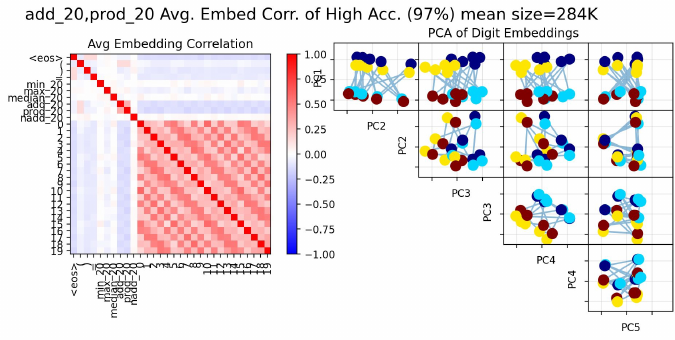}
    \caption{
    \textbf{ADD+PROD embeddings and PCs. PCs colored based modulo 4.}
    }
    \label{fig:color_pca_addprod}
\end{figure}

\begin{figure}[!htbp]
    \centering
    \includegraphics[width=\linewidth]{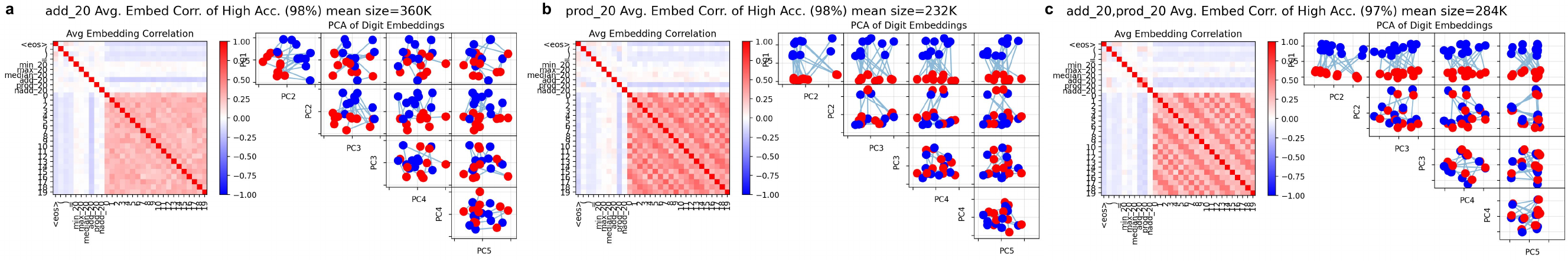}
    \caption{
    \textbf{ADD, PROD, ADD+PROD embeddings and PCs.}
    }
    \label{fig:pca_prod}
\end{figure}

\subsection{Modulo prime - PROD\label{ap:prod_prime} }
\begin{figure}[!htbp]
    \centering
    \includegraphics[width=\linewidth]{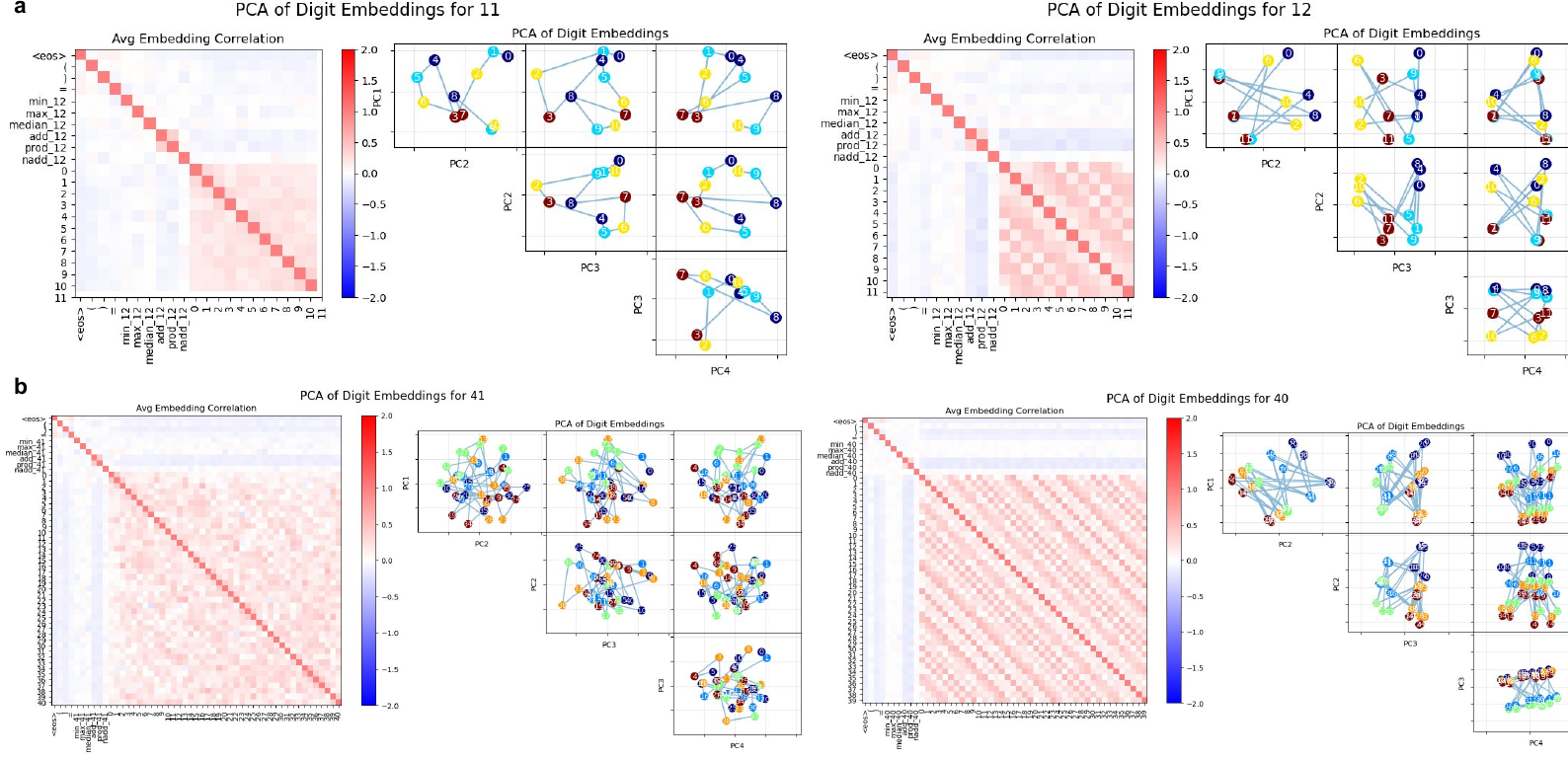}
    \caption{
    \textbf{PCs PROD modulo prime vs not prim. }
    }
    \label{fig:pca_prod_prime}
\end{figure}

\clearpage

\section{Permutation groups.}

\begin{figure}[!htbp]
    \centering
    \includegraphics[width=.95\linewidth]{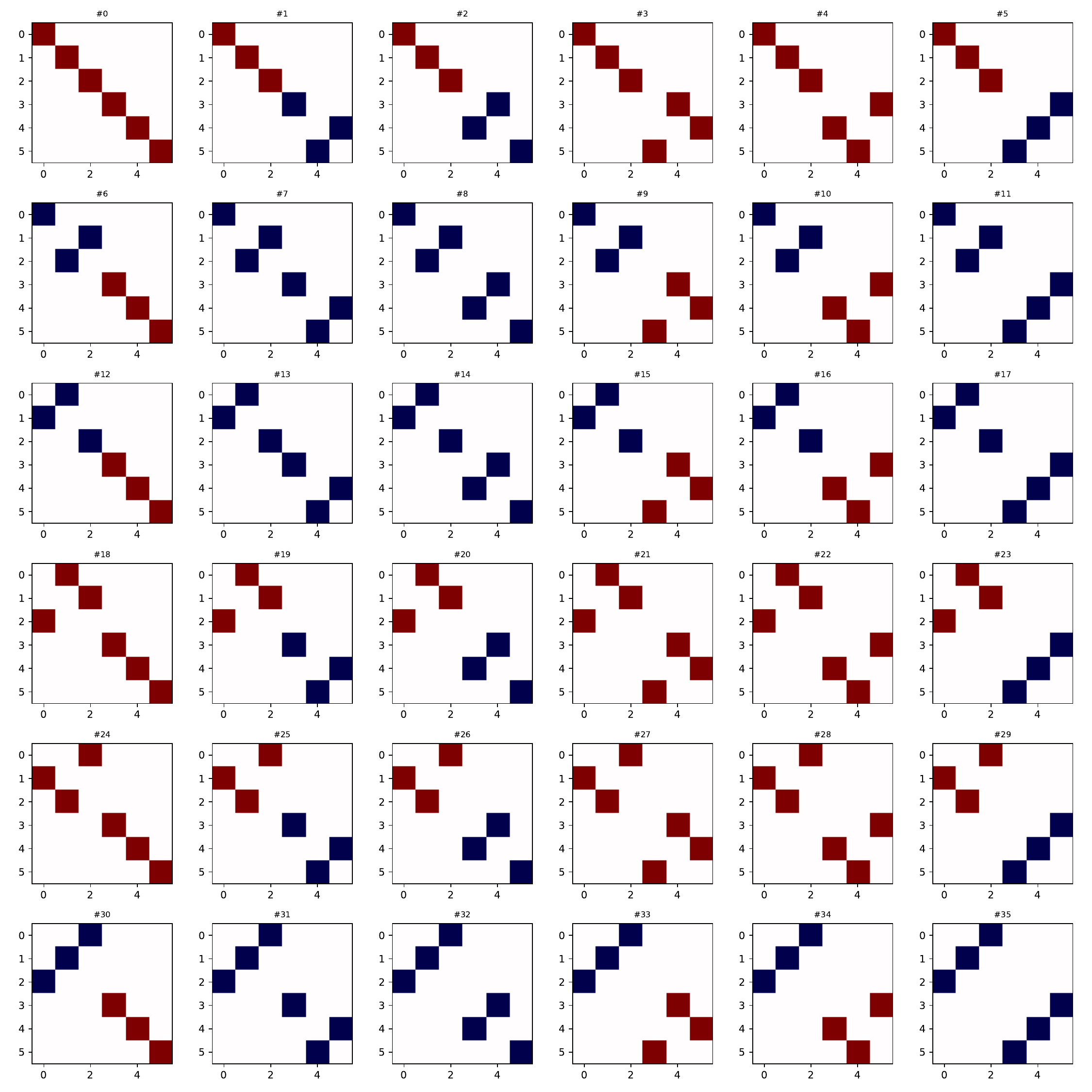}
    \caption{
    \textbf{Permutation - block diagonal with two 3x3 block matrix}
    Red indicates a cycle sub-block, while blue indicates a non-cycle block.
    }
    \label{fig:perm_elements}
\end{figure}

\begin{figure}[!htbp]
    \centering
    \includegraphics[width=.5\linewidth]{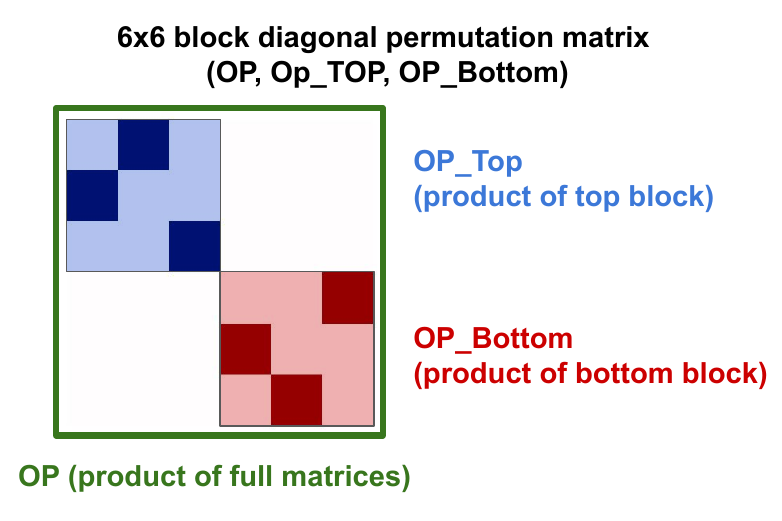}
    \caption{
    \textbf{6x6 block diagonal permutation operations(OP, Op\_TOP, OP\_Bottom)} OP\_Top only acts on the top block matrix and OP\_Bottom only acts on the bottom block matrix, while OP is an operation that acts on both the top and bottm blocks. }
    \label{fig:perm_operator_sketch}
\end{figure}

\begin{figure}
    \centering
    \includegraphics[width=0.3\linewidth]{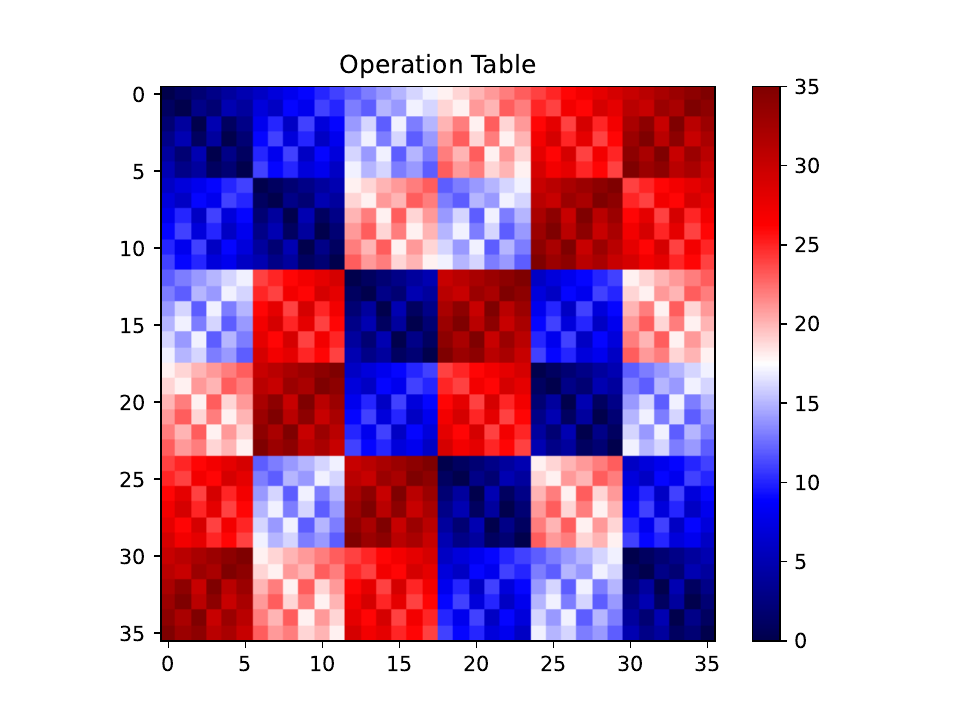}
    \includegraphics[width=0.3\linewidth]{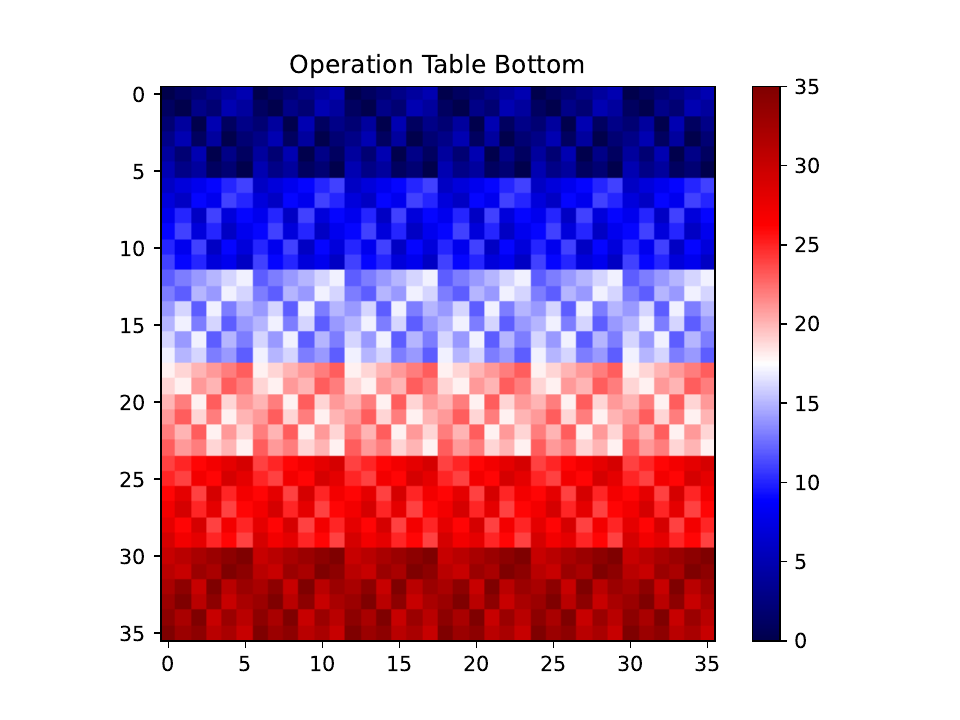}
    \includegraphics[width=0.3\linewidth]{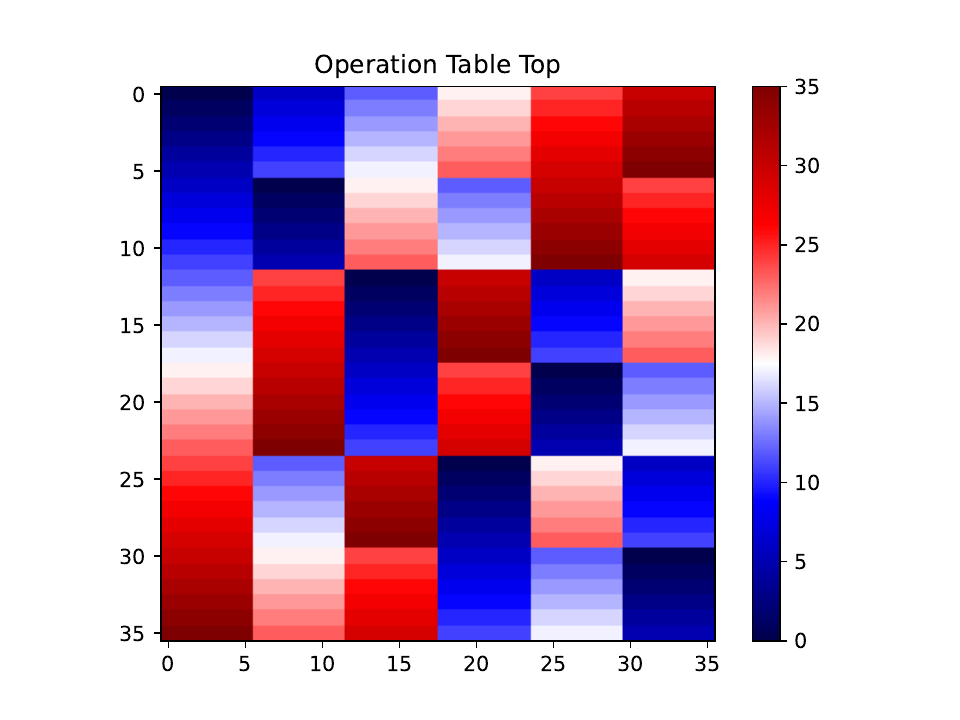}
    \caption{
    \textbf{Permutation data structure.}
    Left: 
    Operation table (OP) - block diagonal with two 3x3 block matrix. Middle: Operation table (OP\_BOTTOM) - block diagonal with two 3x3 block matrix. 
    Right: Operation table (OP\_TOP) - block diagonal with two 3x3 block matrix
    }
    \label{fig:perm_op}
    \label{fig:perm_op_bottom}
    \label{fig:perm_op_top}
\end{figure}

\begin{figure}
    \centering
    \includegraphics[width=0.49\linewidth]{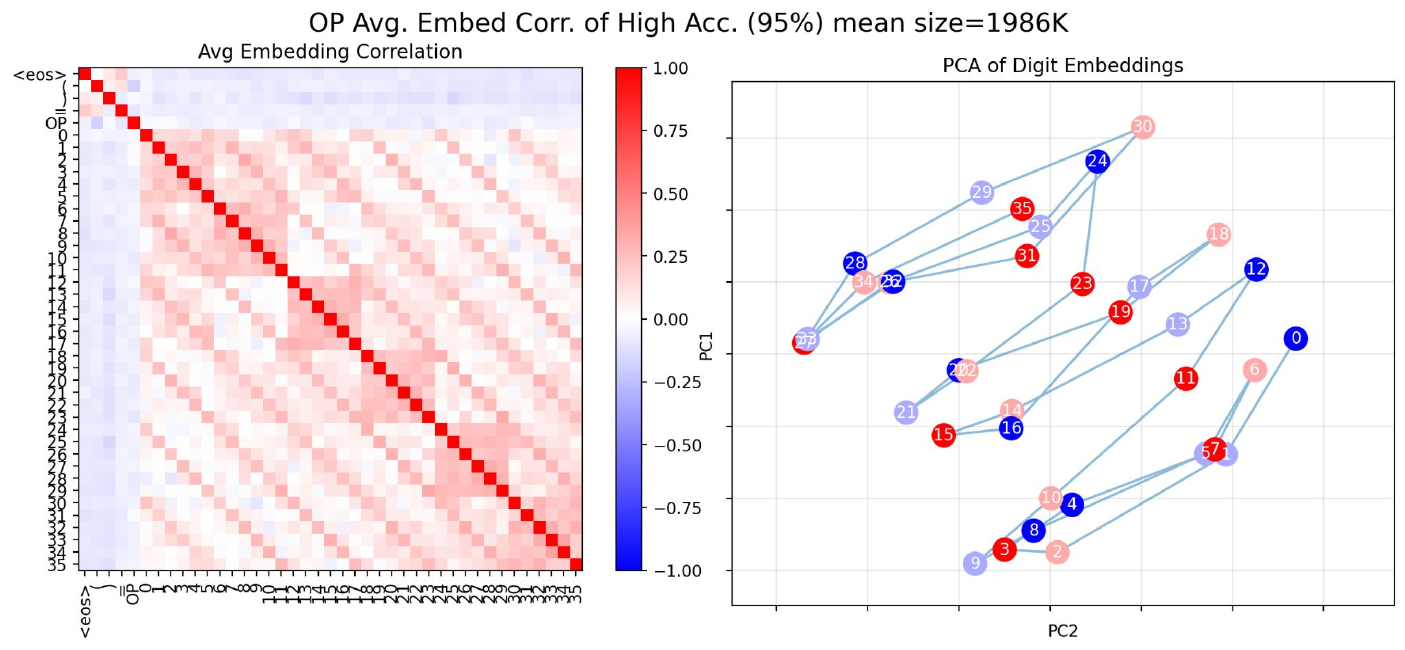}
    \includegraphics[width=0.49\linewidth]{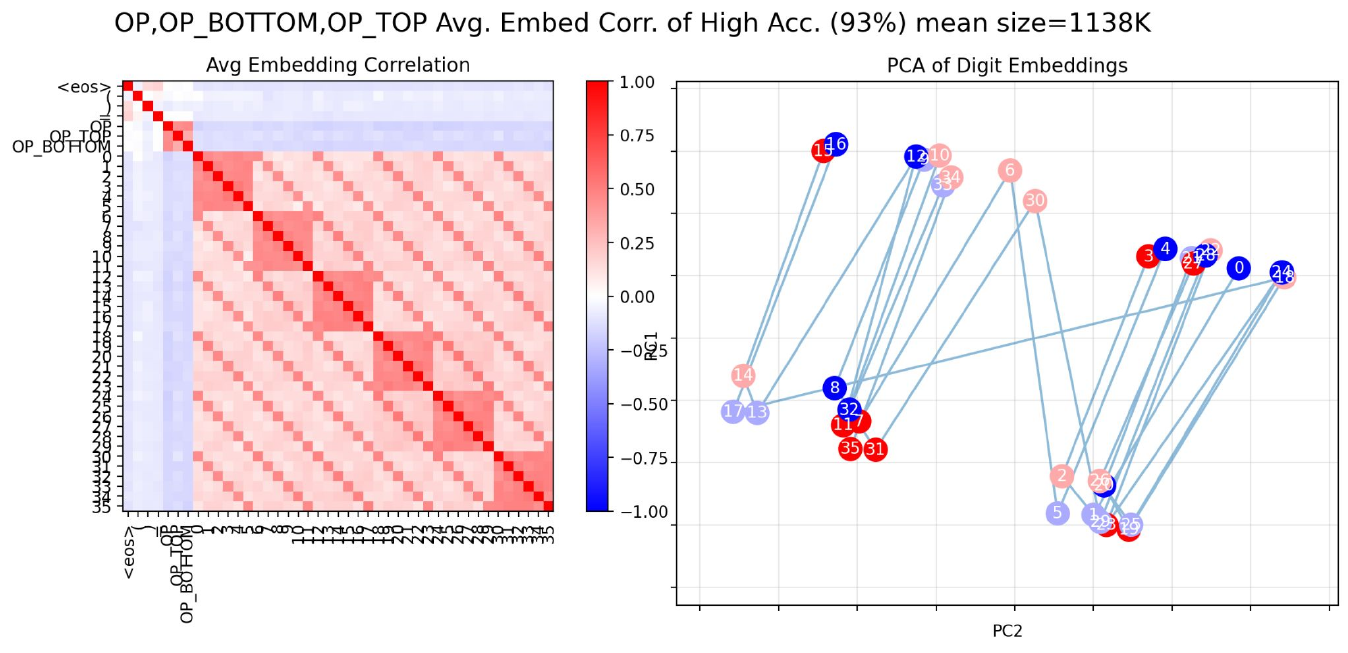}
    \caption{\textbf{Permutation experiments:} The correlation matrix and top 2 PCA embedding f=of permutaion matrices for high performing models in two settings: Left: trianing on the full $6\times 6$ matrix multiplication (OP). Right: trianing on a mixture of OP, and subtasks OP\_BOTTOM (multiplying bottom $3\times3$ block, repeating the top block of first matrix), and and OP\_TOP (multiply top $3\times3$ block, repeat bottom of first matrix). 
    The mixed task model reaches $>90\%$ accuracy at 1.1k parameters, whereas the single OP model requires 1.9k for the same accuracy level.  }
    \label{fig:op_pca}
    \label{fig:op_all_pca}
\end{figure}

\clearpage

\section{Shuffled SUM}
\label{shuffled_sum}
The Shuffled SUM table was constructed using the following code:

\lstset{
  language=Python,
  basicstyle=\ttfamily\footnotesize,
  keywordstyle=\color{blue},
  commentstyle=\color{gray},
  stringstyle=\color{orange},
  showstringspaces=false,
  breaklines=true,
  frame=single,
  rulecolor=\color{black!30},
  columns=flexible
}

\begin{lstlisting}[caption={Generate upper triangle and diagonal matrices}]
# Generate upper triangle and diagonal matrices.
# To ensure commutativity i+j=j+i, we will transpose the
# the upper triangle.
upper_triangle_matrix = {(i, j): (i + j) % MOD
    for i in range(MOD)
    for j in range(MOD) if i < j}

# To ensure uniform distribution after shuffling the values,
# we must shuffle the diagonal separately.
# This is because all off-diag r.h.s. are repeated twice
# for commutativity, but not the diagonal entries.
diagonal_matrix = {(i, i): (2 * i) % MOD for i in range(MOD)}

# Function to shuffle values in a dictionary
def shuffle_dict_values(d):
    keys = list(d.keys())
    values = list(d.values())
    random.shuffle(values)
    return dict(zip(keys, values))

# Apply shuffling
shuffled_triangle = shuffle_dict_values(upper_triangle_matrix)
shuffled_diagonal = shuffle_dict_values(diagonal_matrix)
\end{lstlisting}

As a result, there is no consistent mapping between the original and shuffled values (e.g., the number 1 does not always map to 2), making it difficult for the model to learn a deterministic transformation. While the shuffled sum table remains commutative, we did not enforce associativity. A direct check also confirmed that it does not satisfy the associative property for most triplets.

\begin{figure}[!htbp]
    \centering
    \includegraphics[width=.4\linewidth]{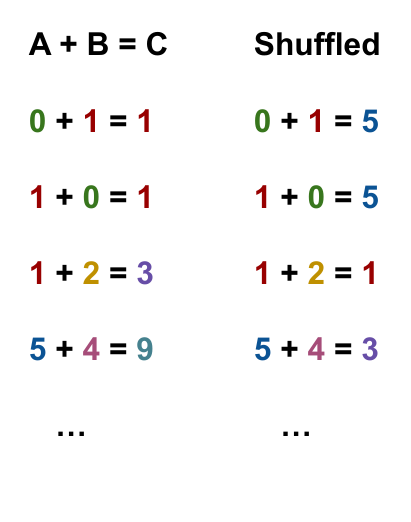}
    \caption{
    \textbf{Shuffle the summation table.} }
    \label{fig:shuffle_sum_table_example}
\end{figure}
\clearpage

\subsection{Shuffled SUM modulo 10}
\begin{figure}[!htpb]
    \centering
    \includegraphics[height=0.25\linewidth]{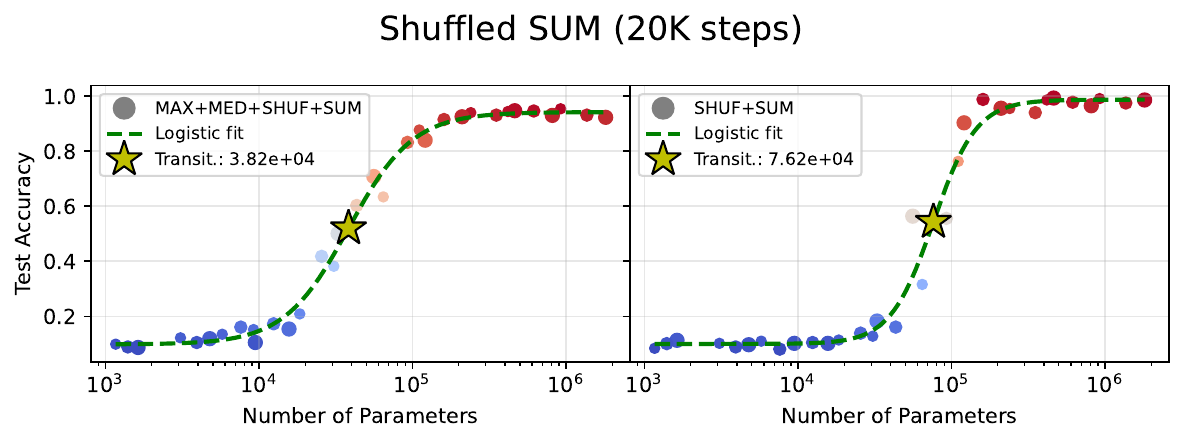}
    \includegraphics[height=0.25\linewidth]{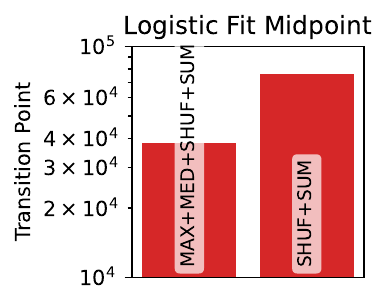}
    \caption{\textbf{Shuffled symmetric sum table, Mod 10:} We find that this shuffled version of sum is again more difficult to learn than any of the math operations except for the original Pure SUM, which remains slightly more difficult than even the shuffled version. 
    Additionally, we observe that MAX+MED+Shuffled SUM is again more difficult than all operations except pure sum, suggesting that the number properties played an important role in the other tasks becoming easier in mixed training. 
    }
    \label{fig:transition-shuffled-sum-10}
\end{figure}

\begin{figure}[htpb]
    \centering
    \includegraphics[width=0.49\linewidth]{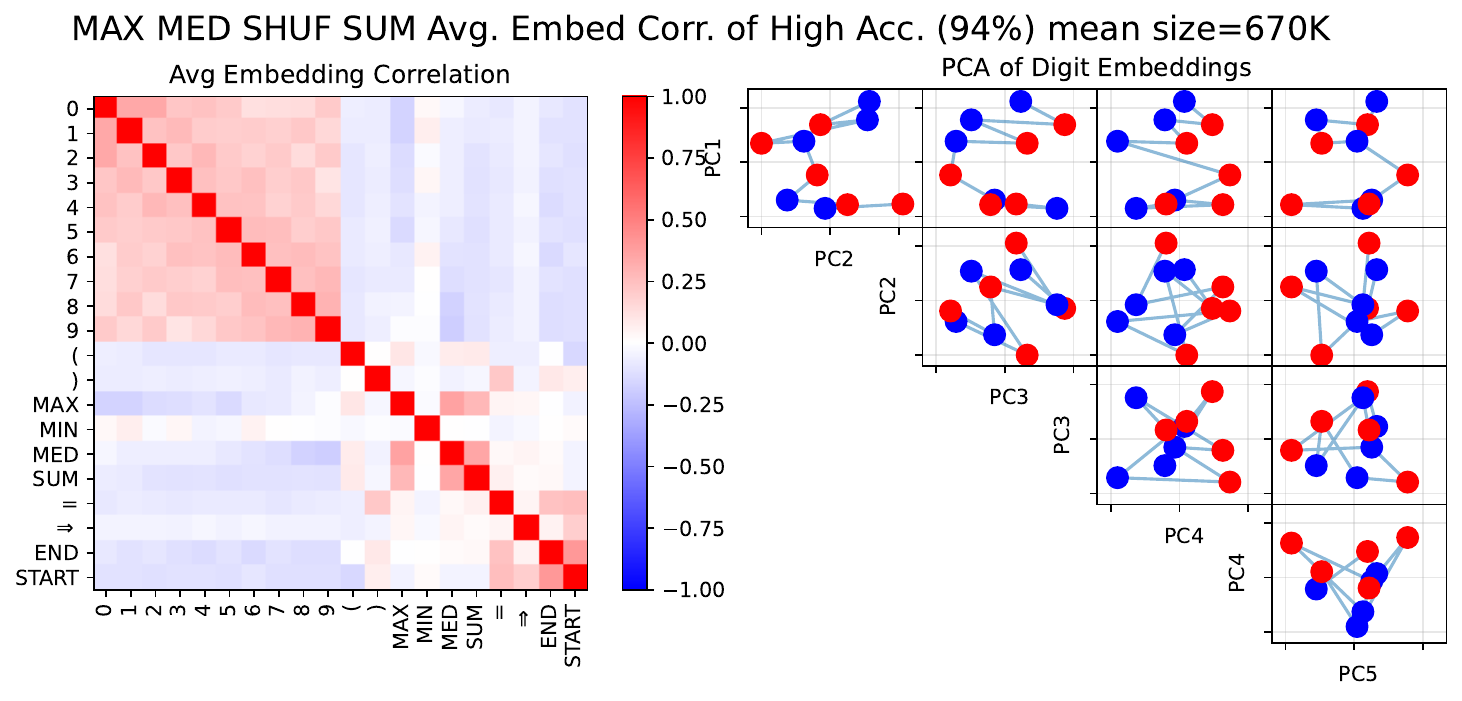}
    \includegraphics[width=0.49\linewidth]{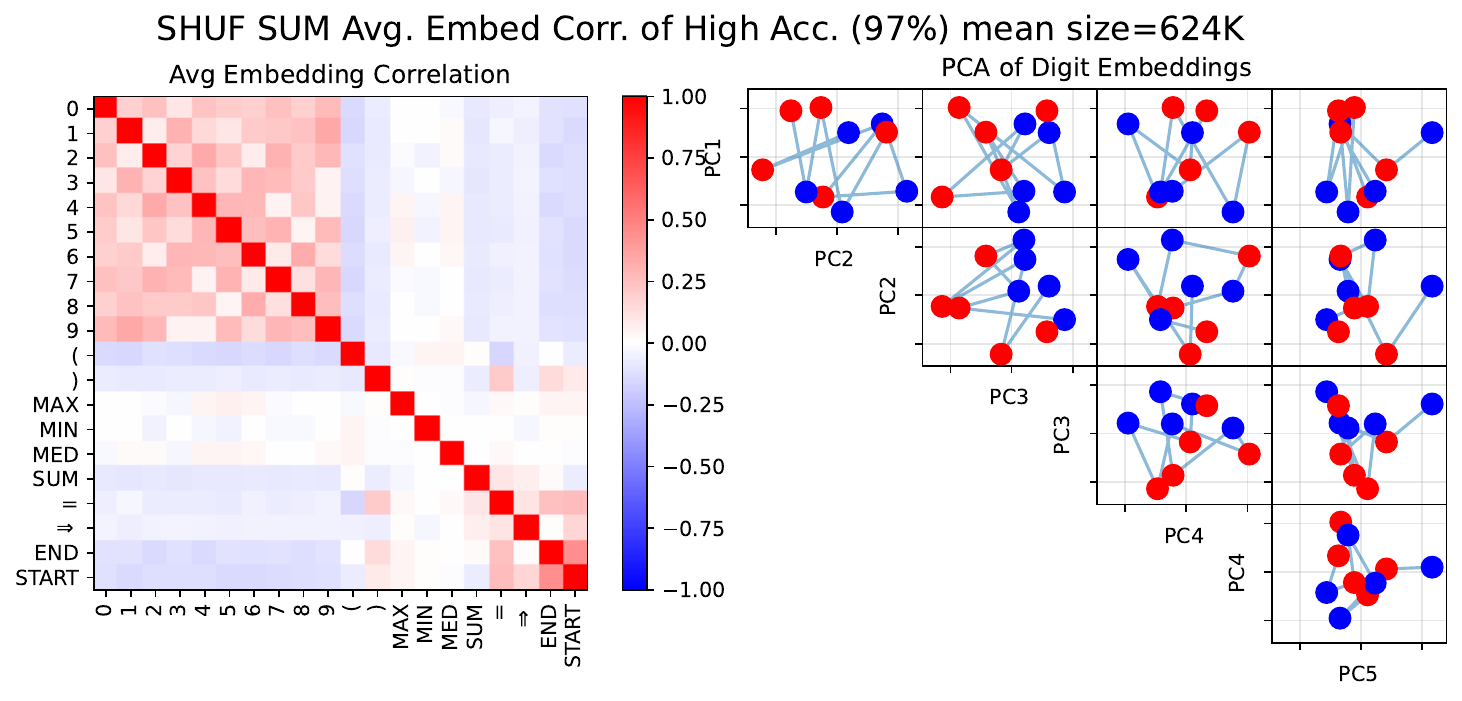}
    \caption{\textbf{PCA of embeddings mod 10 for Shuffled symmetric SUM, pure vs mixed with MAX and MED:}
    The numbers are colored based on parity (odd is red, even is blue). 
    It is not expected that the embeddings show strong signals. 
    There seems to be partial ordering of the numbers, but the clear wave patterns and clear parity separation is not evident. 
    }
    \label{fig:corr-PCA-shuffle-10}
\end{figure}

\subsection{Shuffled SUM modulo 26}
We make a symmetric sum table ($A+B=B+A$), with a randomly shuffled right hand side, meaning where if in the table $A+B=C$, $C$ does not the actual arithmetic sum of $A+B$ modulo 26. 
We do this to test a couple of hypotheses:
\begin{enumerate}
    \item \textbf{Hypothesis 1:} The SUM trained alone is not really learning the logic arithmetic of numbers, but rather memorizing the sum table. 
    \item \textbf{Hypothesis 2:} Joint training with max and med leads to learning number properties. 
\end{enumerate}

If the first hypothesis holds, the shuffled sum would also require similarly high number of parameters as the normal sum and show similarly random patterns in the embedding space. 
If the second hypothesis holds, then joint training of shuffled sum should actually have detrimental effects on learning sum because number properties don't play a role in learning shuffled sum. 
Thus we expect the jointly trained model to struggle to learn all three operations, or require significantly higher number of parameters to master shuffled sum, compared to joint training on regular sum. 

%
\begin{figure}[htpb]
    \centering
    \includegraphics[width=0.72\linewidth]{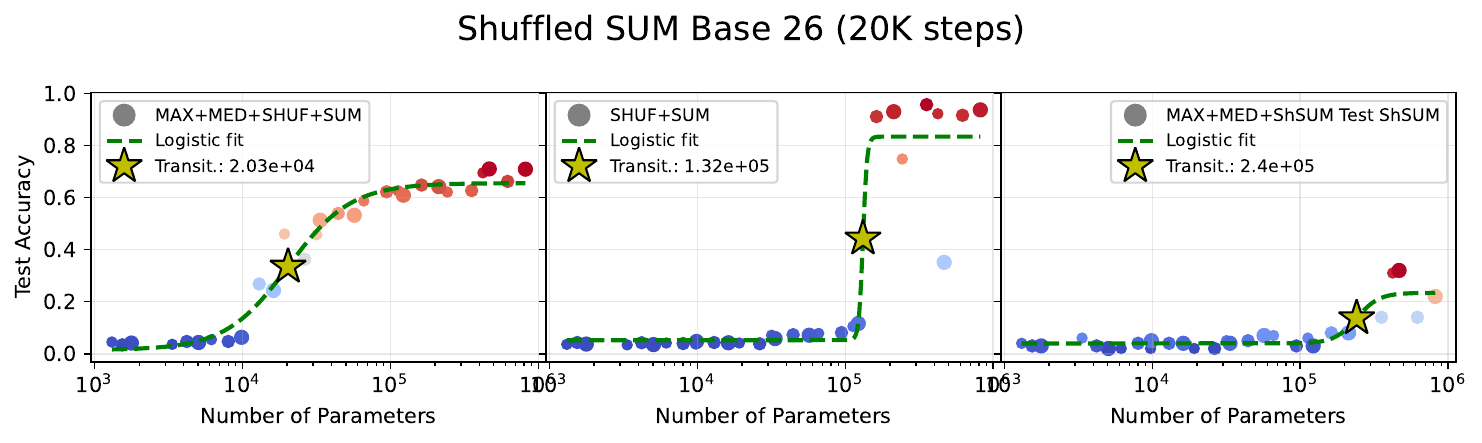}
    \includegraphics[width=0.27\linewidth]{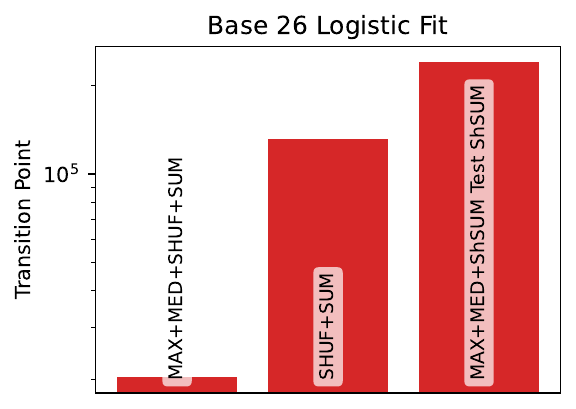}
    \caption{\textbf{Shuffled symmetric sum table, Mod 26, 20k steps:} We find that this shuffled version of sum is again more difficult to learn than any of the math operations except for the original Pure SUM, which remains slightly more difficult than even the shuffled version. 
    We also observe that MAX+MED+Shuffled SUM never reaches more than 80\% accuracy. 
    The third scatter plot from the let shows the accuracy of the MAX+MED+Shuffled SUM model on the Shuffled SUM test set. 
    We see that the accuracy is very low ($\sim 20\%$ top), showing that the mixed model never learned the shuffled SUM. 
    This may suggest that MAX+MED revealed number properties, but Shuffled SUM was incompatible with those properties, leading to a model that overall cannot solve the two problems (MAX+MED and Shuffled SUM) simultaneously. 
    }
    \label{fig:transition-shuffled-sum-26}
\end{figure}

\begin{figure}[htpb]
    \centering
    \includegraphics[width=0.72\linewidth]{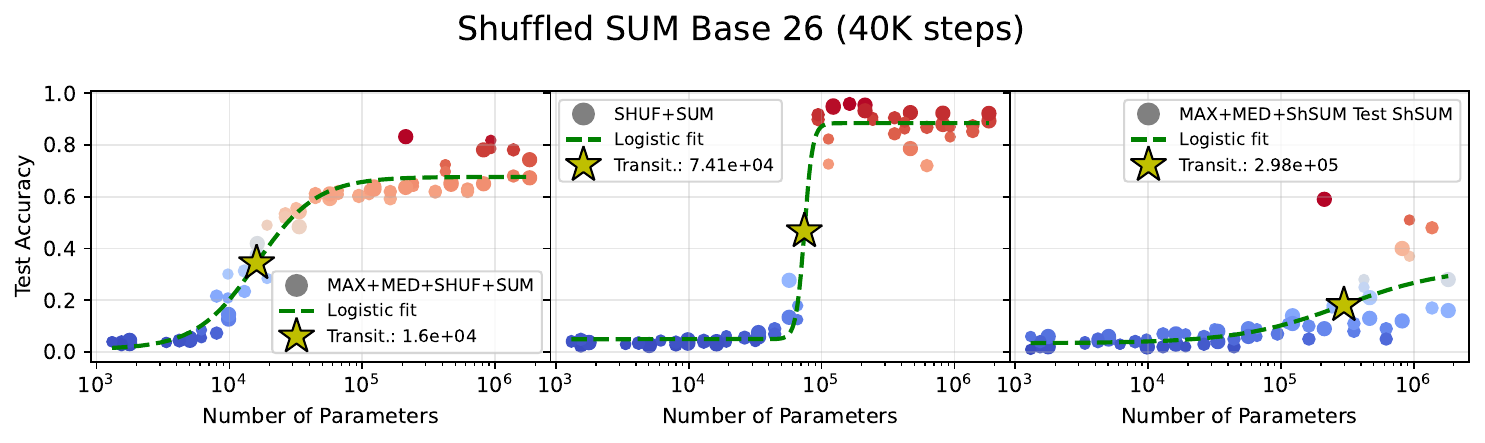}
    \includegraphics[width=0.27\linewidth]{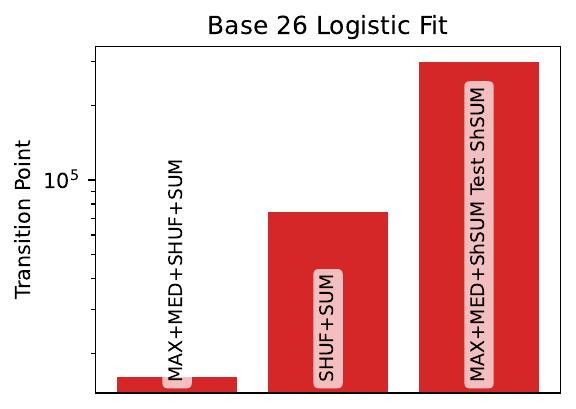}
    \caption{\textbf{Shuffled symmetric sum table, Mod 26, 40 steps:} We find that this shuffled version of sum is again more difficult to learn than any of the math operations except for the original Pure SUM, which remains slightly more difficult than even the shuffled version. 
    We also observe that MAX+MED+Shuffled SUM never reaches more than 80\% accuracy. 
    The third scatter plot from the let shows the accuracy of the MAX+MED+Shuffled SUM model on the Shuffled SUM test set. 
    We see that the accuracy is very low ($\sim 20\%$ top), showing that the mixed model never learned the shuffled SUM. 
    This may suggest that MAX+MED revealed number properties, but Shuffled SUM was incompatible with those properties, leading to a model that overall cannot solve the two problems (MAX+MED and Shuffled SUM) simultaneously. 
    }
    \label{fig:transition-shuffled-sum-26-40k}
\end{figure}

\begin{figure}[htpb]
    \centering
    \includegraphics[width=0.49\linewidth]{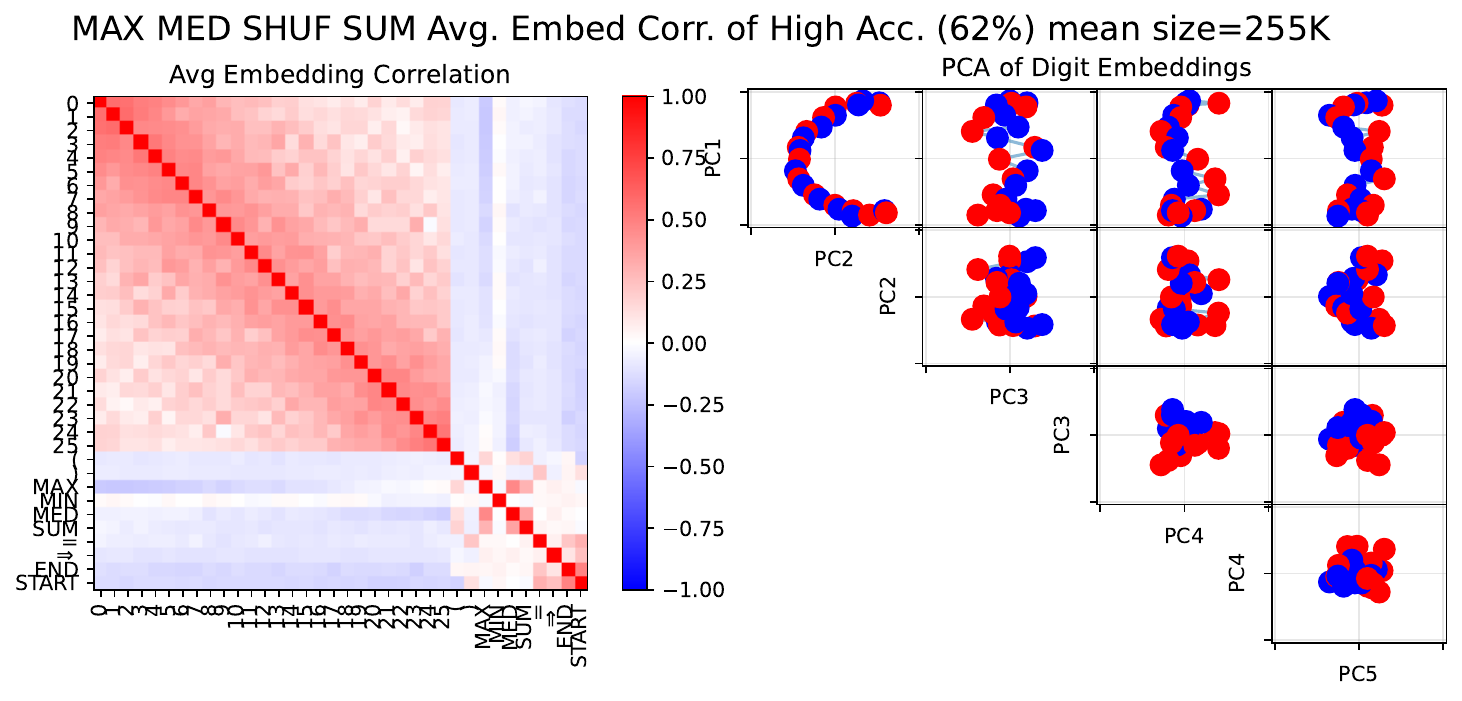}
    \includegraphics[width=0.49\linewidth]{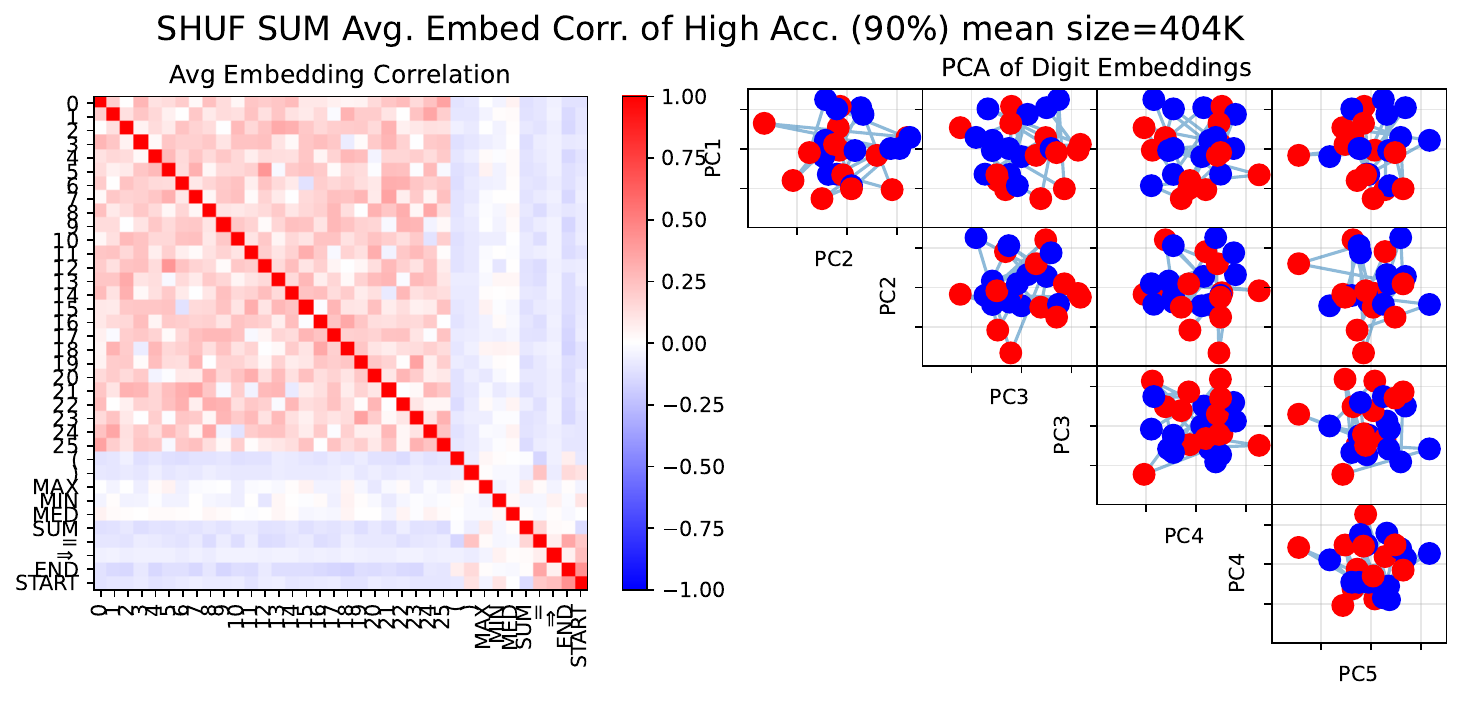}
    \caption{\textbf{PCA of embeddings for Shuffled symmetric SUM, pure vs mixed with MAX and MED, Mod 26:}
    The numbers are colored based on parity (odd is red, even is blue). 
    There seems to be partial ordering of the numbers, but the clear wave patterns are not visible. 
    We do almost observe parity separation in PC3 and PC4, albeit with some noise.
    It is curious that the system still learns partial parity, but evidently this feature did not allow the system to learn the shuffled SUM perfectly. 
    \nd{Was the Shuffled SUM table the same for the two? Can we test All3 on Shuffled SUM data?}
    }
    \label{fig:corr-PCA-shuffle-26}
\end{figure}

\clearpage
\section{Deep transformer vs. Recurrent Transformer}

\begin{figure}[!htbp]
    \centering
    \includegraphics[width=.5\linewidth]{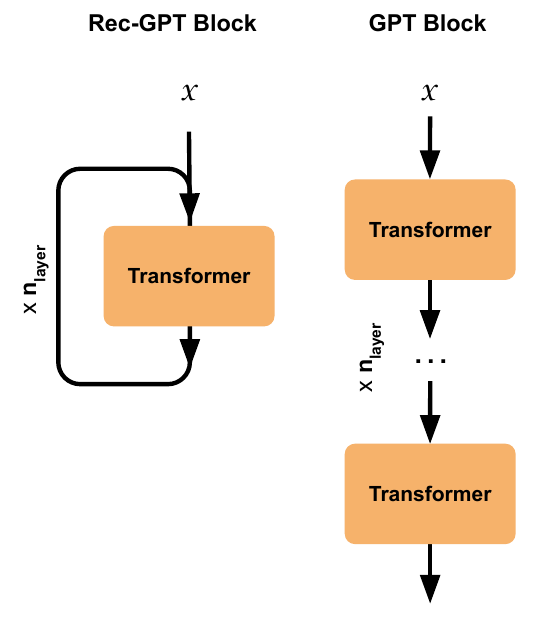}
    \caption{
    \textbf{Recurrent Transformer vs. Deep transformer} In the Rec-GPT block, there is only one transformer block that is applied recursively \textit{n} times, so each output becomes the input for the next step. In contrast, in a traditional GPT block, there are \textit{n} transformer blocks stacked sequentially on top of each other. }
    \label{fig:rec-gpt_sketch}
\end{figure}

\section{Results using Deep Transformer based models}
\label{resutls_deep_transformer}
\subsection{Experiment settings}
\begin{enumerate}
    \item \textbf{Model}: GPT model (\cite{karpathy2022nanogpt})
    \item \textbf{Number of layers}: 1, 2, 3, 4, 5, 6
    \item \textbf{Number of head}: 1
    \item \textbf{Embedding dimension}: 4, 8, 12, 16, 24, 32, 48, 64, 96, 128, 192, 256
    \item \textbf{Number of random seed}: 5
    \item \textbf{Context window size}: 128
    \item \textbf{Batch size}: 64
    \item \textbf{Optimizer}: Adam
    \item \textbf{Minimum learning rate}: 1e-4
    \item \textbf{Maximum interation steps}: 20000/50000
    \item \textbf{Early Stopping criteria}: Early stopping was applied after 2000 iterations when the change in training loss was below $\Delta_{min}$ = 2.5e-4 for 10 consecutive evaluation steps.
    \item \textbf{Data} We generate 50,000 initial equations and split them into training and test sets by excluding 100 randomly selected triplets. The training set consists of approximately 45,000 examples that do not contain any of the excluded triplets. The test set comprises around 2,000 examples, carefully curated to ensure that the excluded triplets do not appear, even in the final step of each equation. We evaluate model performance on a subset of 1,000 final test examples.
    \item \textbf{Vocab - Base 10 ListOps}: \verb|%()+-/0123456789=>es|
    \item \textbf{Vocab - Base 26 ListOps}: \verb|ABCDEFGHIJKLMNOPQRSTUVWXYZse()+-/%=>|
    \item \textbf{Hardware}: All simulations were run on a mix of GPUs, including NVIDIA A100, H200, RTX 4090, and A30, as well as on a standard modern laptop. The experiments are lightweight, requiring approximately 1.5GB of RAM, and can be executed efficiently on any recent laptop-class device. 
\end{enumerate}

\subsection{Experiments on ListOps.}
\begin{enumerate}
    \item \textbf{Base 10}: - no CoT/CoT: MAX, MED, SUM, MAX/MED/SUM
    \item \textbf{Base 10}: - all combination: MAX, MIN, MED, SUM + MAX/MED/SUM-RANDOM SHUFFLED
    \item \textbf{Base 26}: - all combination: MAX, MIN, MED, SUM + MAX/MED/SUM-RANDOM SHUFFLED
    \item \textbf{Triplet}: MAX+MED+SUM 2700 (2700 training data samples = 900 MAX + 900 MED + 900 SUM),  MAX+MED+SUM 900 (900 training data samples = 300 MAX + 300 MED + 300 SUM), SUM (900 training data samples)
    
\end{enumerate}

\subsection{Data and processing}\label{ap:data}
\subsubsection{Dataset Notation} \label{ap:listops-notation}
ListOps consists of nested mathematical equations involving operations such as min, max, median, and sum modulo 10 applied to single-digit numbers (0-9). 
It uses the Polish notation: (operation, [inputs])
For example: \verb|max(3,min(7,4,9))=4|.
\begin{quote}
    \verb|max(3,min(7,4,9))=4| $\quad \Rightarrow\quad $ \textbf{Polish: } \verb|(max,3,(min,7,4,9))=4|
\end{quote}
To disentangle any complexity arising from tokenization we further simplify these expression by representing the by symbols: '$+$' for max, '$-$' for min, '$/$' for median, and '$\%$' for sum modulo 10.
For example:
\begin{quote}
    \verb|(max,3,(min,7,4,9))=4| $\quad \Rightarrow\quad $ \textbf{Our notation: } \verb|s(+3(-749))=4e|
\end{quote}
In this notation, 's' denotes the start of the expression, 'e' marks the end, and parentheses indicate nesting levels. 

\subsubsection{Tokenization}\label{ap:token}
We employ word-level tokenization where each semantically meaningful unit is a single token:
\begin{itemize}
    \item \textbf{Operations:} Each operation with its modulus is one token (e.g., \verb|add_20|, \verb|max_10|, \verb|prod_19|).
    \item \textbf{Numbers:} Each number is one token, regardless of digit count (e.g., \verb|3|, \verb|13|, \verb|104| are each single tokens).
    \item \textbf{Symbols:} Parentheses \verb|(|, \verb|)|, equals \verb|=|, CoT separator \verb|>|, and end token \verb|<eos>| are each single tokens.
\end{itemize}
Note that commas are not needed as token boundaries are implicit.
For modulo $n$, the vocabulary size is approximately $n$ + (number of operations) + 5 (for symbols).
This tokenization ensures that each number is atomic, preventing the model from learning spurious digit-level patterns.

\subsubsection{Chain of Thought Implementation} \label{ap:cot}
\begin{figure}[t]
    \centering
    \includegraphics[width=1\linewidth]{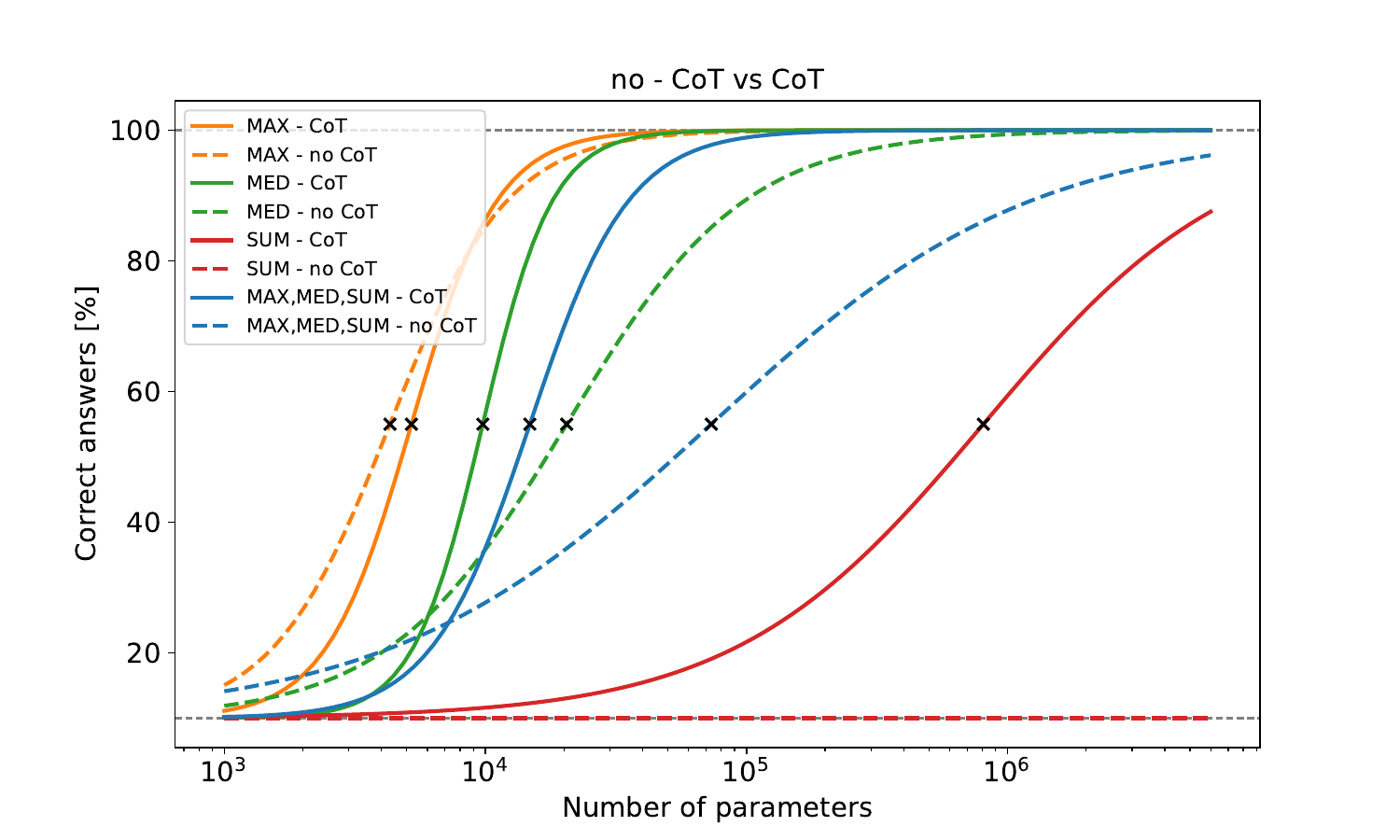}
    \caption{\textbf{no - CoT vs. CoT.} 
    Providing solutions as chain-of-thought (CoT) helps models learn the tasks. In almost all cases, CoT accelerates learning, enabling smaller models to succeed. This effect is especially strong for the sum operation, which cannot be learned without CoT. 
    \nd{May remake fig without all3 and showing faint data points, also fit  with discrete transition.
    "nosol" experiments are for no-CoT.} 
    }
    \label{fig:cot}
\end{figure}

Directly solving nested ListOps in one step is challenging for transformers (Fig.~\ref{fig:cot}).
Even with maximum three nesting levels and three operands, GPT models with over 10 million parameters fail to learn.
To address this, we use a chain-of-thought (CoT) approach that resolves the innermost operation at each step.

For example, with the functional notation:
\begin{quote}
    \verb|add_10(1 2 add_10(3 4))>add_10(1 2 7)>0=0<eos>|
\end{quote}
The CoT proceeds as:
\begin{itemize}
    \item Initial expression: \verb|add_10(1 2 add_10(3 4))|
    \item Step 1: Resolve innermost \verb|add_10(3 4)| $\to$ \verb|7|
    \item Step 2: Resolve \verb|add_10(1 2 7)| $\to$ \verb|0|
    \item Final: \verb|0=0<eos>|
\end{itemize}

This CoT approach guides the model through step-by-step reasoning, provides granular supervision, and significantly improves performance on challenging operations (Fig.~\ref{fig:cot}).

\subsection{Modulo 26 \label{ap:mod26} }

We define the following token vocabulary \verb|ABCDEFGHIJKLMNOPQRSTUVWXYZse()+-/%=>|, where letters are mapped to integers such that \texttt{A}~$\rightarrow$~0, \texttt{B}~$\rightarrow$~1, ..., \texttt{Z}~$\rightarrow$~25. 

\begin{figure}[h]
    \centering
    \includegraphics[width=0.49\linewidth]{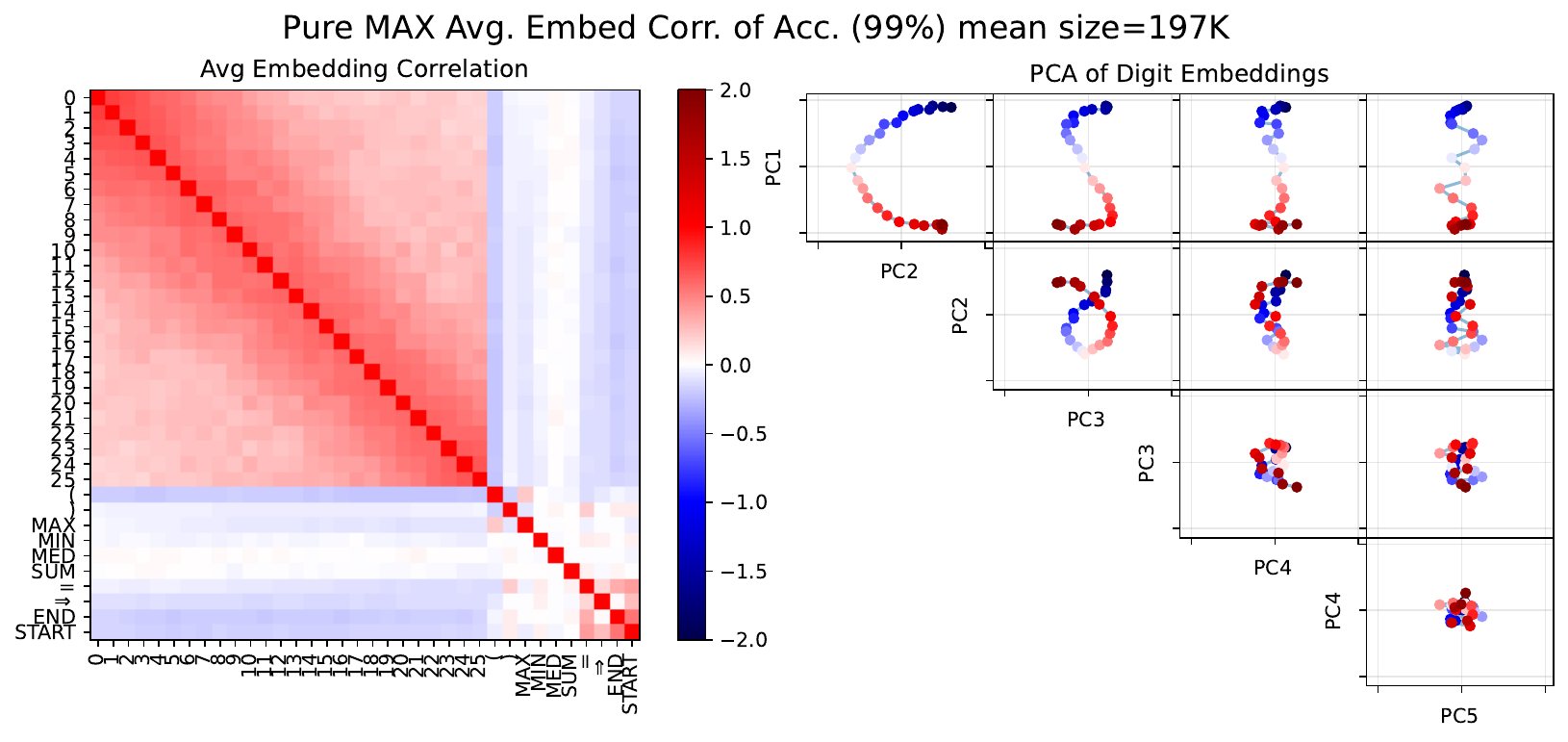}
    \includegraphics[width=0.49\linewidth]{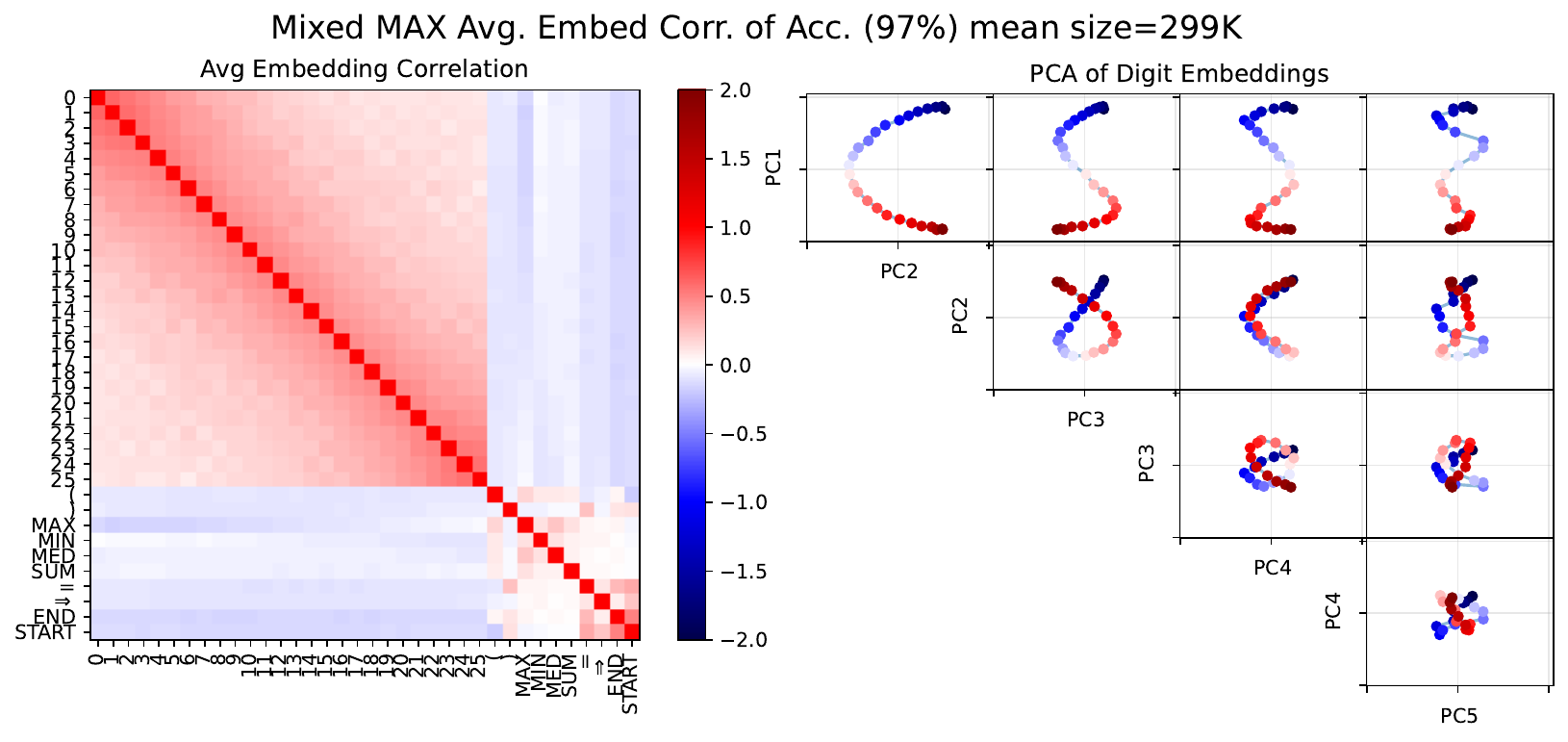}
    \includegraphics[width=0.49\linewidth]{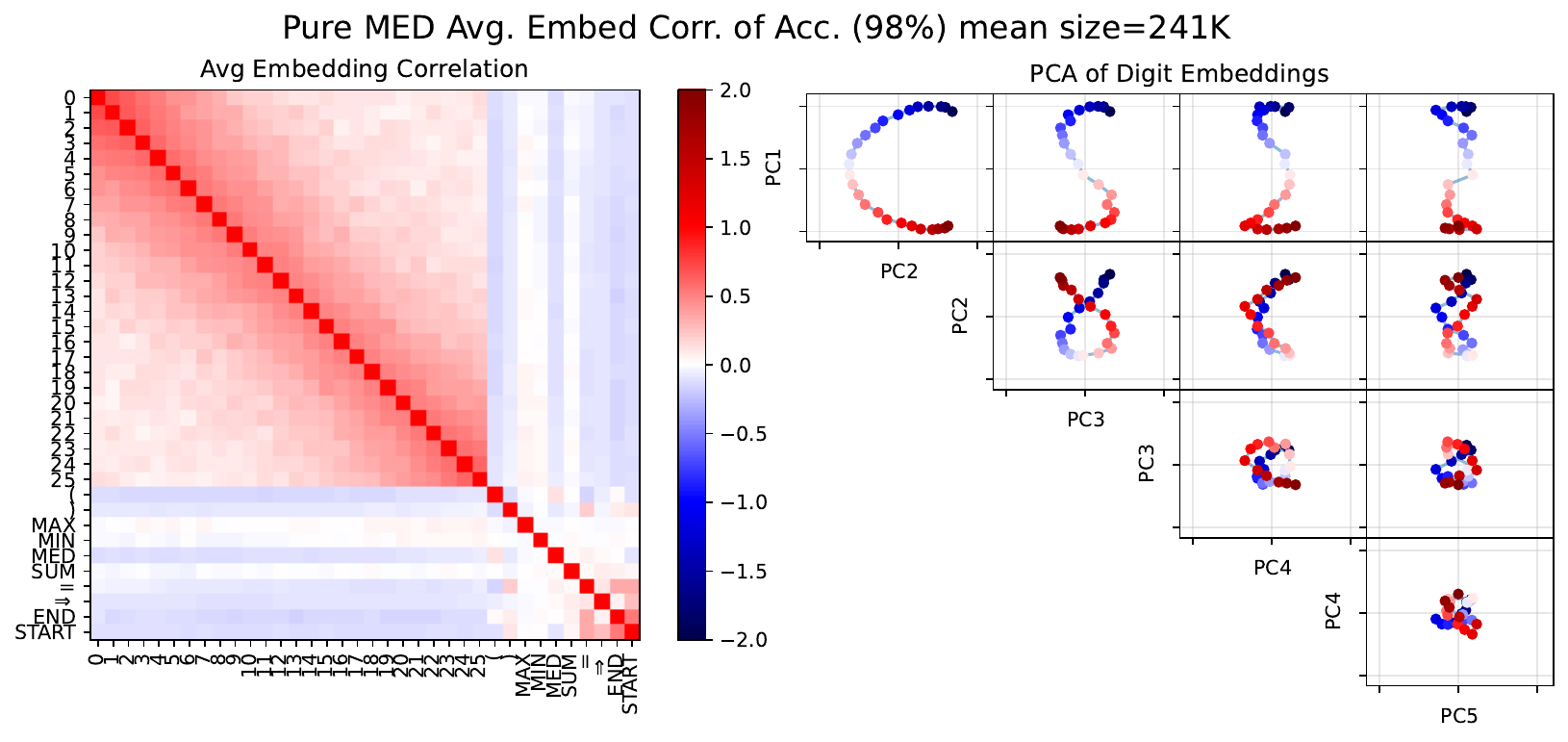}
    \includegraphics[width=0.49\linewidth]{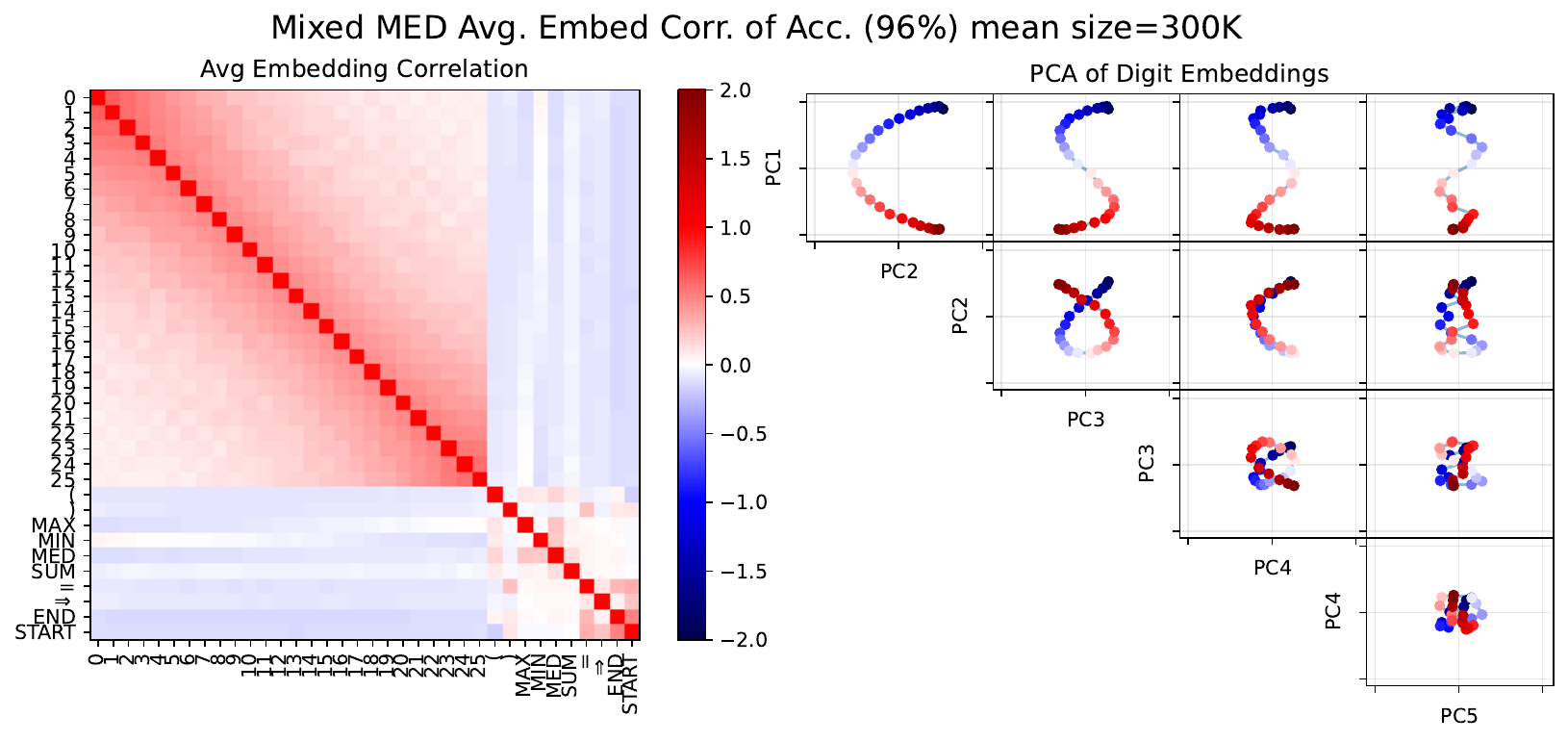}
    \includegraphics[width=0.49\linewidth]{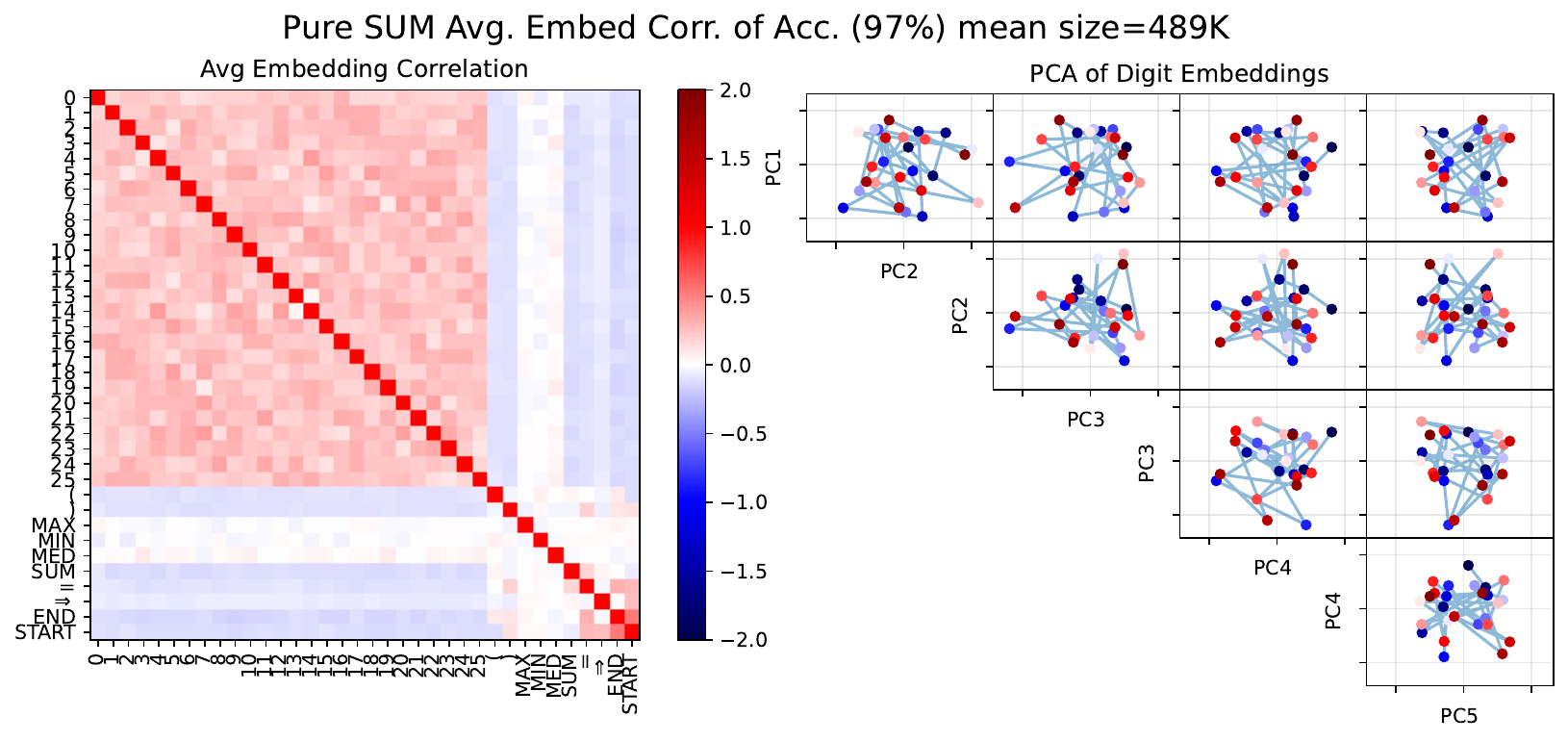}
    \includegraphics[width=0.49\linewidth]{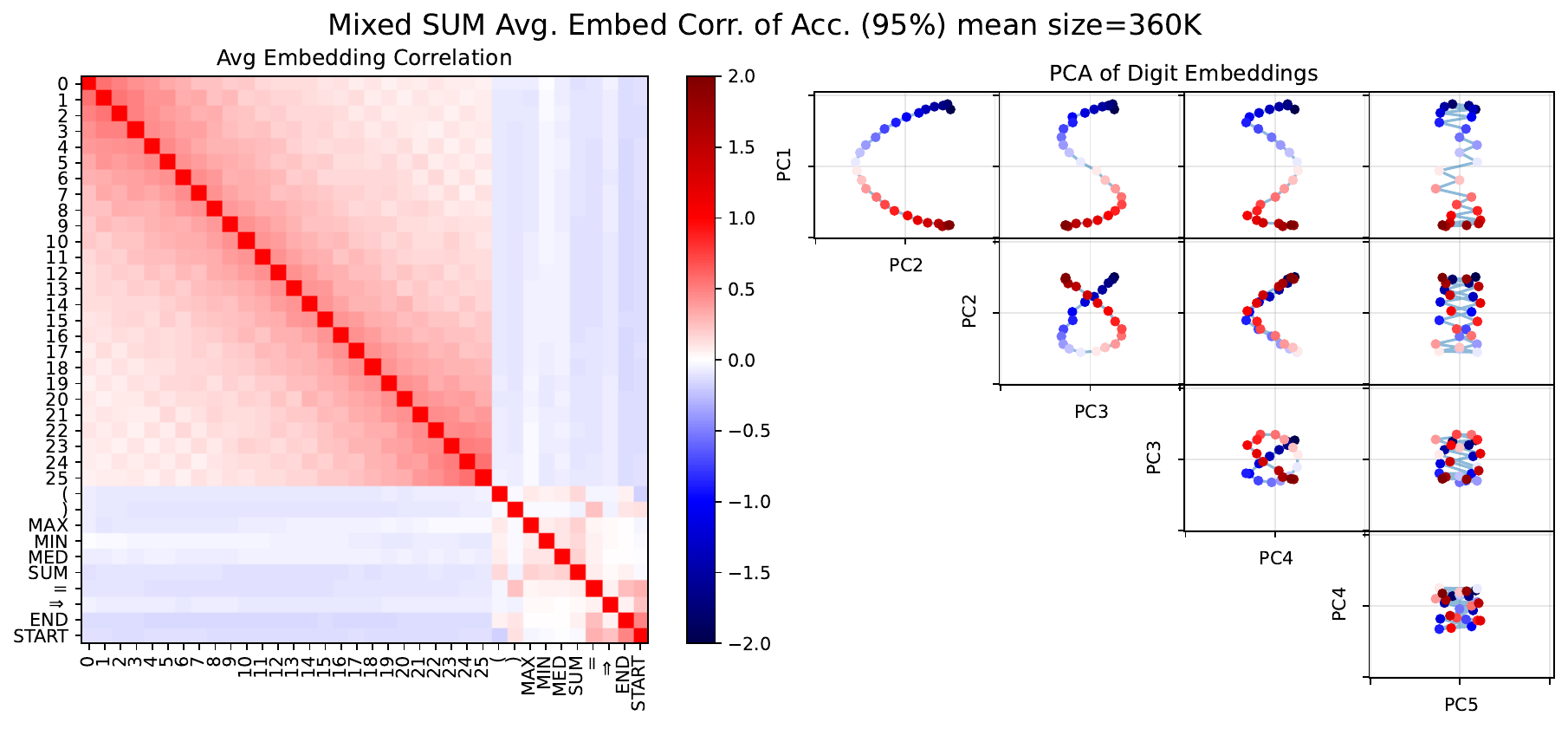}
    \caption{
    \textbf{PCA of embeddings:}
    We choose all models which reached over $95$\% test accuracy. 
    Each row shows the average correlation matrix and top PCs for models trained on either a single operation, e.g. Pure+MAX, or all mixtures involving a given operation, e.g. Mixed+MAX. 
    Again, pure SUM does not show a discernible structure in the embeddings, whereas all cases do. 
    }
    \label{fig:PCA-pure-mixed-26-all}
\end{figure}

\begin{figure}[h]
    \centering
    \includegraphics[width=0.49\linewidth]{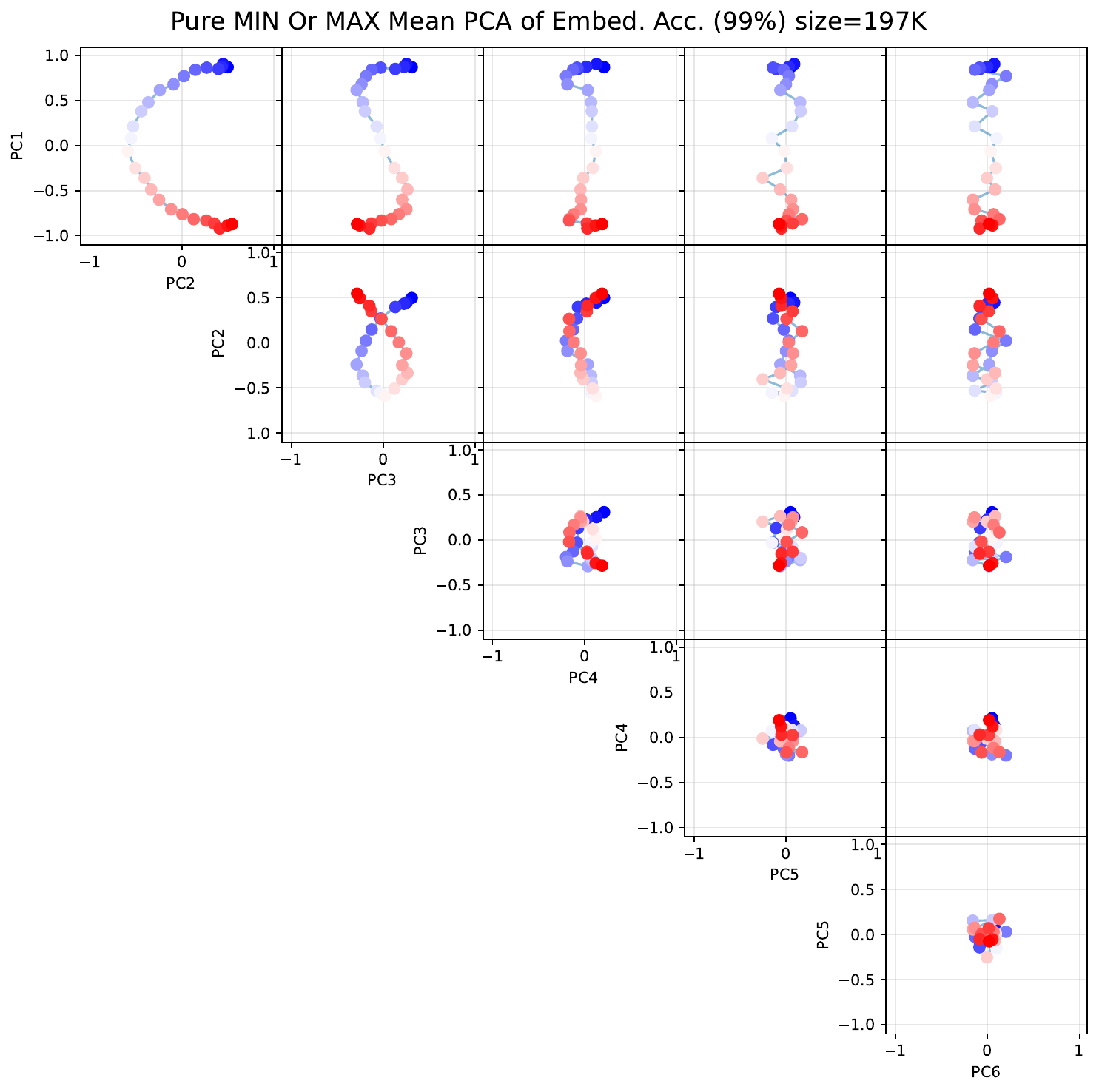}
    \includegraphics[width=0.49\linewidth]{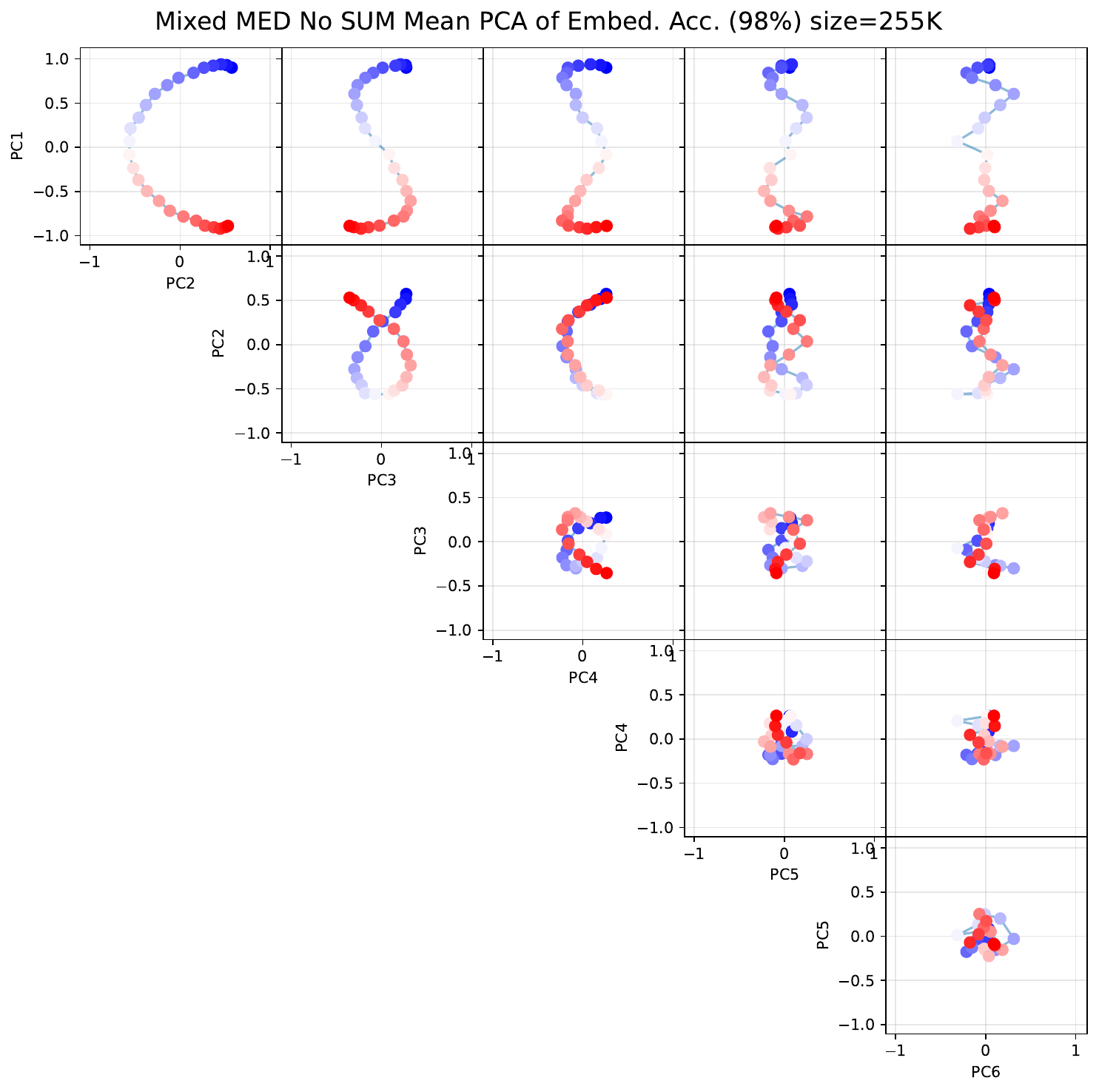}
    \includegraphics[width=0.49\linewidth]{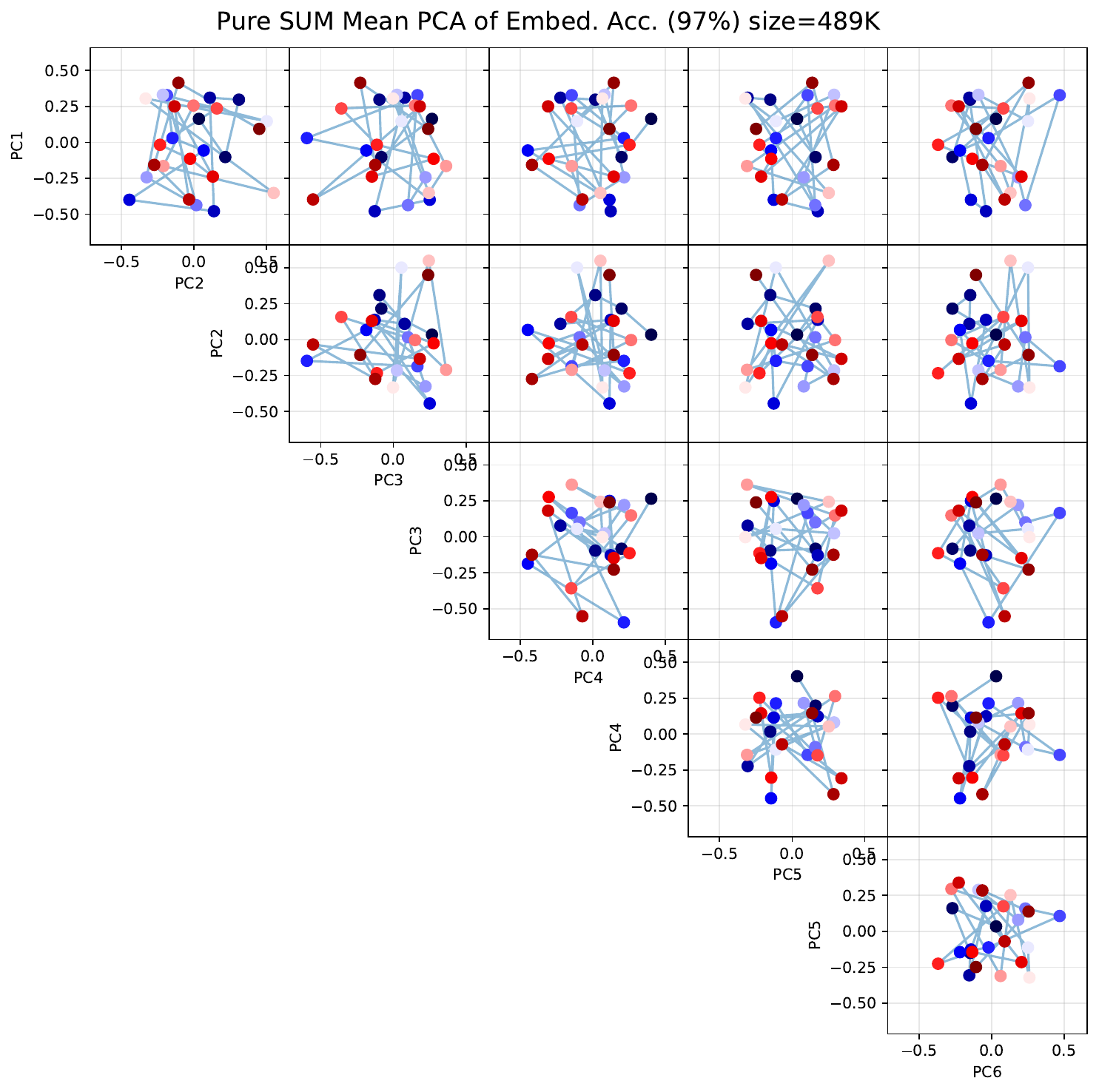}
    \includegraphics[width=0.49\linewidth]{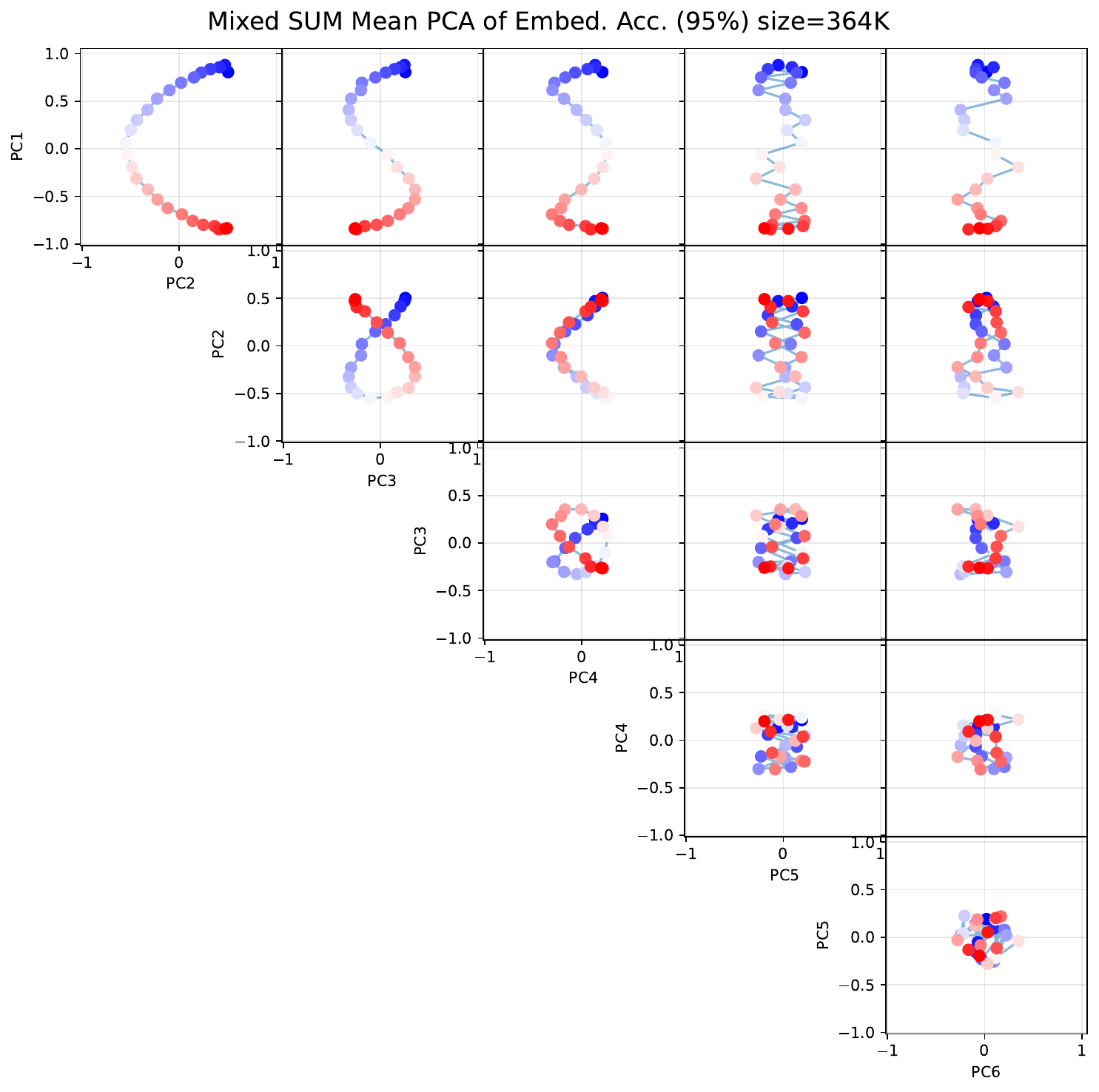}
    \caption{
    \textbf{PCA of embeddings:}
    We choose all models which reached over $95$\% test accuracy. 
    Each row shows the average correlation matrix and top PCs for models trained on either a single operation, e.g. Pure+MAX, or all mixtures involving a given operation, e.g. Mixed+MAX. 
    Again, pure SUM does not show a discernible structure in the embeddings, whereas all cases do. 
    }
    \label{fig:PCA-pure-mixed-26}
\end{figure}

\begin{figure}
    \centering
    \includegraphics[width=0.49\linewidth]{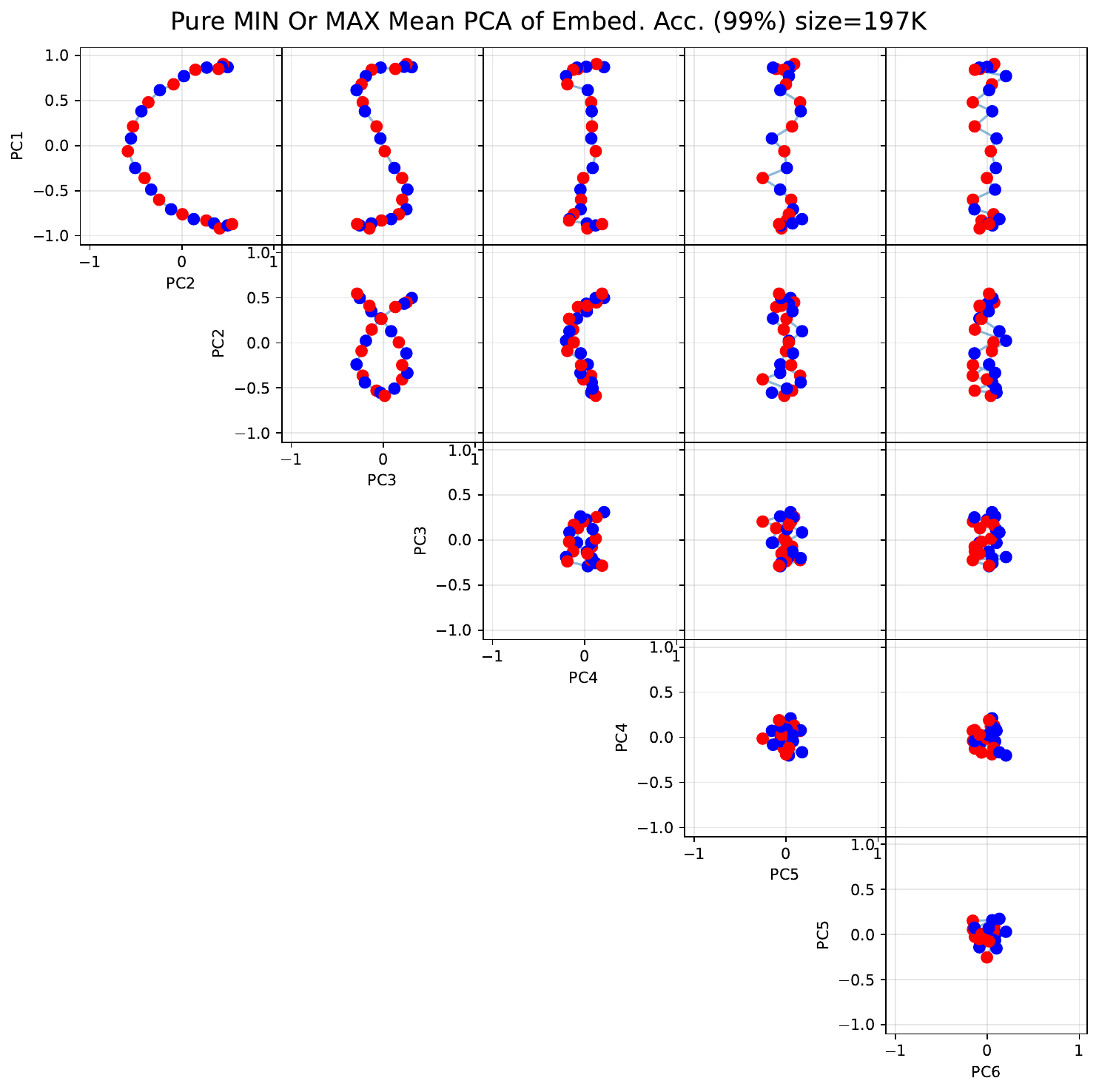}
    \includegraphics[width=0.49\linewidth]{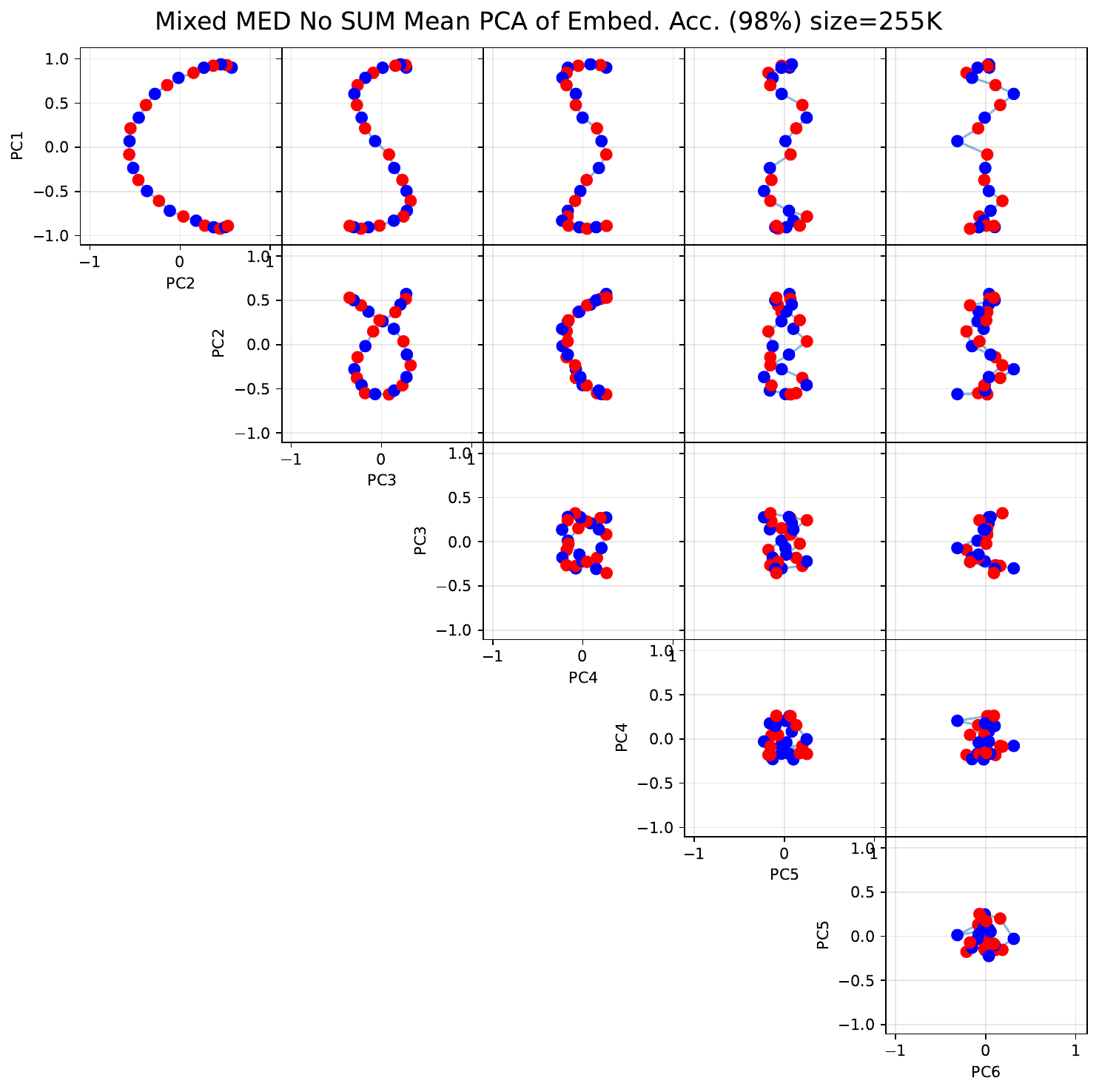}
    \includegraphics[width=0.49\linewidth]{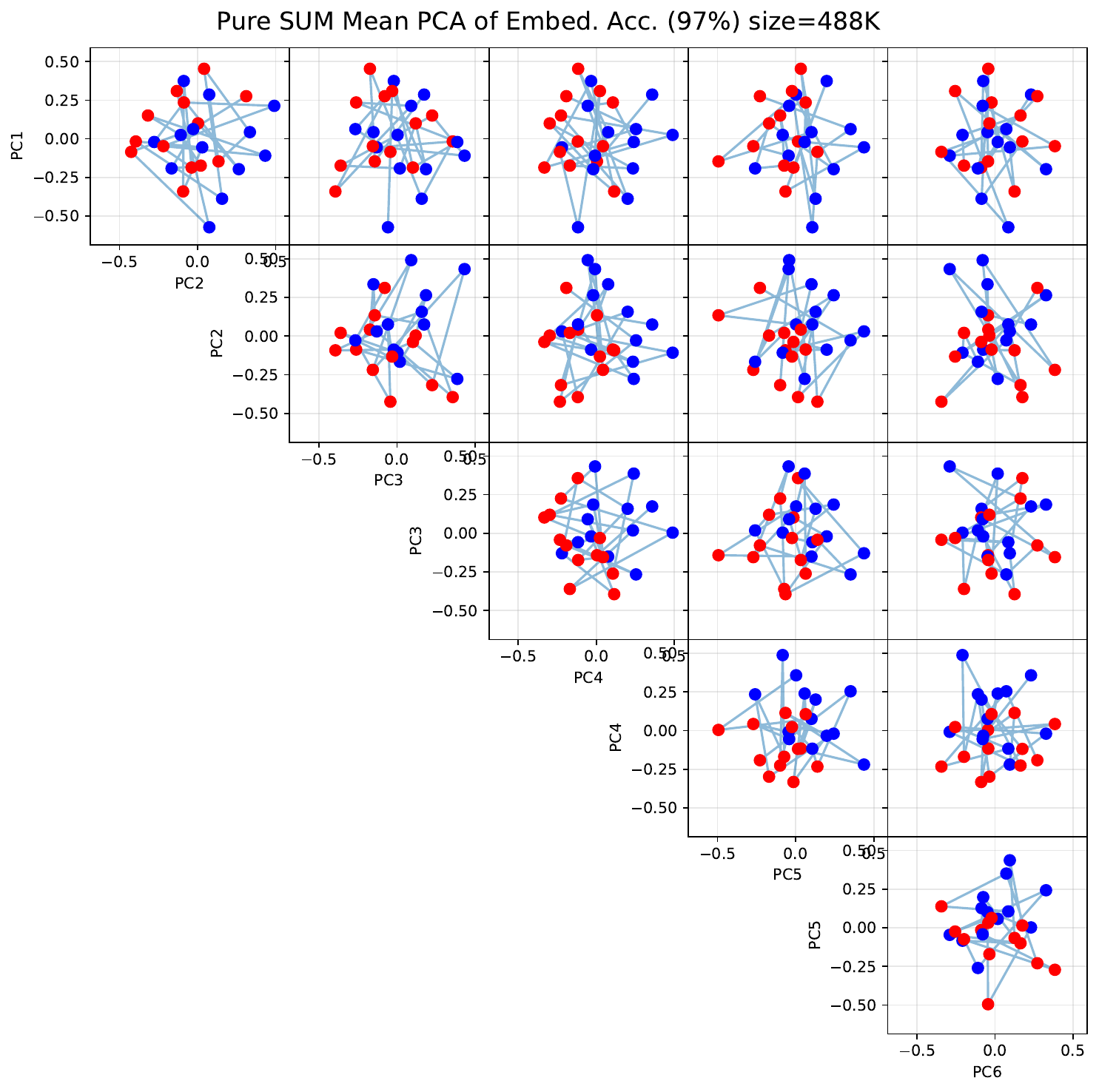}
    \includegraphics[width=0.49\linewidth]{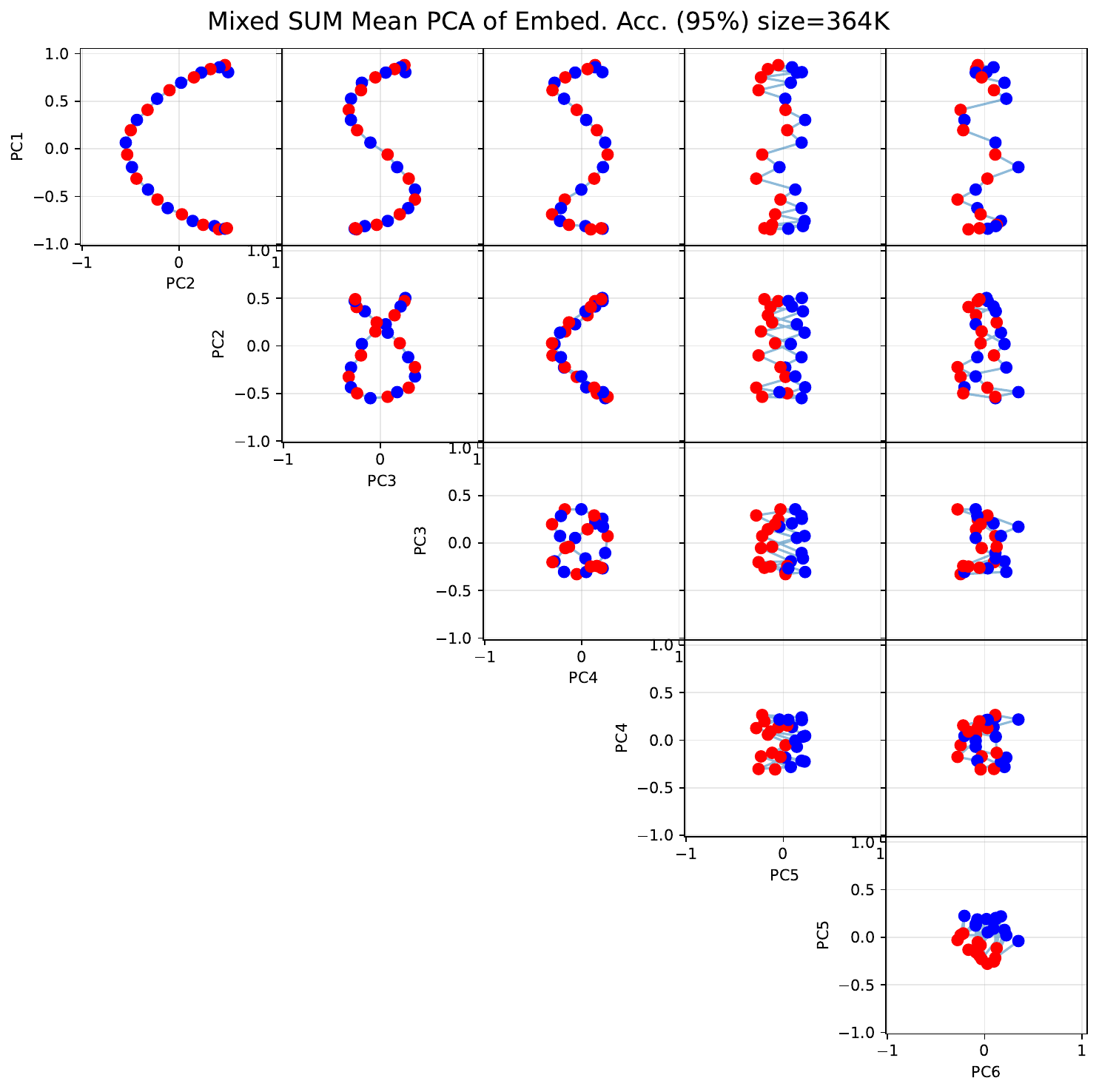}
    \caption{
    \textbf{PCA of embeddings Odd vs Even:}
    Same plot as above, only odd numbers colored red and even colored blue. 
    Mixed SUM shows a clear odd-even separation in a few of the top PCs. 
    Such a separation is not clearly observed in other cases.  
    Interestingly, Pure SUM approximately separates odd-even, suggesting such separation may play a role in its algorithm. 
    }
    \label{fig:PCA-pure-mixed-26-oddeven}
\end{figure}

\begin{figure}
    \centering
    \includegraphics[width=0.49\linewidth]{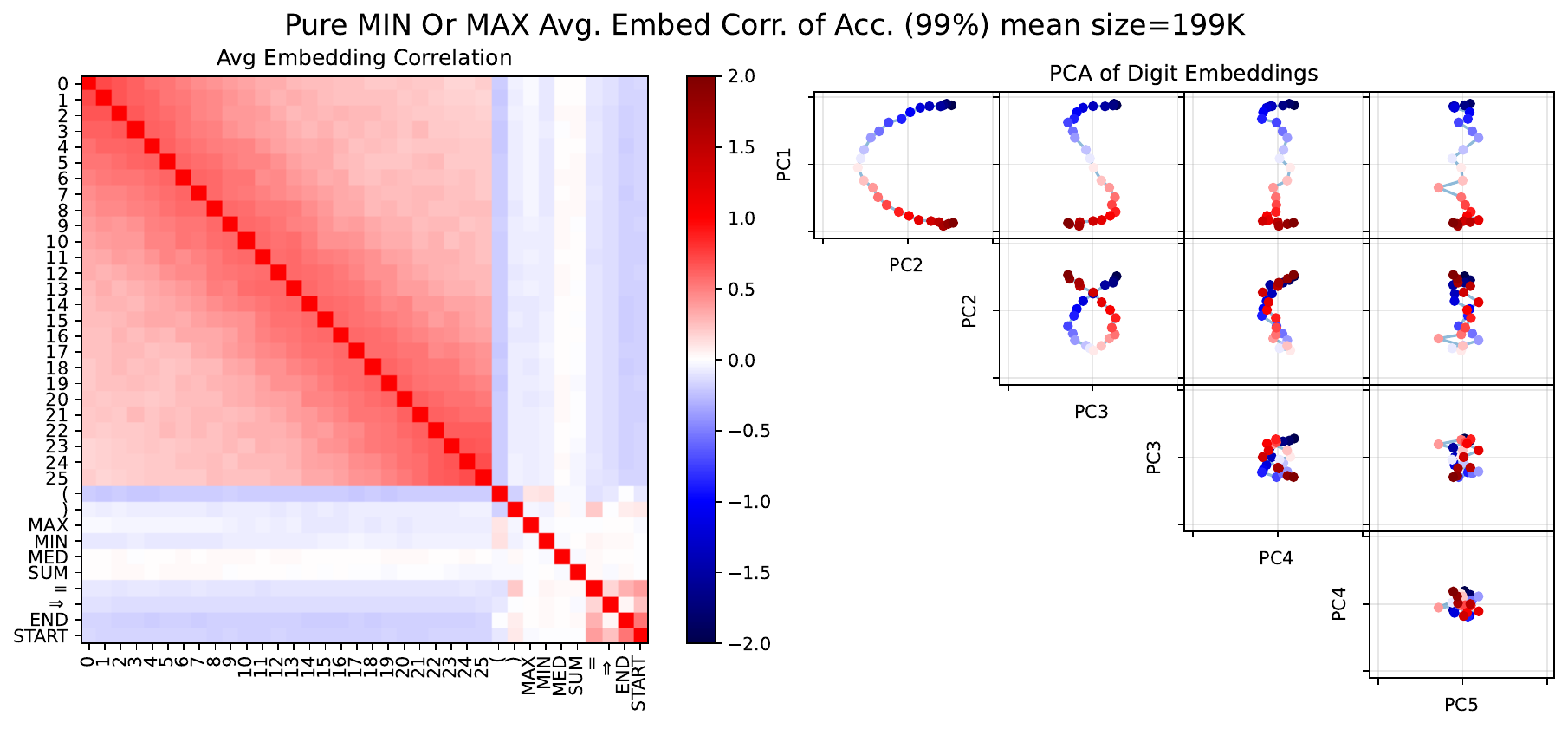}
    \includegraphics[width=0.49\linewidth]{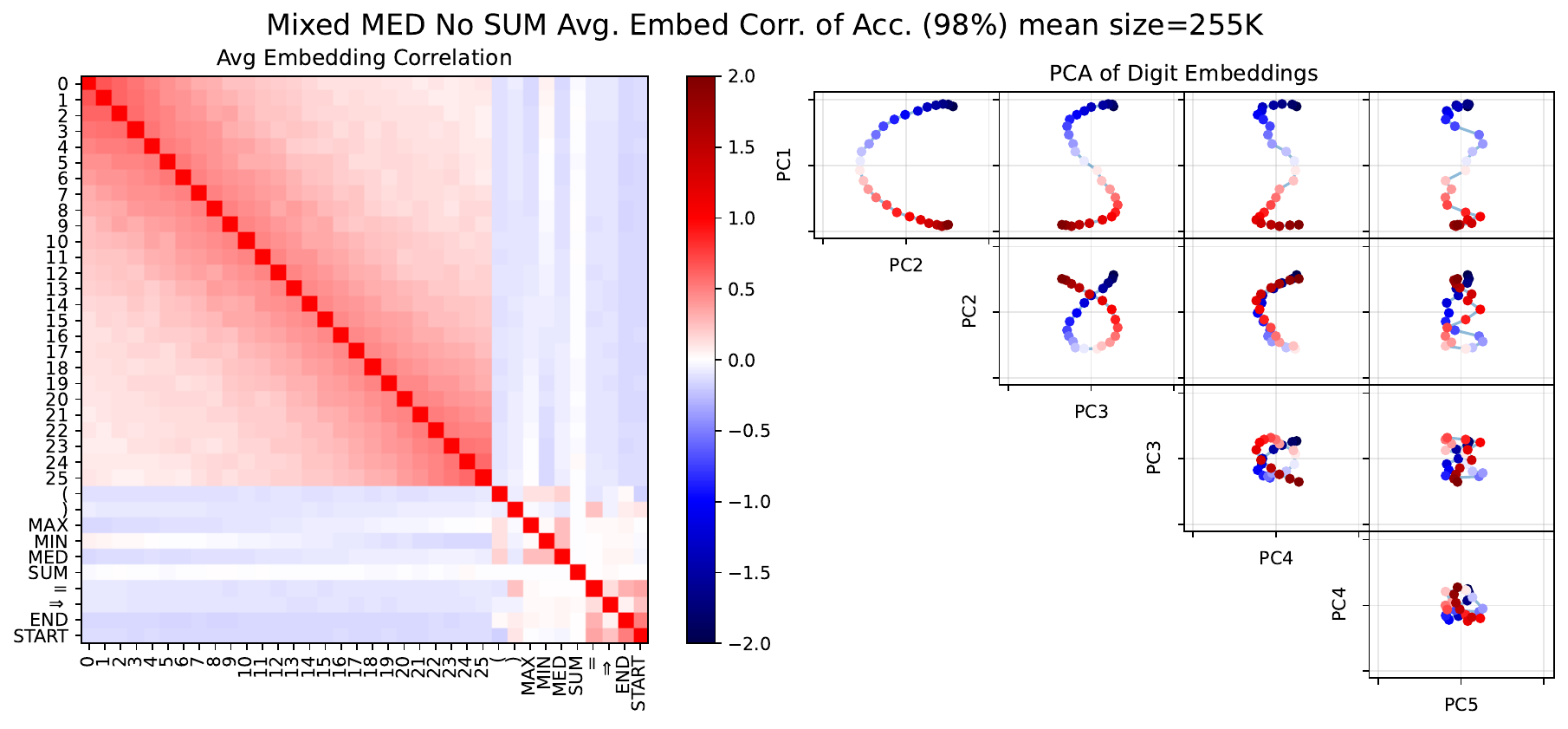}
    \includegraphics[width=0.49\linewidth]{figs/2025-05/base26/average_corr_pca_grouped_Mixed_SUM.pdf}
    \includegraphics[width=0.49\linewidth]{figs/2025-05/base26/average_corr_pca_grouped_Pure_SUM.pdf}
    \caption{
    \textbf{PCA of embeddings:}
    We choose all models which reached over $90$\% test accuracy. 
    Each row shows the average correlation matrix and top PCs for models trained on either a single operation, e.g. Pure MAX, or all mixtures involving a given operation, e.g. Mixed SUM. 
    Interestingly, pure SUM does not show a discernible structure in the embeddings, whereas all other cases do. 
    Notably, Mixed SUM models exhibit a prominent odd-even separation in PC5.  
    }
    \label{fig:corr-PCA-pure-mixed-26}
    \vspace{-10pt}
\end{figure}

\begin{figure}
    \centering
    \includegraphics[width=1\linewidth]{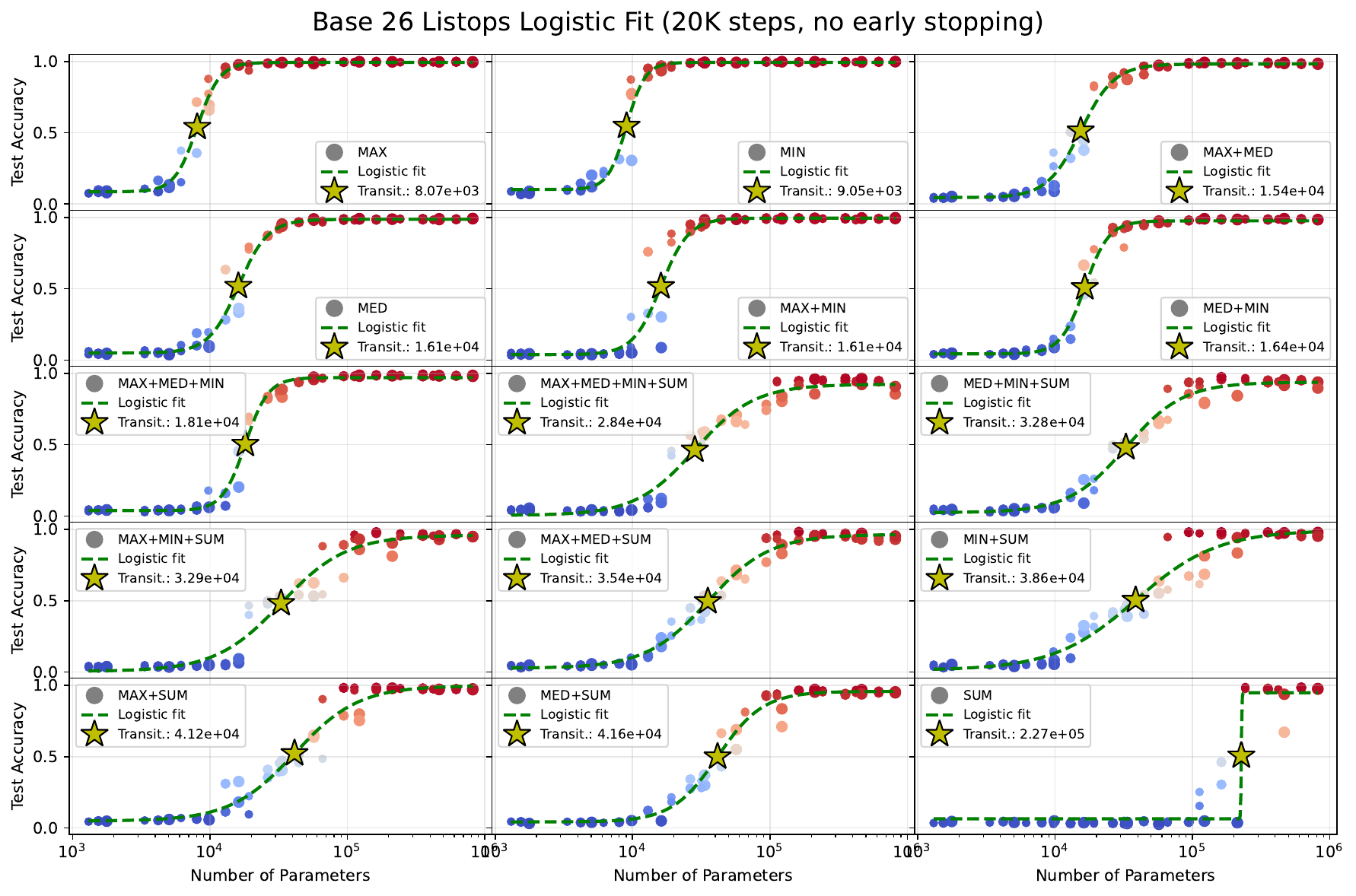}
    \includegraphics[width=1\linewidth]{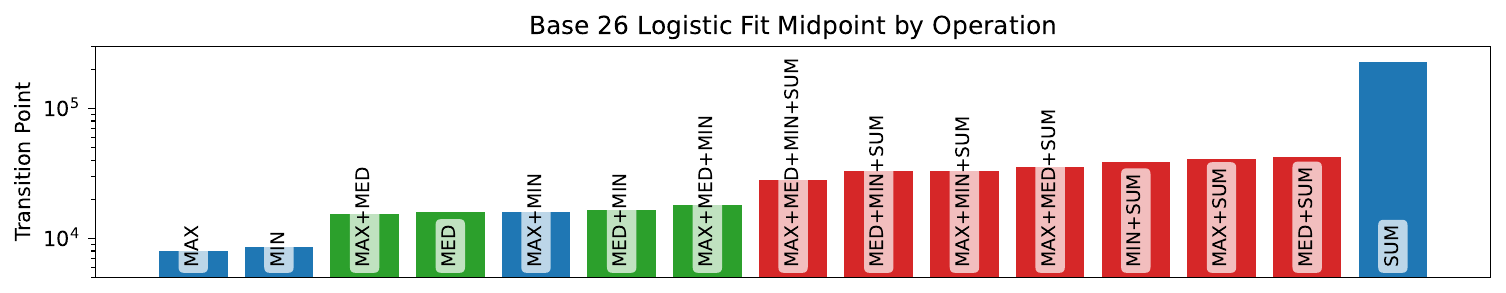}
    \caption{\textbf{Emergence of abilities in ListOps:} 
    Each plot shows the same group of small transformer models trained on a different mix of the four operations MAX, MIN, MED, and SUM. 
    Red dots are models reaching more than 50\% accuracy, and blue dots are less than 50\%. 
    The dashed green line is a logistic fit, and the yellow star indicates the transition point at 50\%. 
    The x-axis is the model size (number of parameters), and the plots are sorted in ascending order of transition points. 
    The bottom panel shows a bar plot of the model sizes at the transition points, with each group distinguished by a different color. 
    }
    \label{fig:transition26}
\end{figure}

\begin{figure}[h]
    \centering
    \includegraphics[width=1\linewidth]{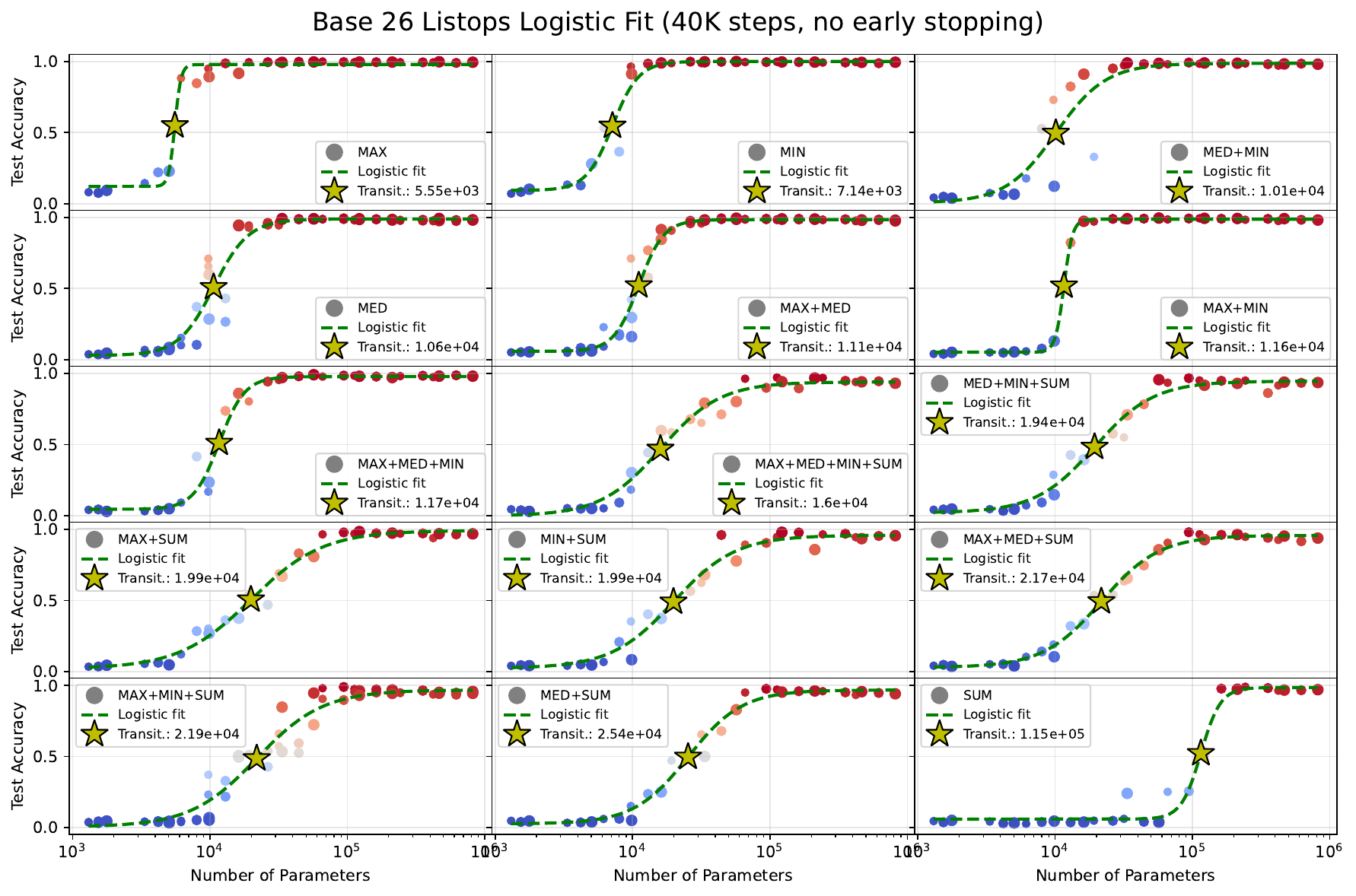}
    \includegraphics[width=1\linewidth]{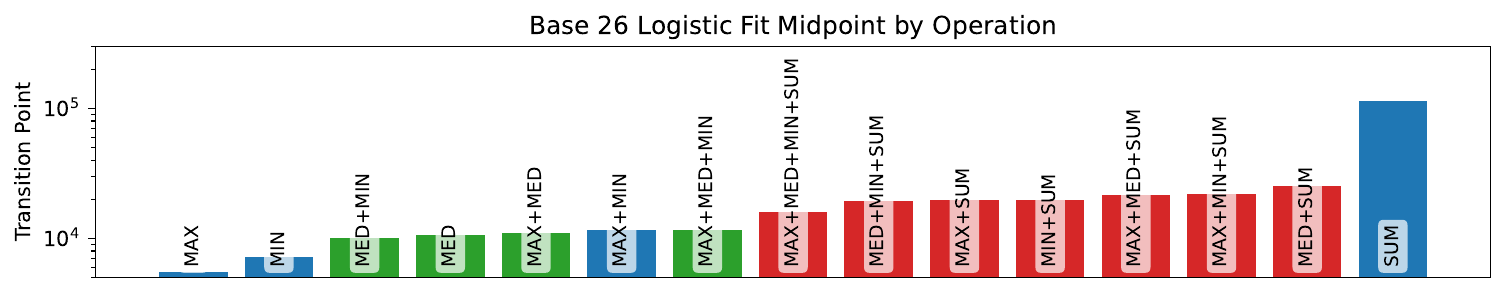}
    \caption{\textbf{Emergence of abilities in ListOps, 40k steps:} 
    Long training. 
    Each plot shows the same group of small transformer models trained on a different mix of the four operations MAX, MIN, MED, and SUM. 
    Red dots are models reaching more than 50\% accuracy, and blue dots are less than 50\%. 
    The dashed green line is a logistic fit, and the yellow star indicates the transition point at 50\%. 
    The x-axis is the model size (number of parameters), and the plots are sorted in ascending order of transition points. 
    The bottom panel shows a bar plot of the model sizes at the transition points, with each group distinguished by a different color. 
    }
    \label{fig:transition26-40k}
\end{figure}

\clearpage 
\subsection{Modulo 10 \label{ap:mod10} }
We conducted the same experiments also on mod 10. 
The smaller number of numbers makes definitive statements about some of the patterns more challenging. 
But all the patterns we observed in mod 26 also have parallels in mod 10, including the prominent odd-even split for operations involving SUM. 
Token vocabulary: \verb|%()+-/0123456789=>es|


\begin{figure}
    \centering
    \includegraphics[width=0.49\linewidth]{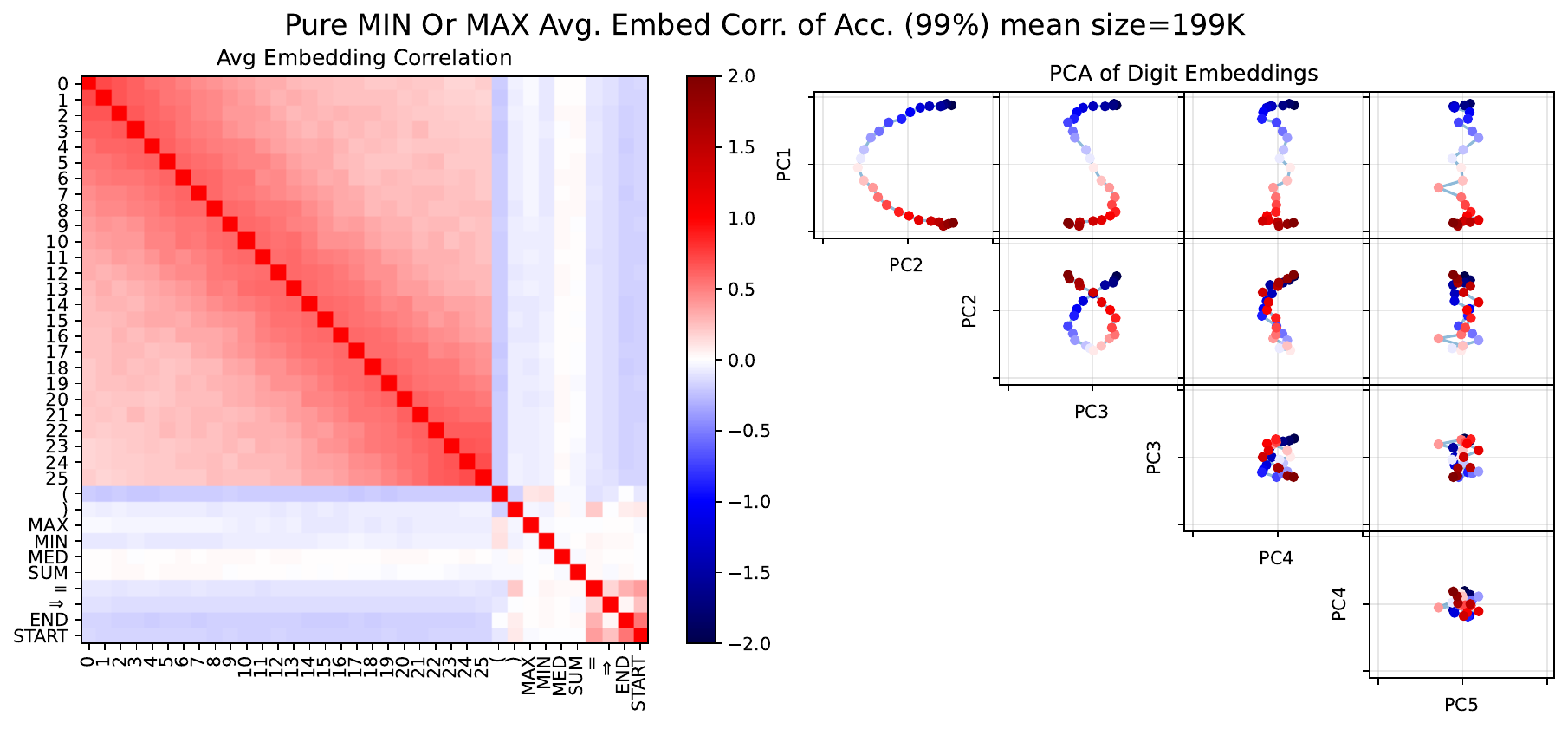}
    \includegraphics[width=0.49\linewidth]{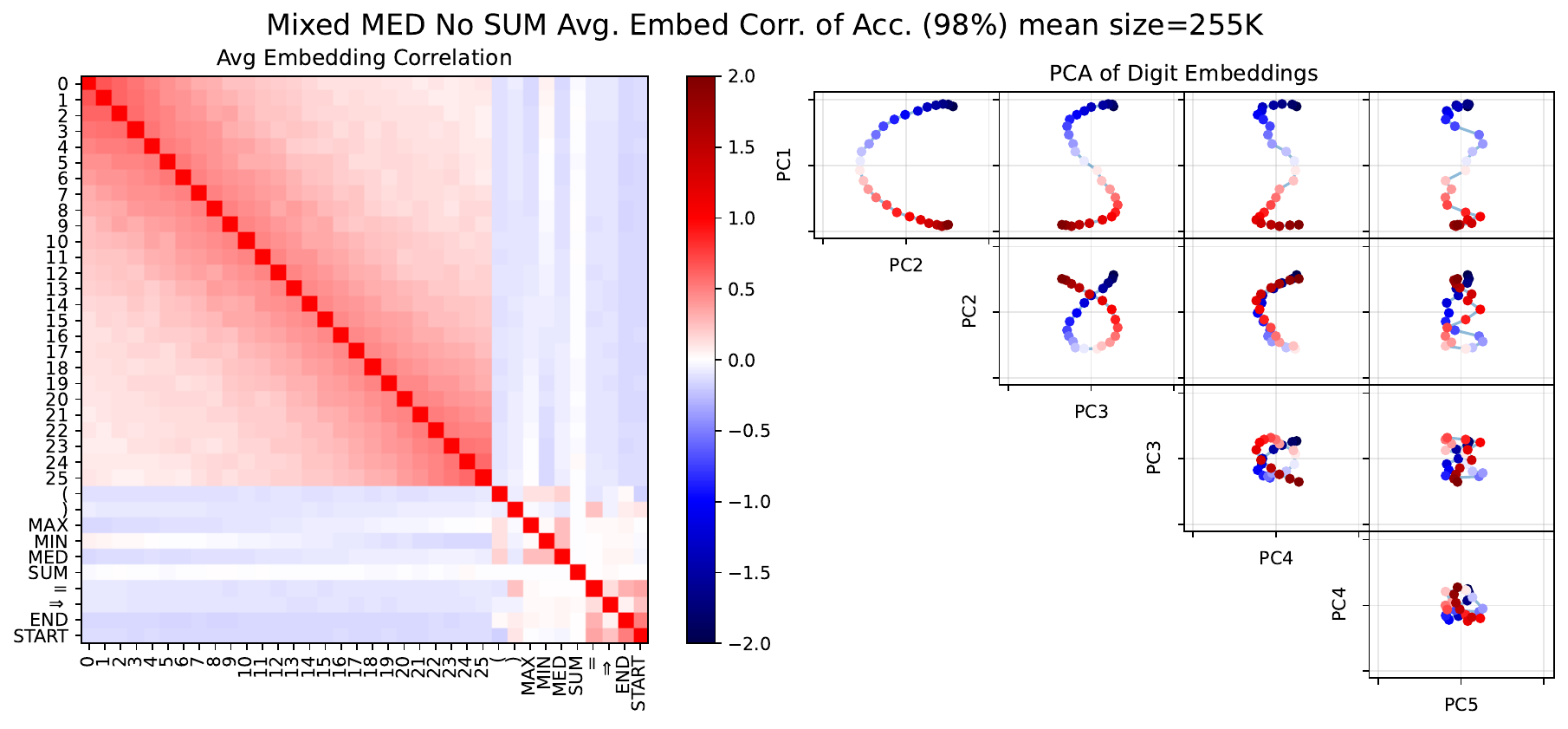}
    \includegraphics[width=0.49\linewidth]{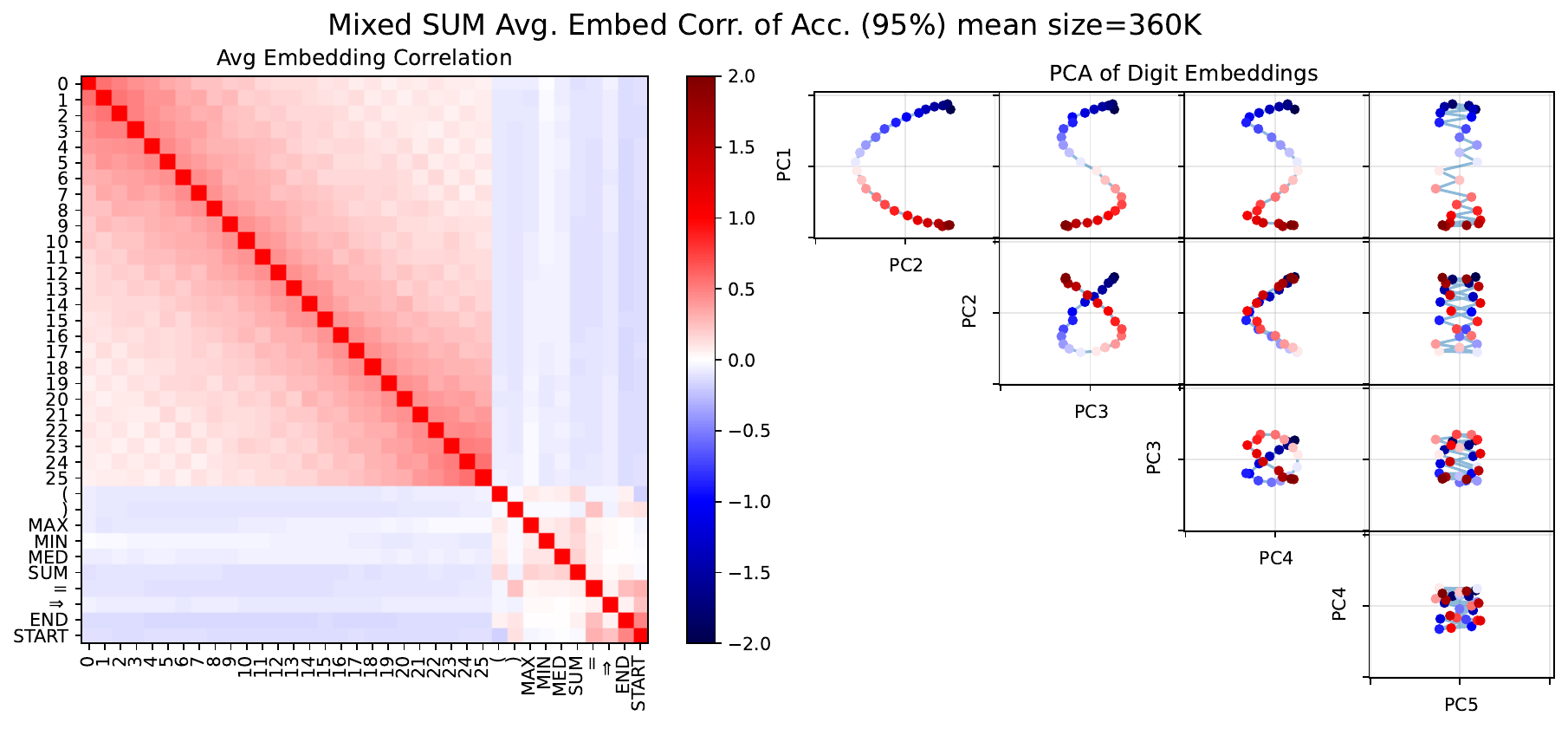}
    \includegraphics[width=0.49\linewidth]{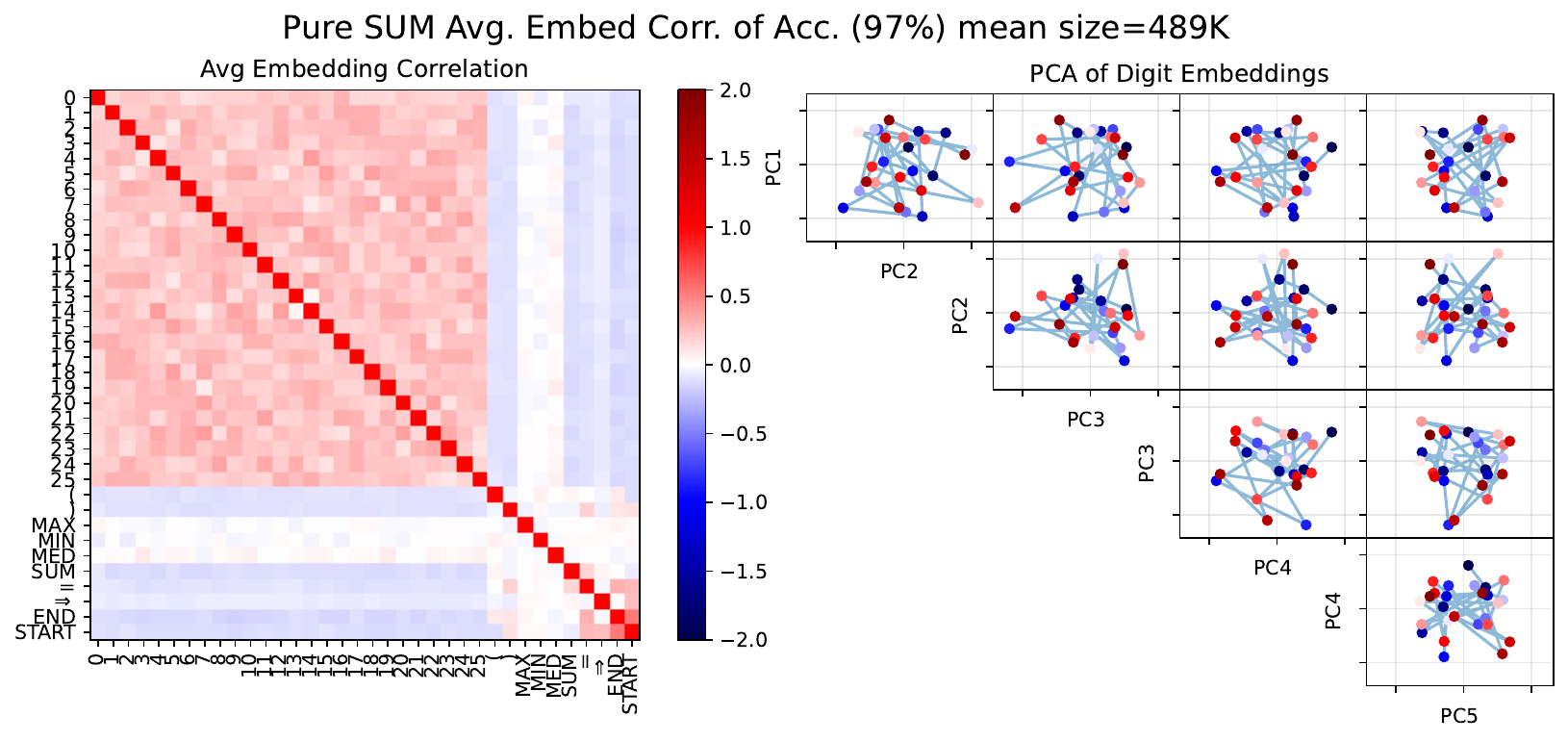}
    \caption{
    \textbf{PCA of embeddings:}
    We choose all models which reached over $90$\% test accuracy. 
    Each row shows the average correlation matrix and top PCs for models trained on either a single operation, e.g. Pure MAX, or all mixtures involving a given operation, e.g. Mixed SUM. 
    Interestingly, pure SUM does not show a discernible structure in the embeddings, whereas all other cases do. 
    Notably, Mixed SUM models exhibit a prominent odd-even separation in PC5.  
    \nd{Replace!!!}
    }
    \label{fig:corr-PCA-pure-mixed-10}
    \vspace{-10pt}
\end{figure}

\begin{figure}
    \centering
    \includegraphics[width=0.49\linewidth]{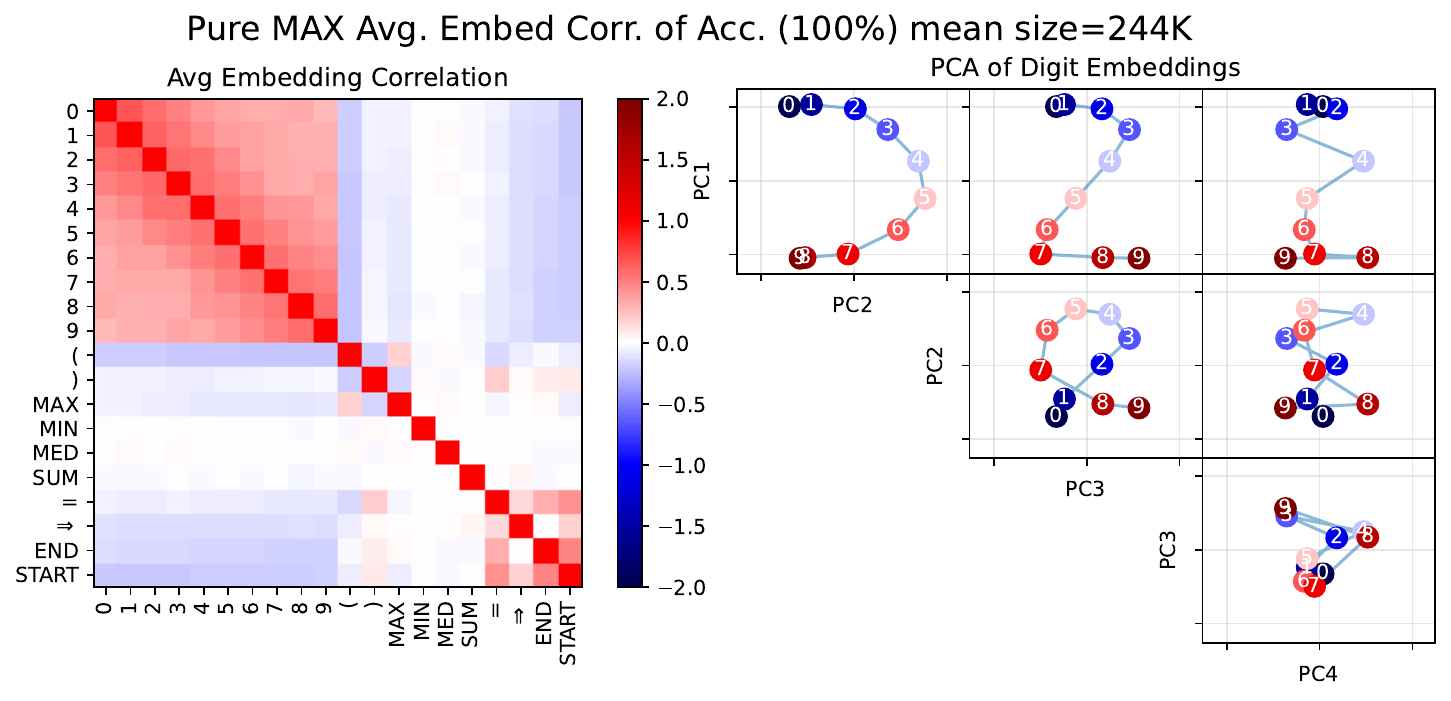}
    \includegraphics[width=0.49\linewidth]{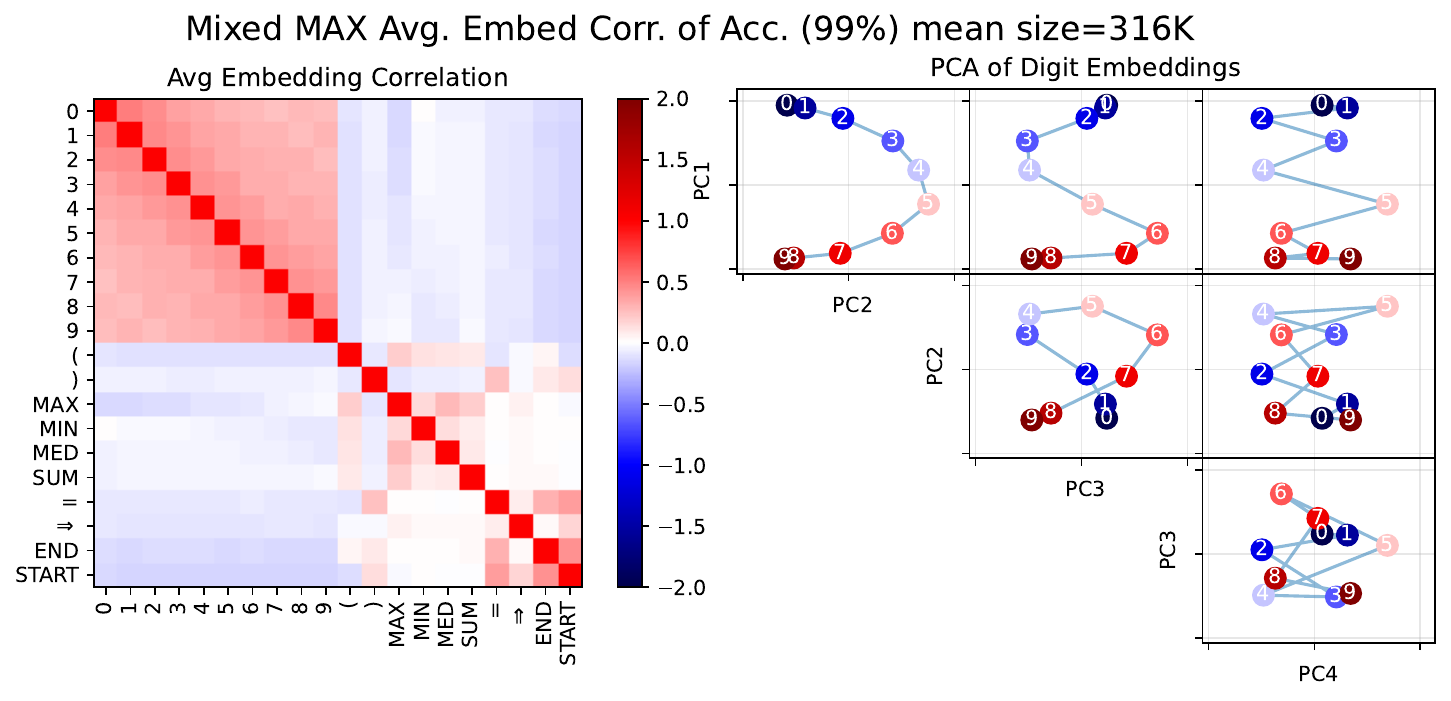}
    \includegraphics[width=0.49\linewidth]{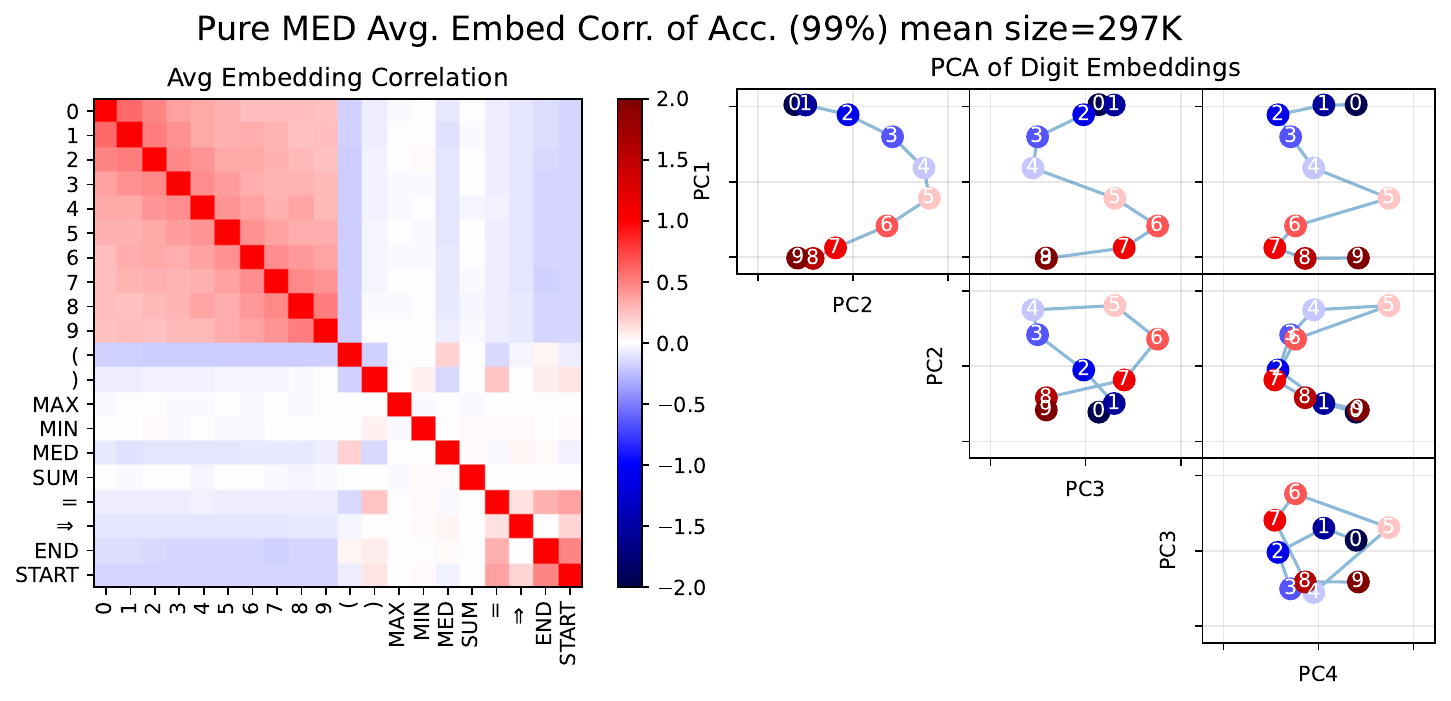}
    \includegraphics[width=0.49\linewidth]{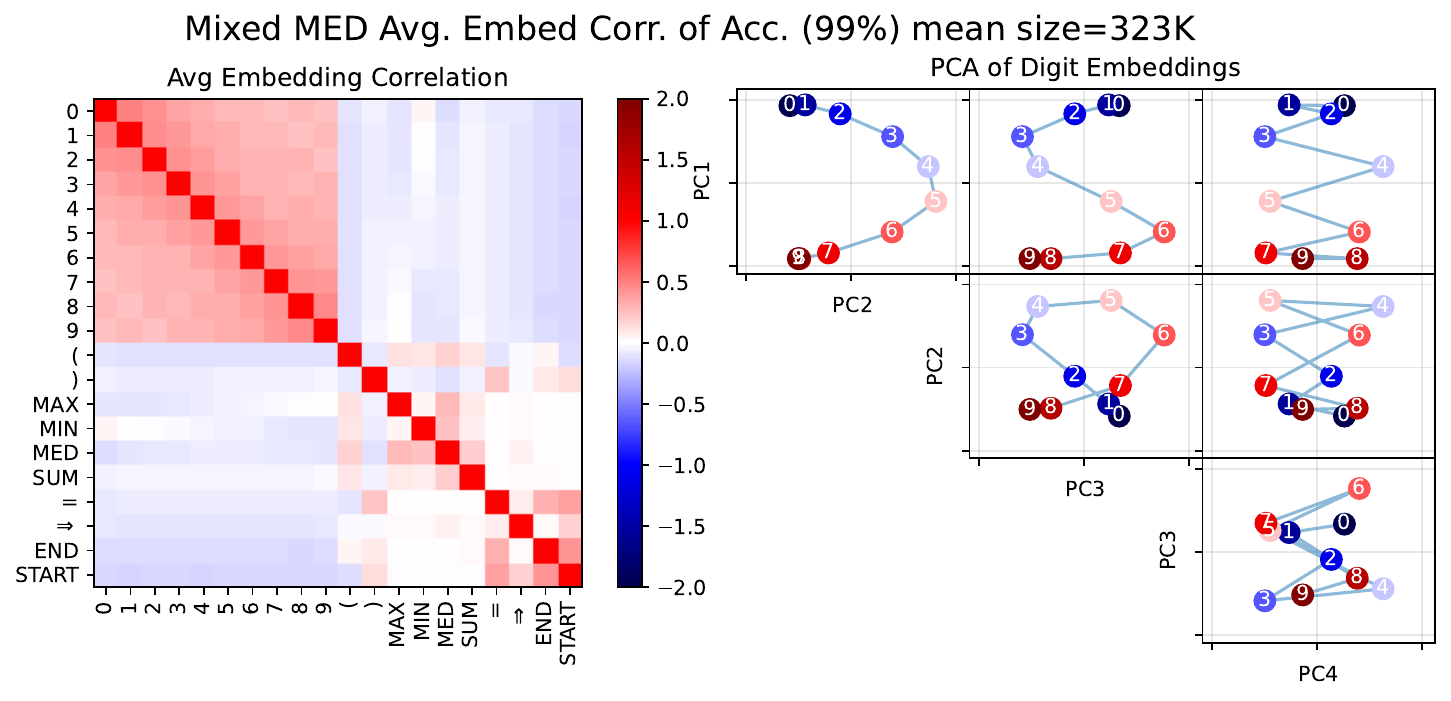}
    \includegraphics[width=0.49\linewidth]{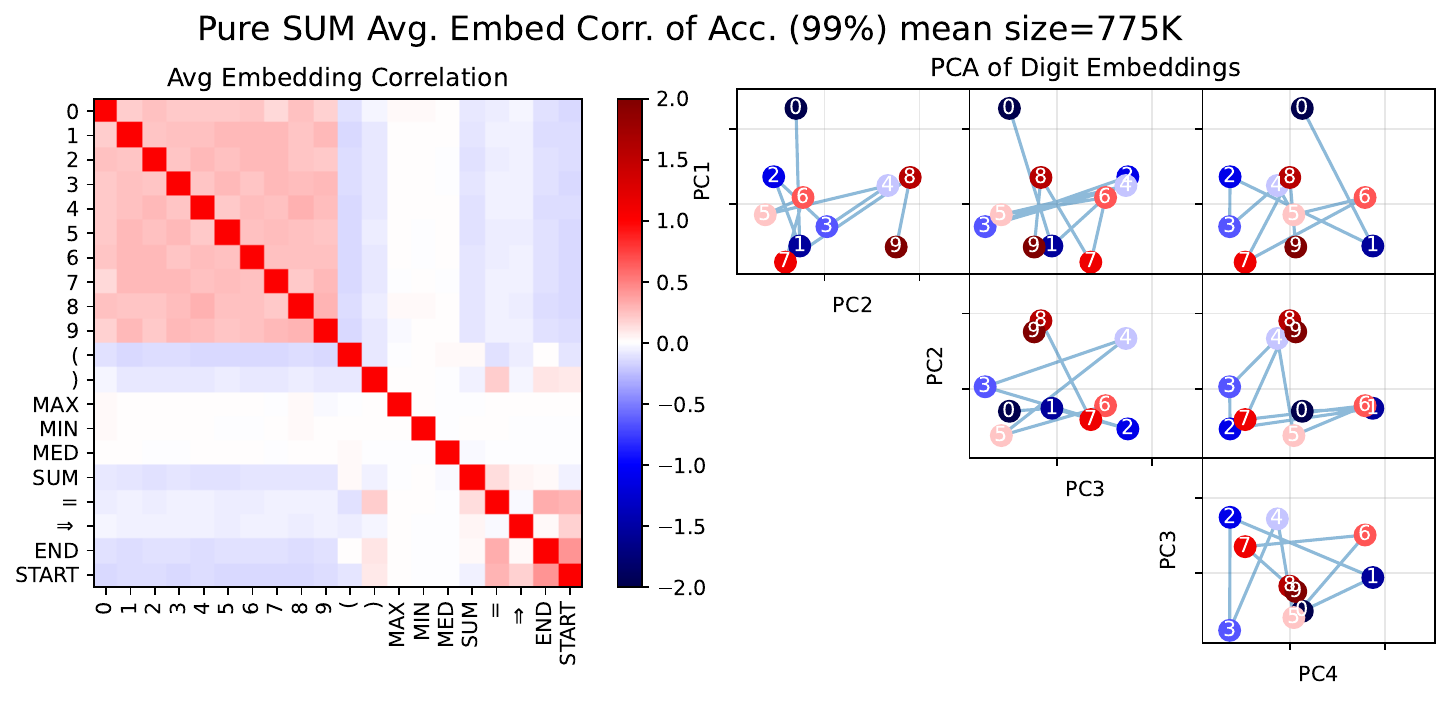}\includegraphics[width=0.49\linewidth]{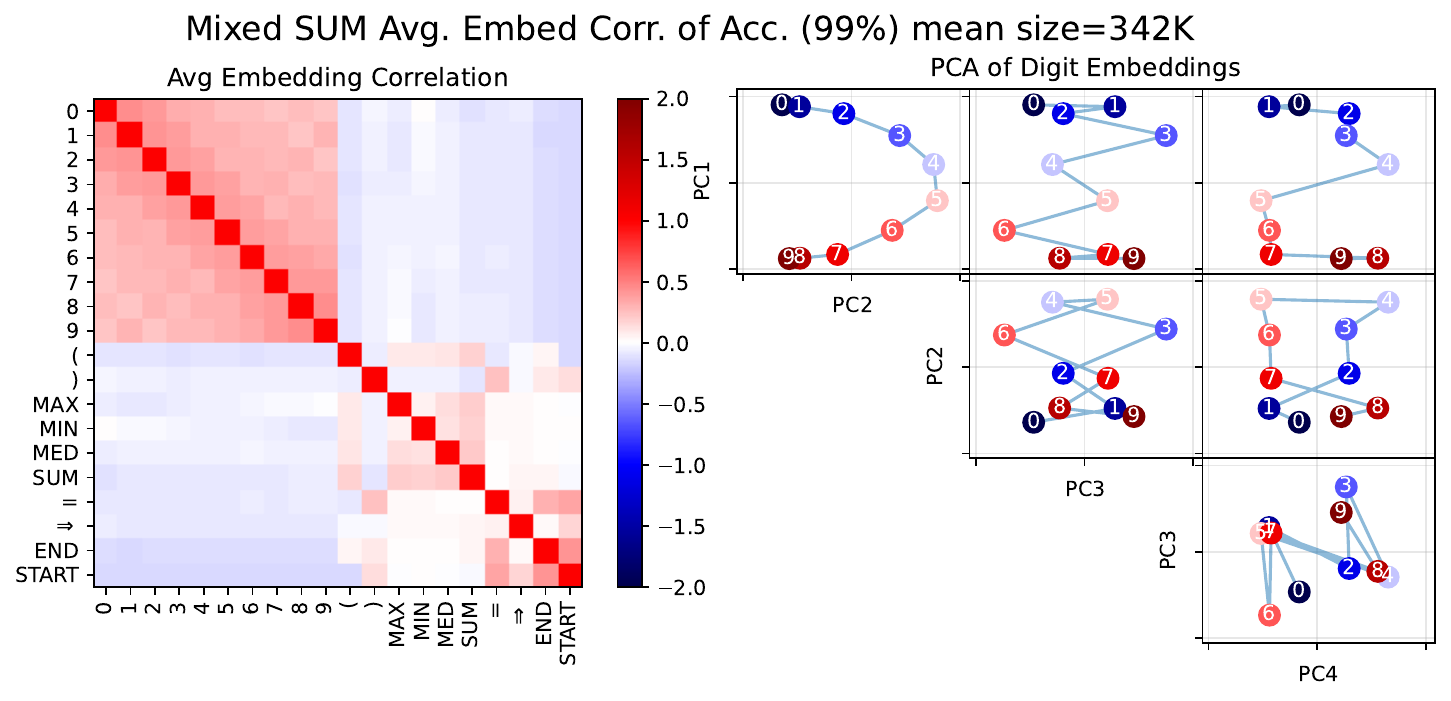}
    \caption{
    \textbf{PCA of embeddings:}
    We choose all models which reached over $95$\% test accuracy. 
    Each row shows the average correlation matrix and top PCs for models trained on either a single operation, e.g. Pure+MAX, or all mixtures involving a given operation, e.g. Mixed+MAX. 
    Again, pure SUM does not show a discernible structure in the embeddings, whereas all cases do. 
    }
    \label{fig:corr-PCA-pure-mixed}
\end{figure}

\begin{figure}
    \centering
    \includegraphics[width=0.49\linewidth]{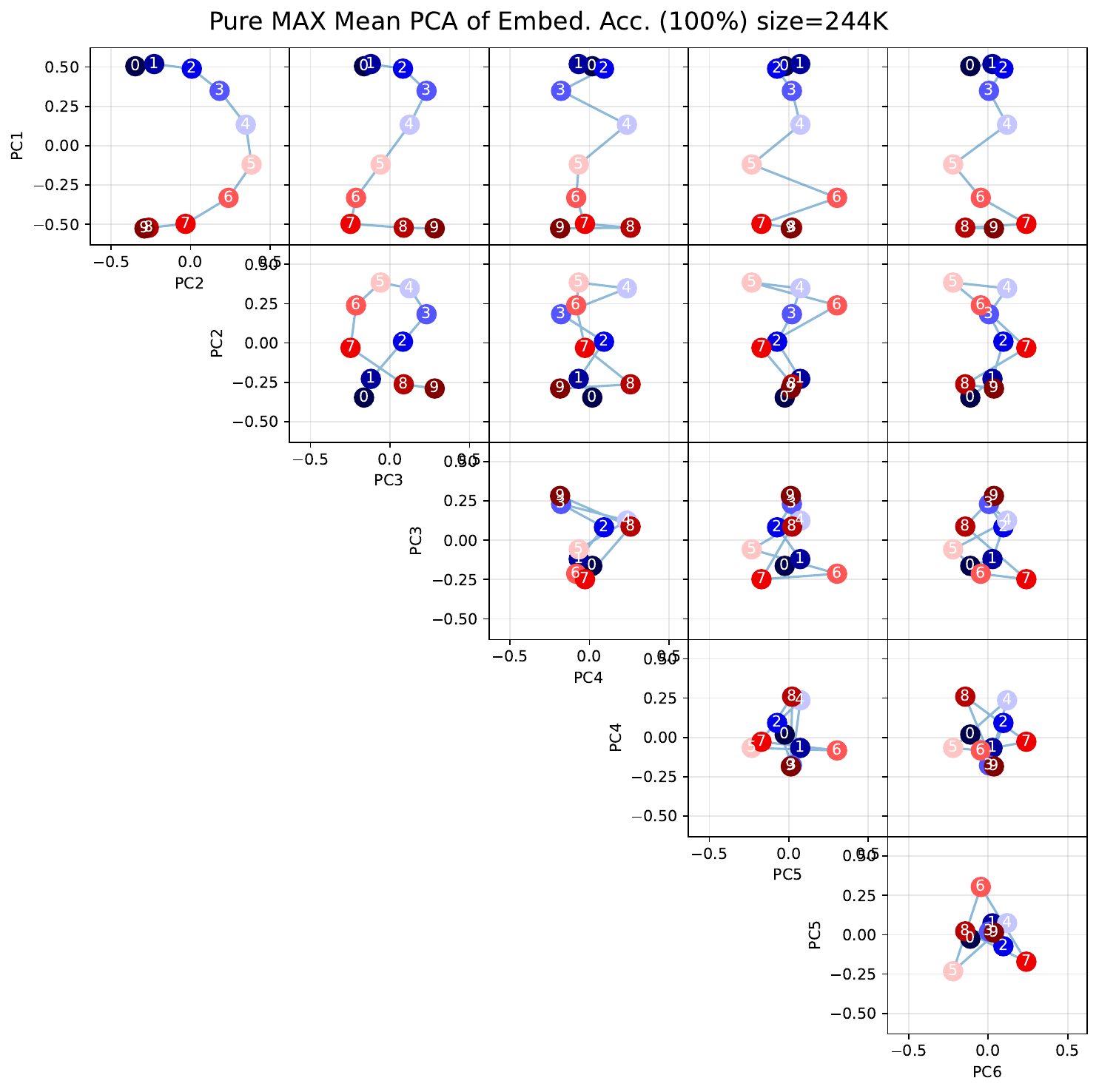}
    \includegraphics[width=0.49\linewidth]{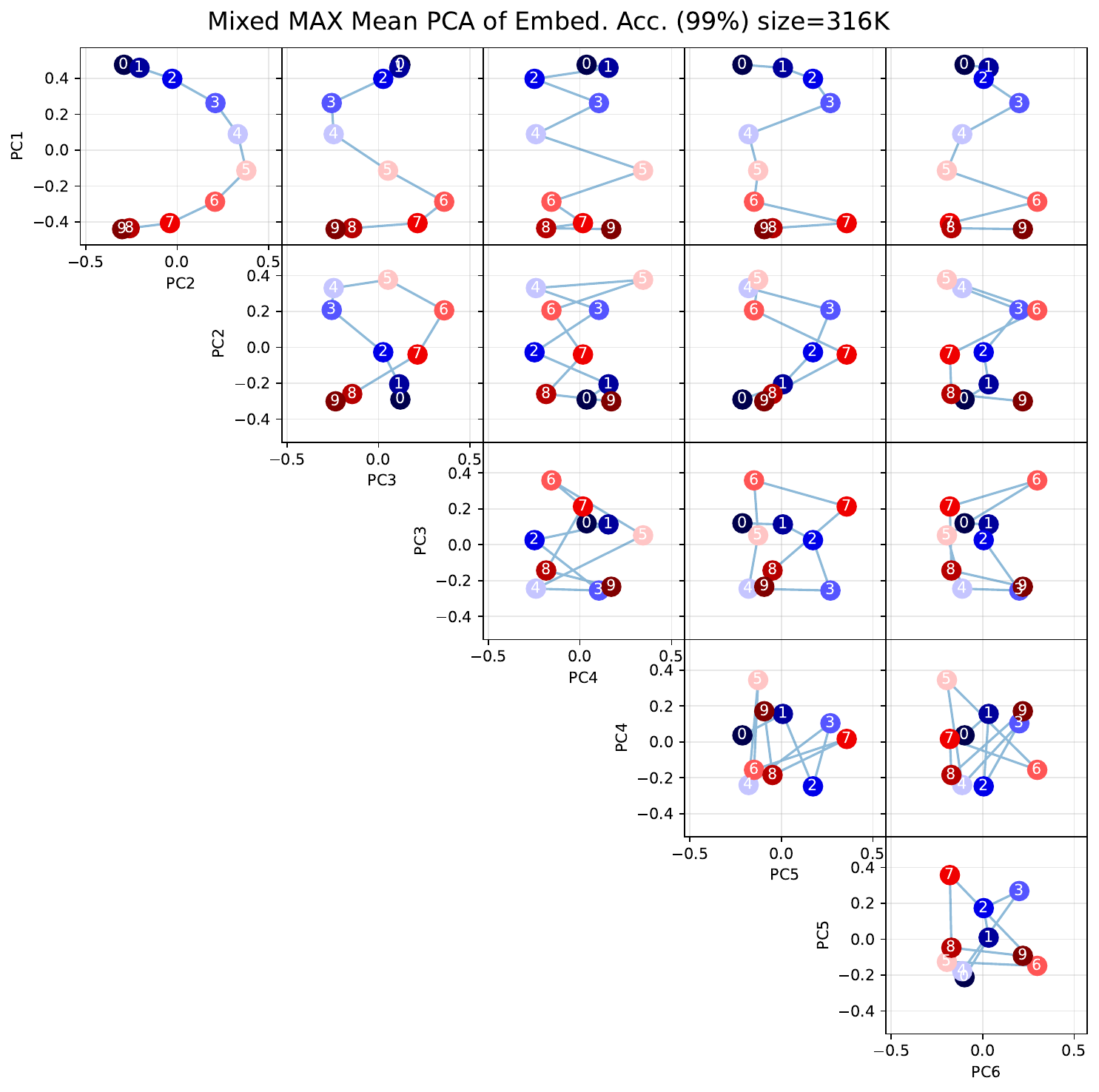}
    \includegraphics[width=0.49\linewidth]{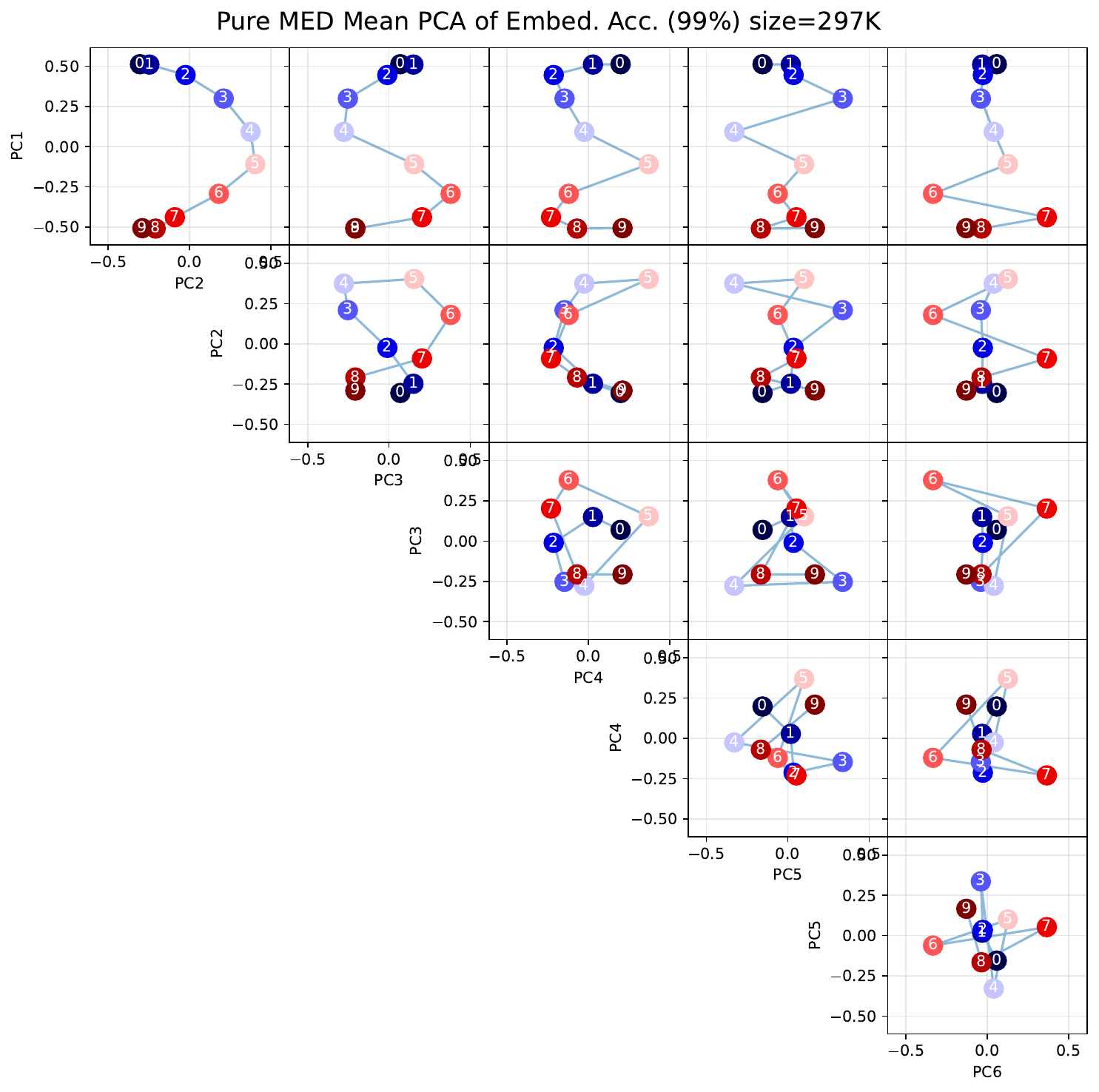}
    \includegraphics[width=0.49\linewidth]{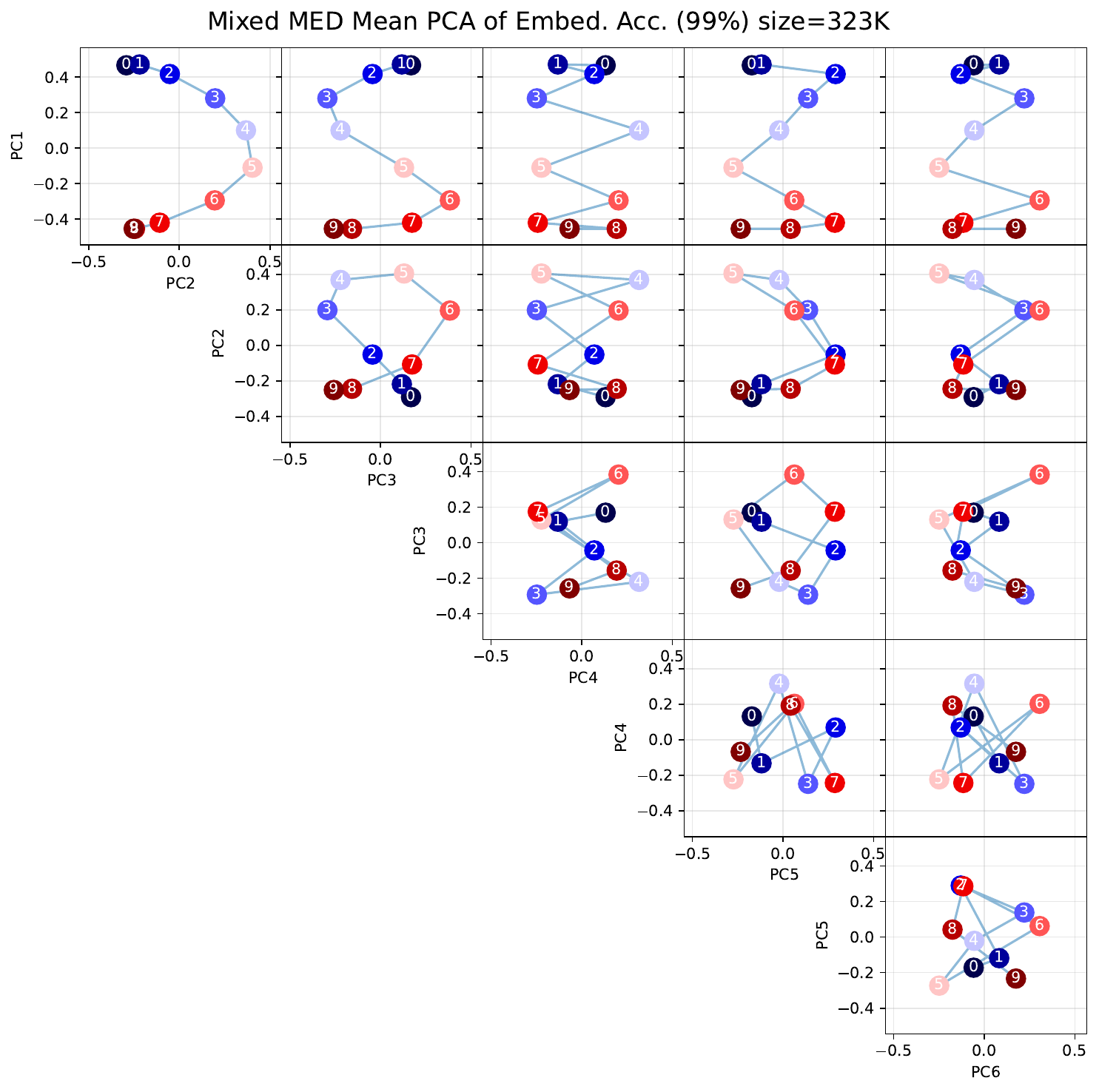}
    \includegraphics[width=0.49\linewidth]{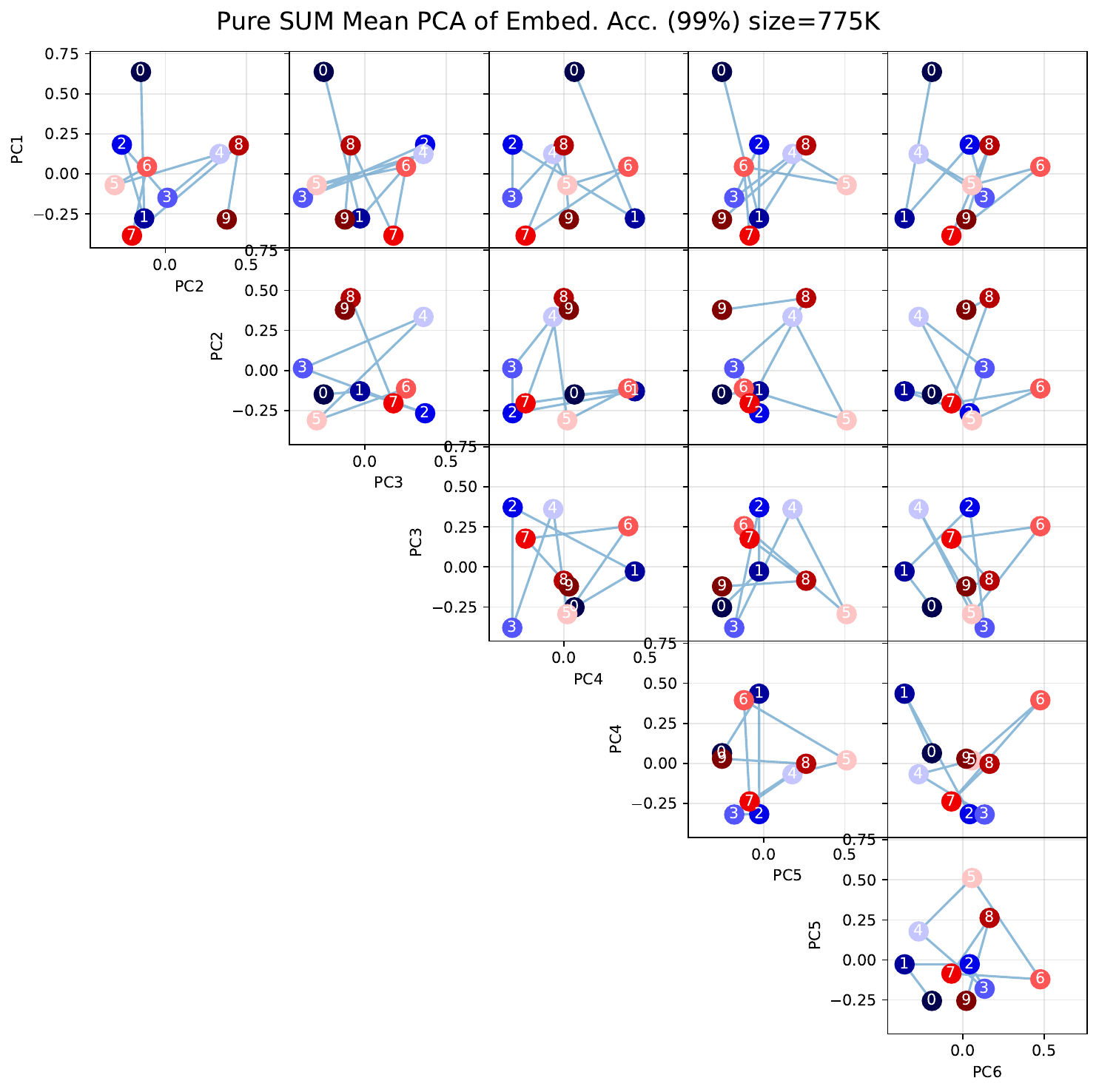}
    \includegraphics[width=0.49\linewidth]{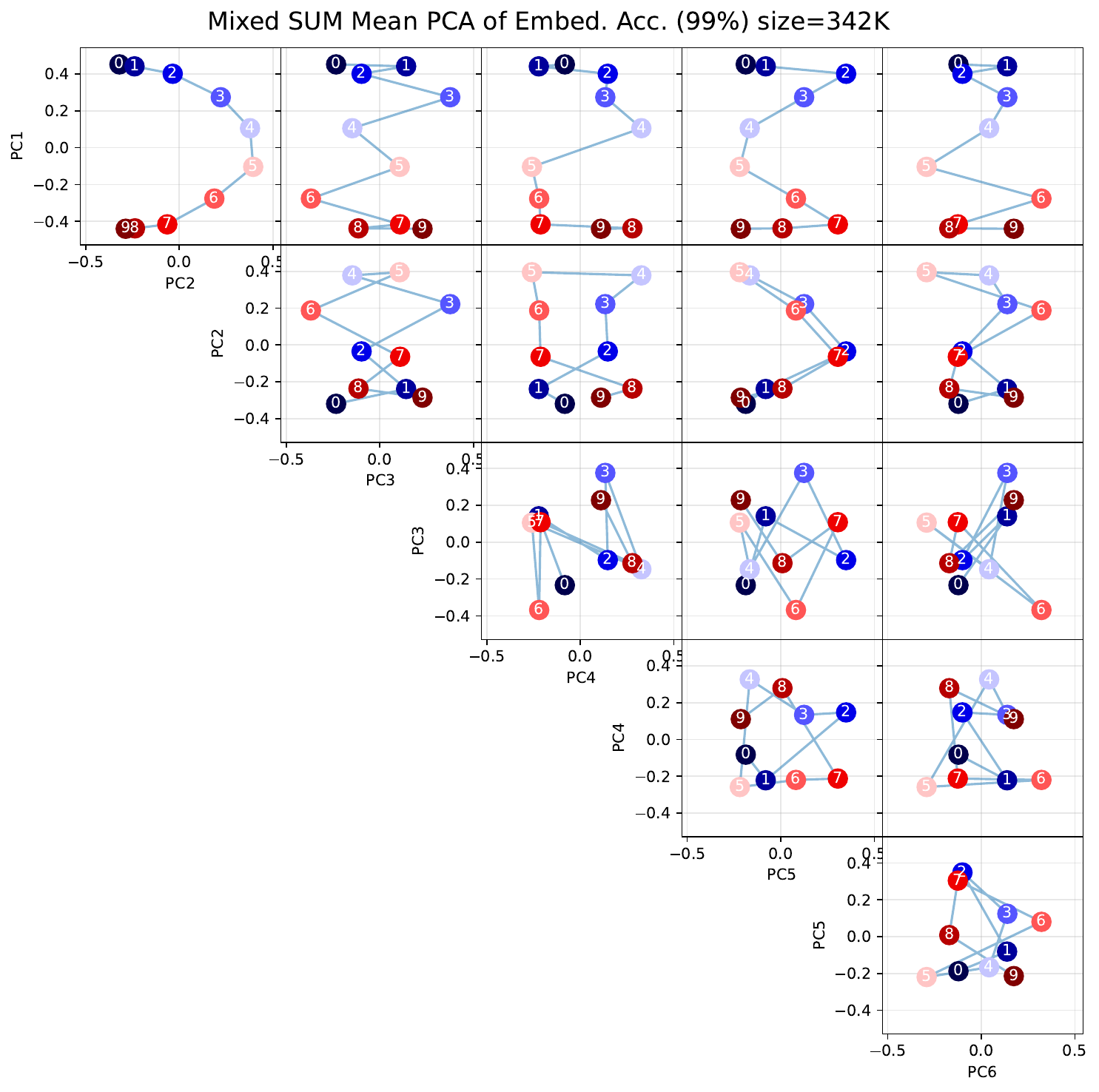}
    \caption{
    \textbf{PCA of embeddings:}
    We choose all models which reached over $95$\% test accuracy. 
    Each row shows the average correlation matrix and top PCs for models trained on either a single operation, e.g. Pure+MAX, or all mixtures involving a given operation, e.g. Mixed+MAX. 
    Again, pure SUM does not show a discernible structure in the embeddings, whereas all cases do. 
    }
    \label{fig:PCA-pure-mixed}
\end{figure}

\begin{figure}
    \centering
    \includegraphics[width=1\linewidth]{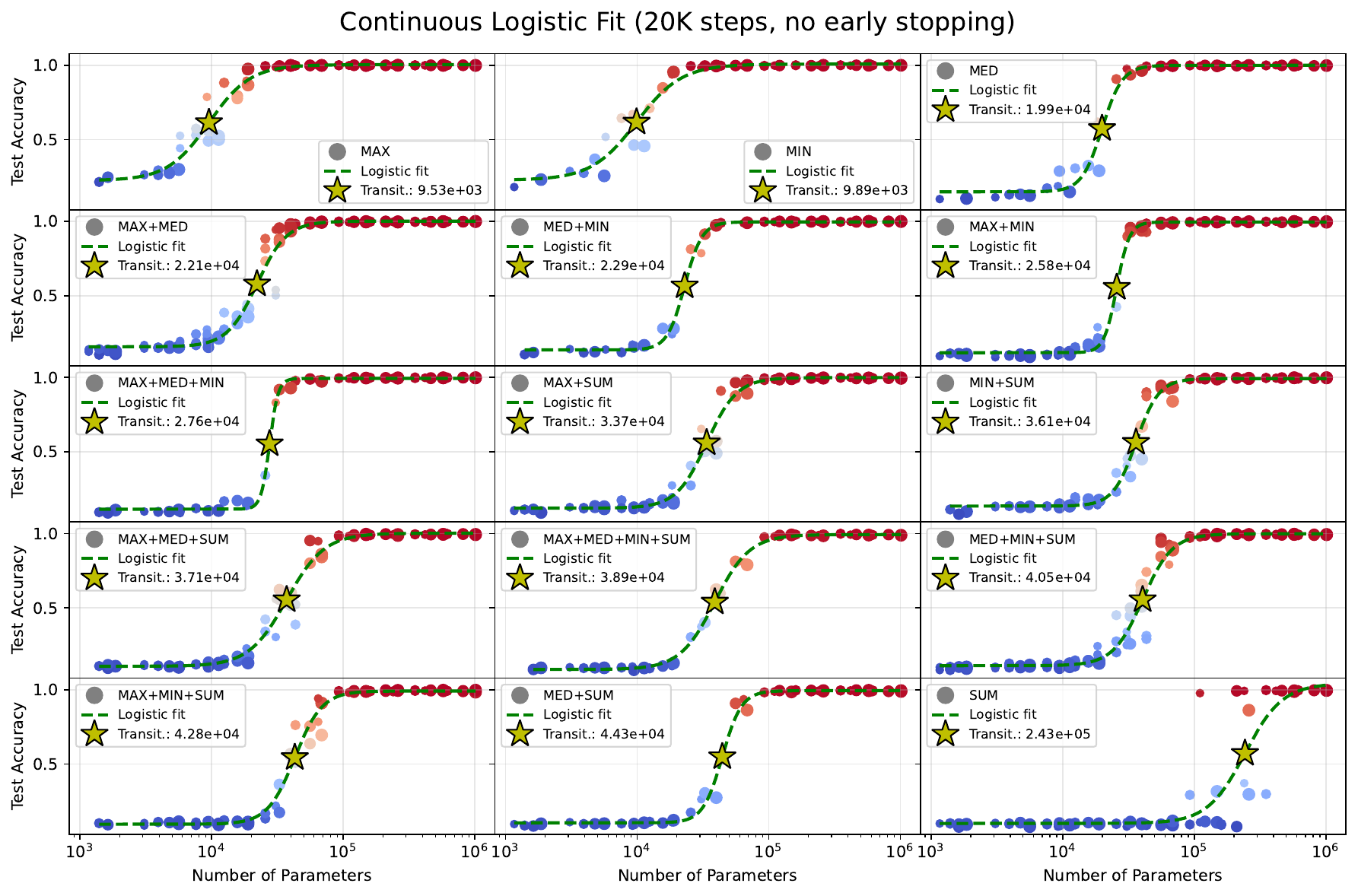}
    \includegraphics[width=1\linewidth]{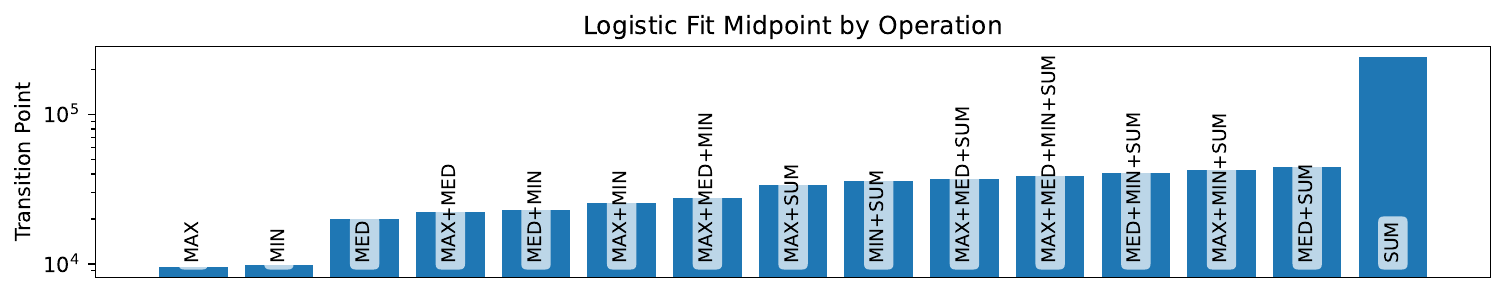}
    \caption{\textbf{Emergence of of abilities in ListOps, Mod 10:} 
    Each plot shows the same group of small transformer models trained on different variants of ListOps.  
    Each variant uses a different mix of the four operations MAX, MIN, MED, SUM. 
    Red dots are model reaching more than 50\% accuracy, and blue are less than 50\%. 
    The dashed green line is a logistic fit and the yellow star indicates the transition point at 50\%. 
    The x axis is the model size and the plots are sorted in ascending order of transition points. 
    The bottom is a bar plot showing the model size at the transition point. 
    We observe that SUM is a clear outlier, with
    models requiring significantly more parameters to learn SUM. 
    Surprisingly, combining SUM with other operations dramatically reduces the transition point, with model less than half the size easily reaching 100\% accuracy.  
    }
    \label{fig:transition}
\end{figure}

\begin{figure}
    \centering
    \includegraphics[width=0.49\linewidth]{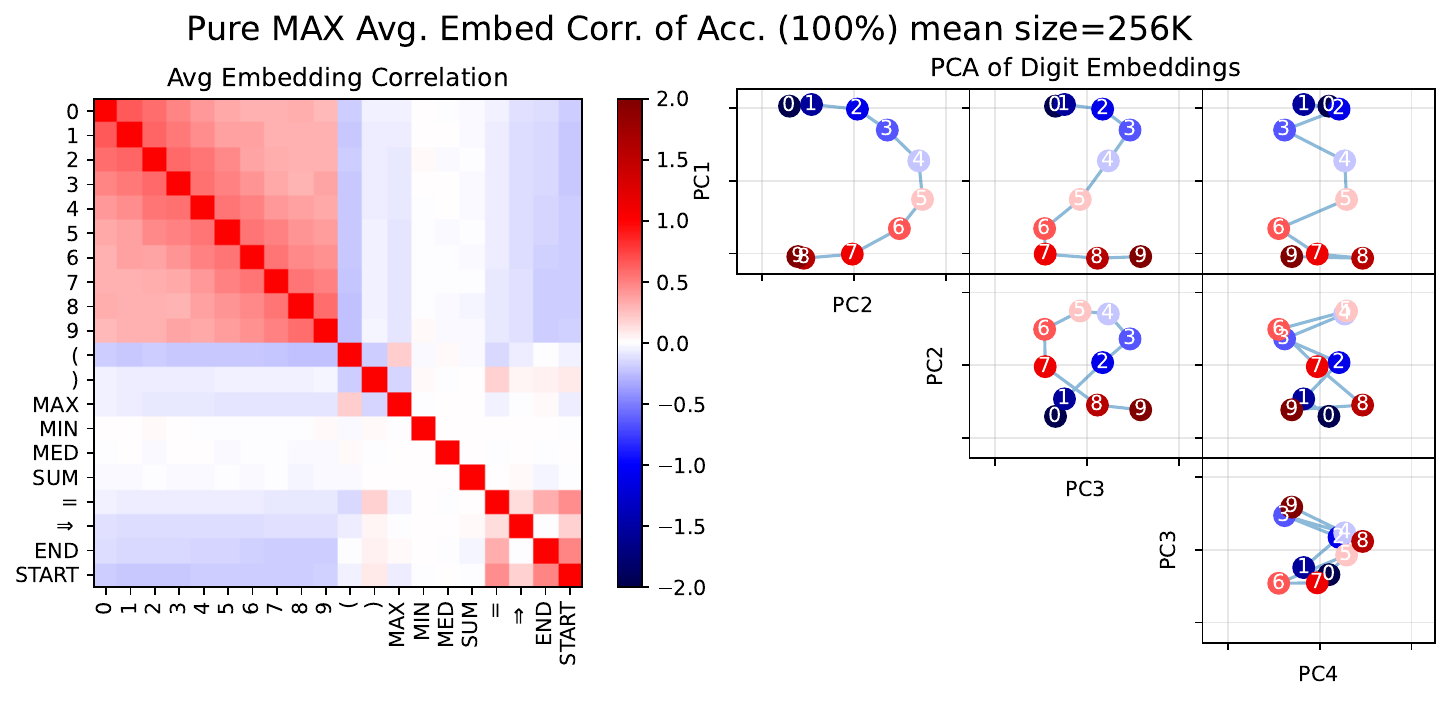}
    \includegraphics[width=0.49\linewidth]{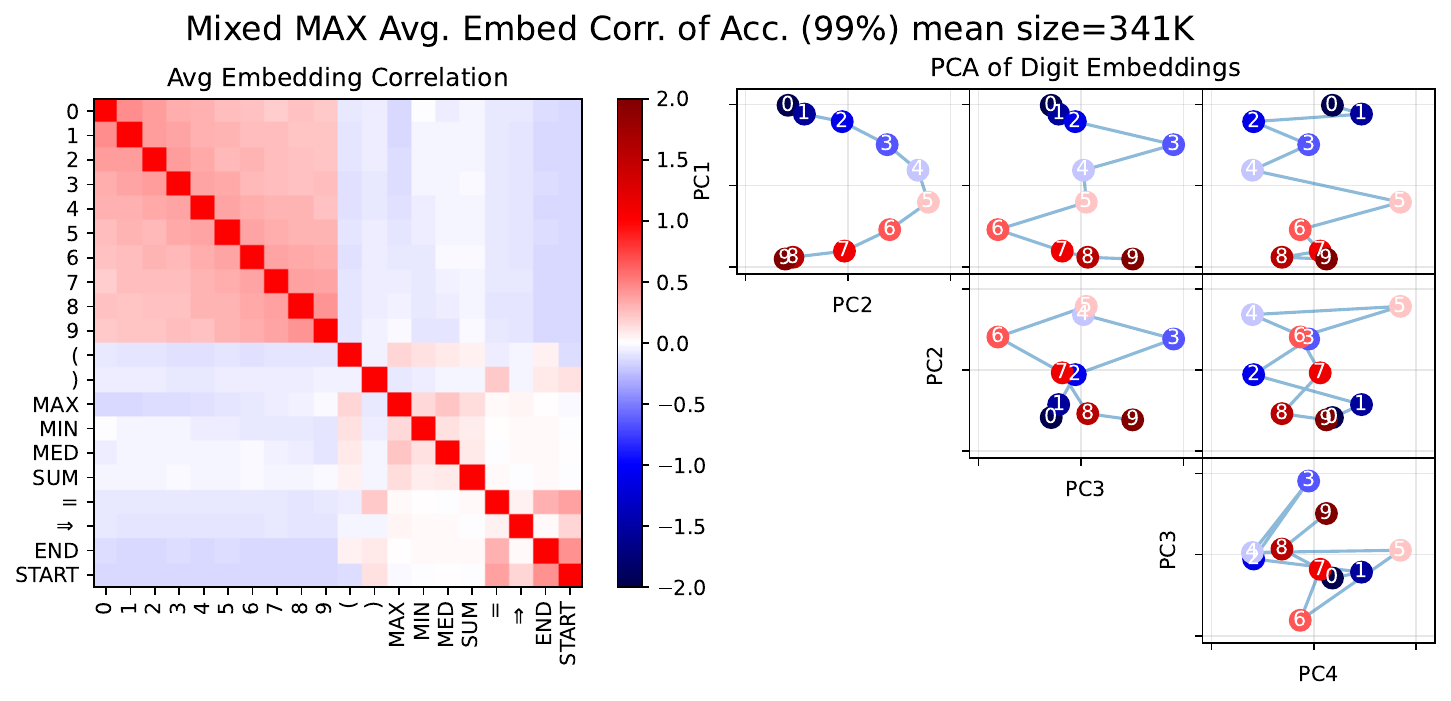}
    \includegraphics[width=0.49\linewidth]{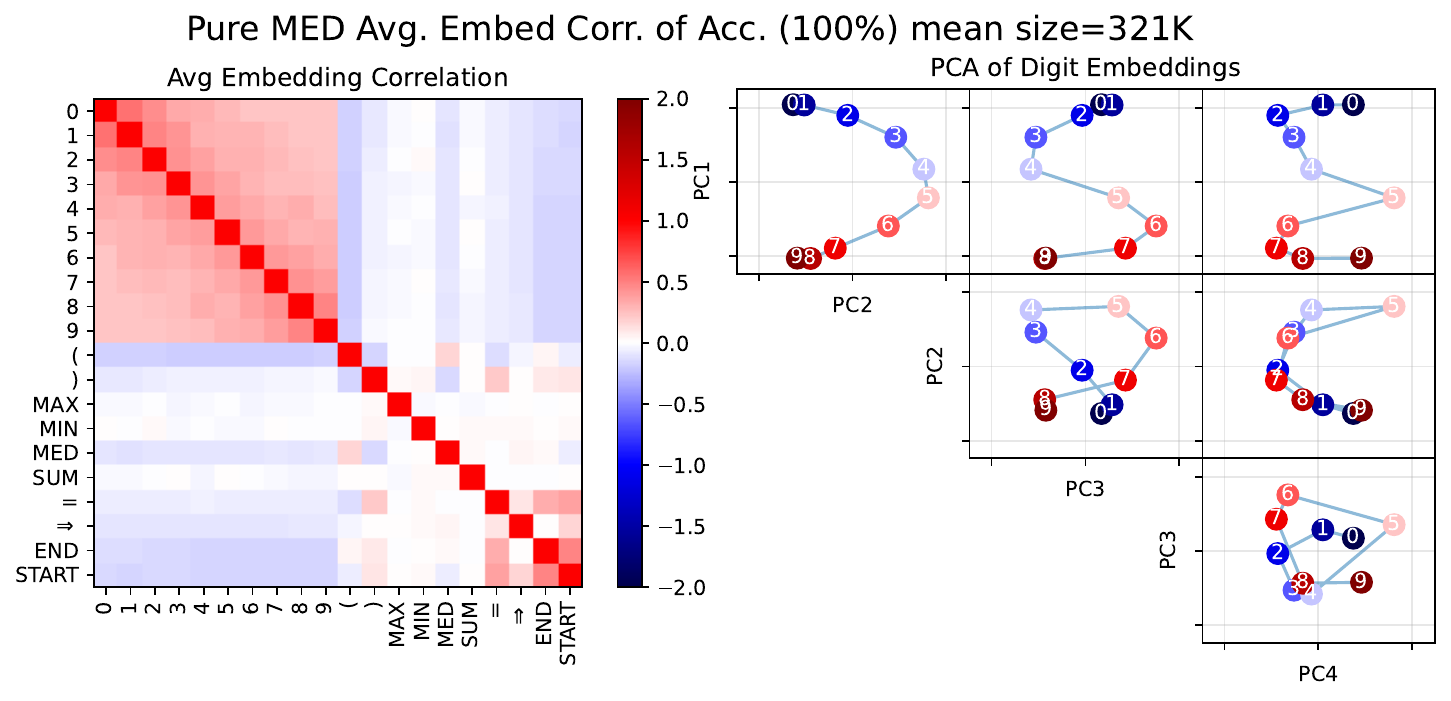}
    \includegraphics[width=0.49\linewidth]{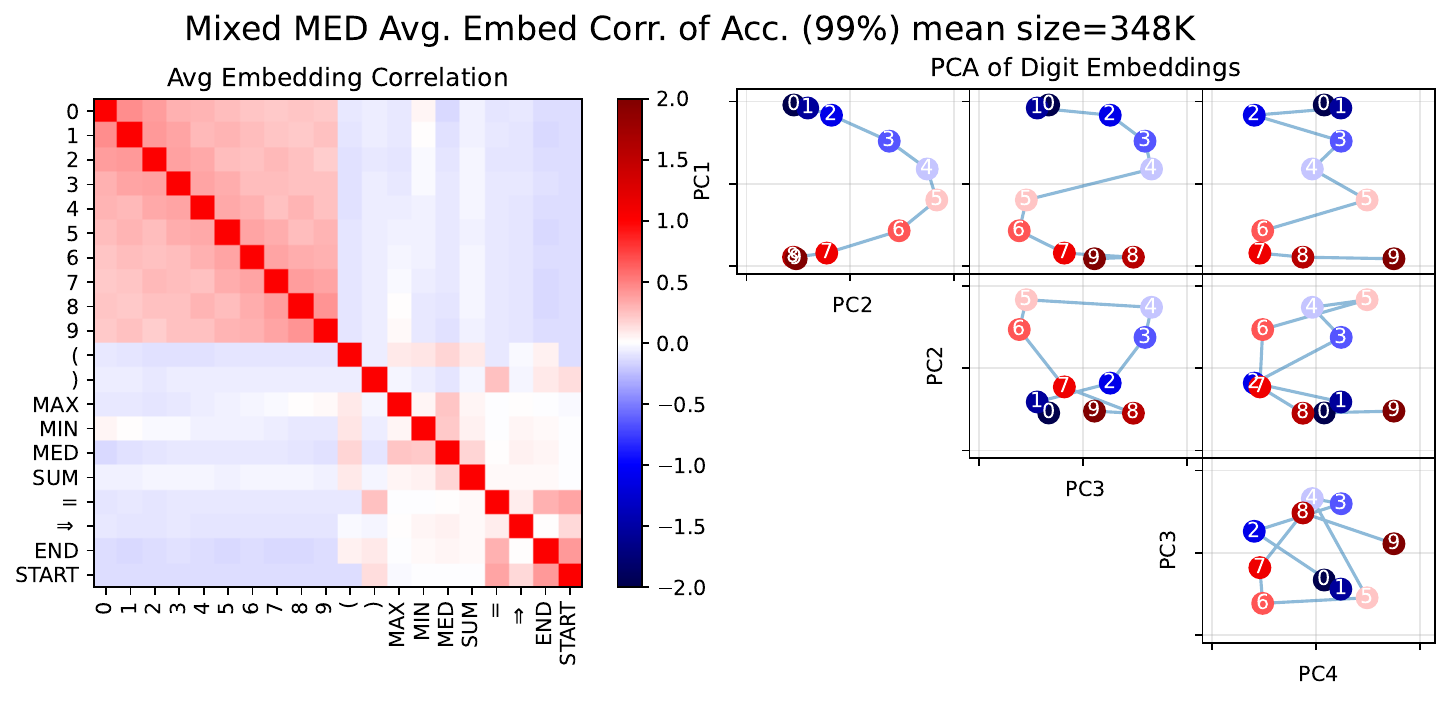}
    \includegraphics[width=0.49\linewidth]{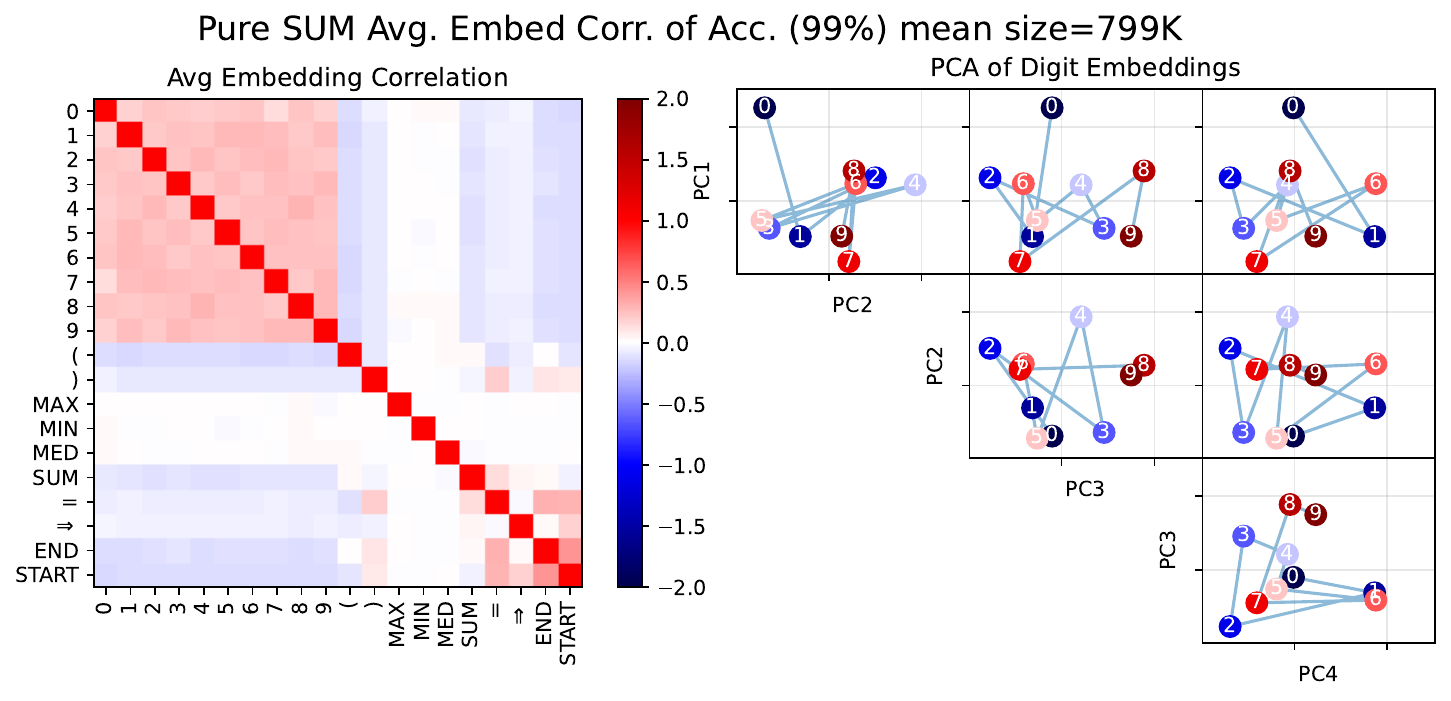}\includegraphics[width=0.49\linewidth]{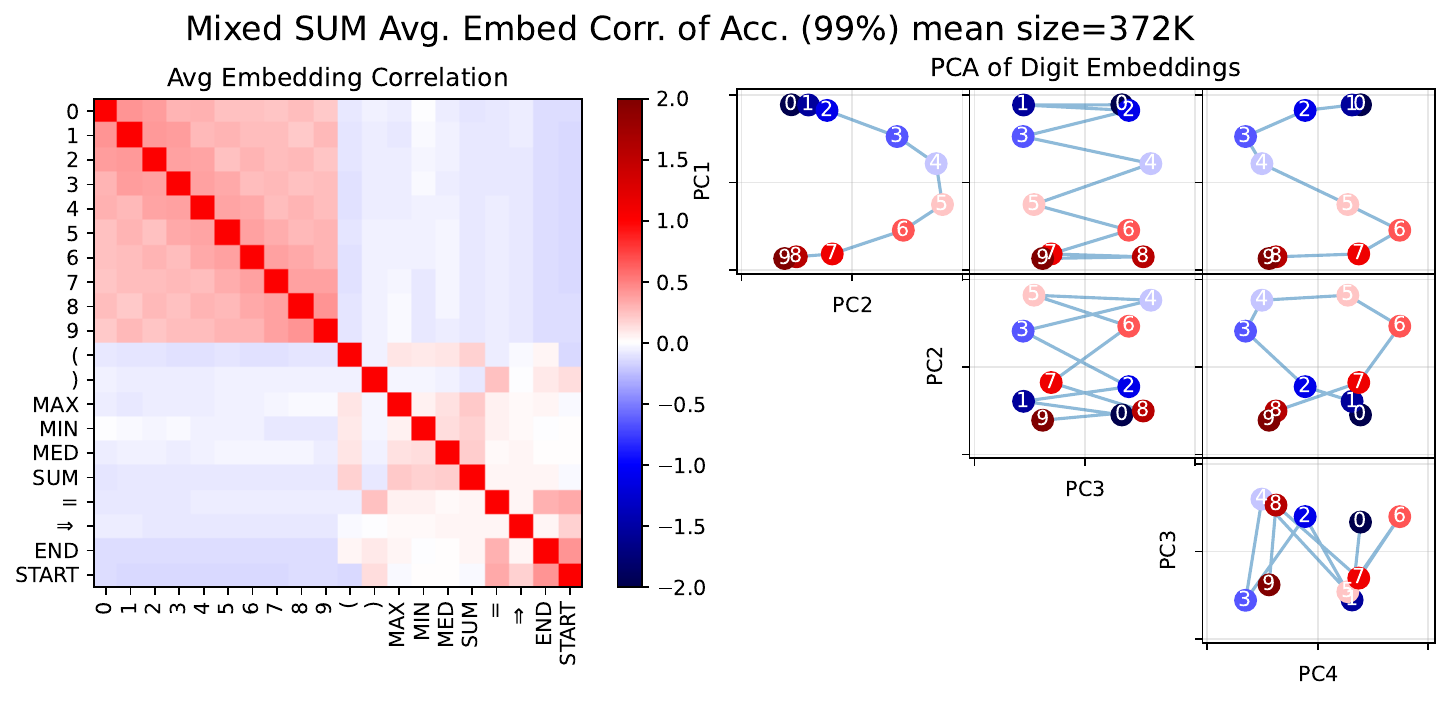}
    \caption{\textbf{PCA of embeddings with the same sample size (67):} 
    We choose the smallest 67 models which reached over $95$\% test accuracy. 
    Each row shows the average correlation matrix and top PCs for models trained on either a single operation, e.g. Pure+MAX, or all mixtures involving a given operation, e.g. Mixed+MAX. 
    We observe that the PCs get slightly distorted with the smaller sample size compared to fig. \ref{fig:PCA-pure-mixed}, but the overall structure stays the same. 
    Again, pure SUM does not show a discernible structure in the embeddings, whereas all cases do. 
    }
    \label{fig:PCA-pure-mixed-rand-samples}
\end{figure}


\begin{figure}
    \centering
    \includegraphics[width=1\linewidth]{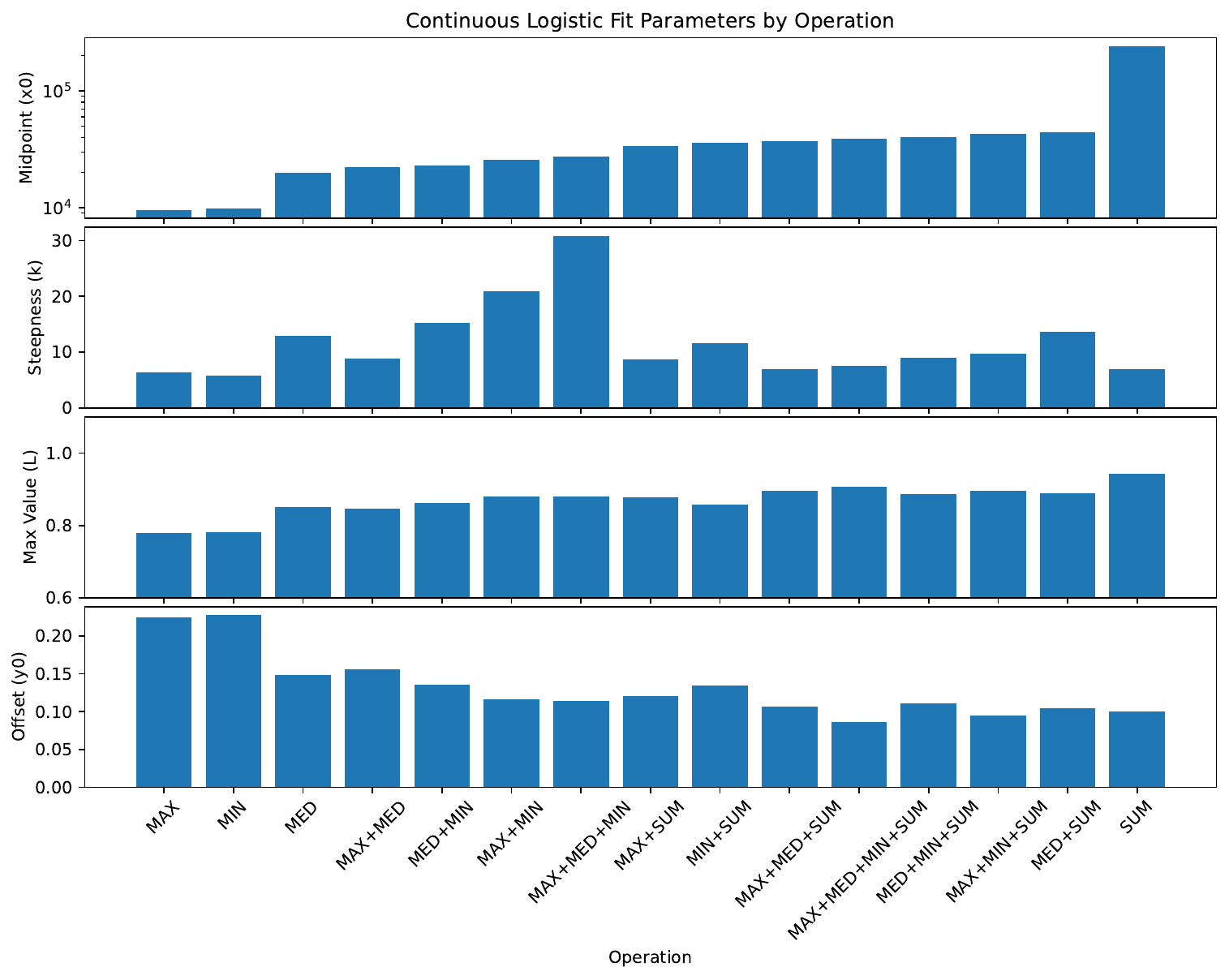}
    \caption{\textbf{Parameters of continuous logistic regression fitting accuracy to number of parameters (No Early stopping)}.
    }
    \label{fig:transition-logistic-params}
\end{figure}



\begin{figure}
    \centering
    \includegraphics[width=\linewidth]{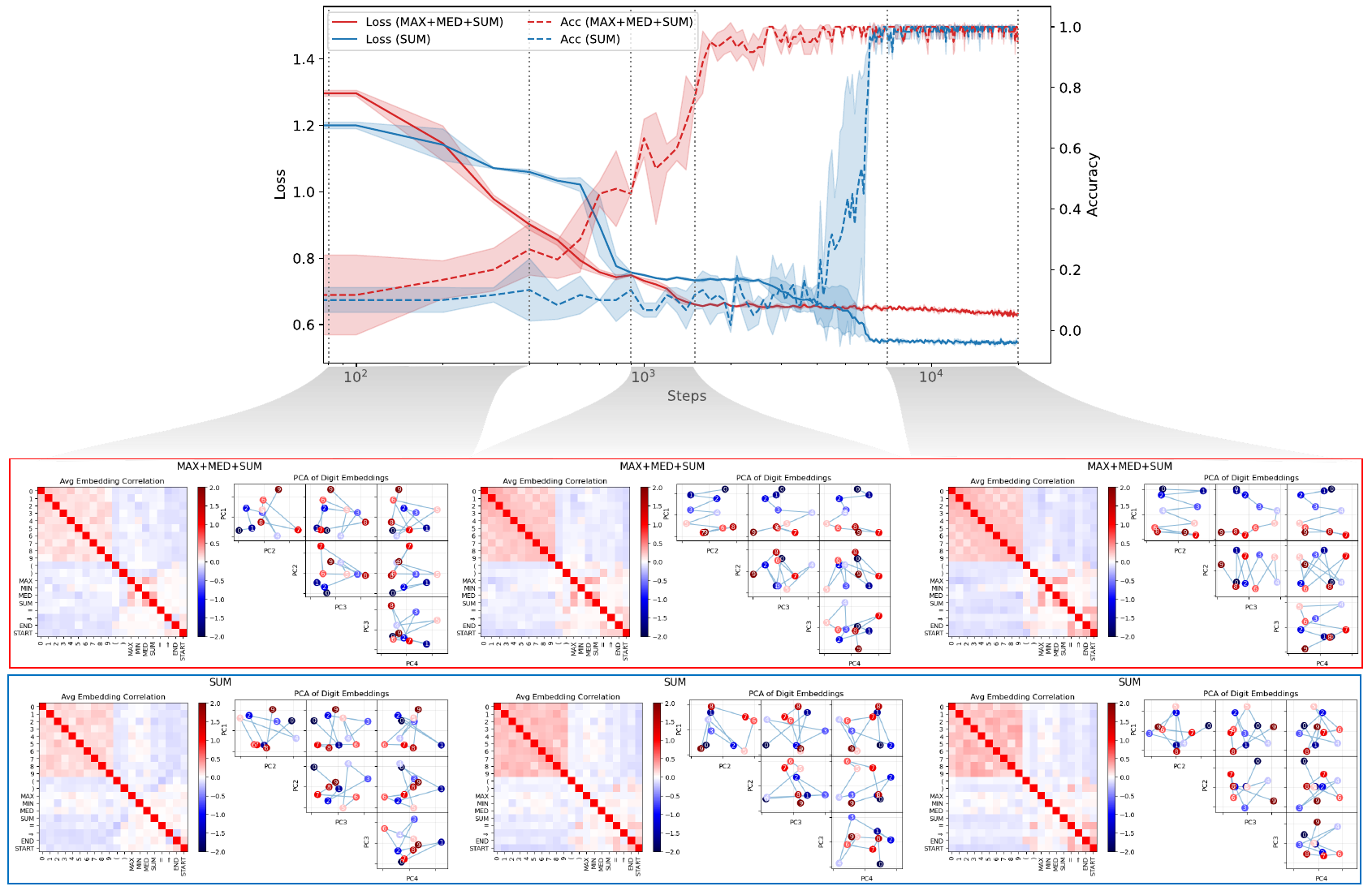}
    \caption{
    \textbf{Evolution of training loss, accuracy, and principal components of cosine similarities in the embedding layer. Modulo 10 ListOps.}
    The top main figure shows the evolution of training loss (solid lines) and test accuracy (dashed lines) for models with an embedding dimension of 128 and 3 layers, trained either on SUM-only data (blue) or on mixed MAX+MED+SUM data (red). Curves represent the mean across three independent runs, with shaded regions indicating one standard deviation. All models were trained for 20000 iterations. The red and blue boxes beneath the main plot display the average embedding representations at different training stages (indicated by vertical gray dashed lines). PCA reveals that models trained on MAX+MED+SUM data progressively develop a structured representation of numerical concepts, accompanied by a steady decrease in loss. In contrast, models trained solely on SUM data exhibit no clear structure in the embedding space and show long plateaus in the loss curve. This suggests that discovering an effective algorithm for the SUM operation in isolation requires significantly more exploration during training, in contrast to the more efficient joint training setting.
    }
    \label{fig:evo-loss-acc-emb}
\end{figure}


\clearpage

\subsection{Triplet Experiments}
\begin{figure}[!htbp]
    \centering
    \includegraphics[width=\linewidth]{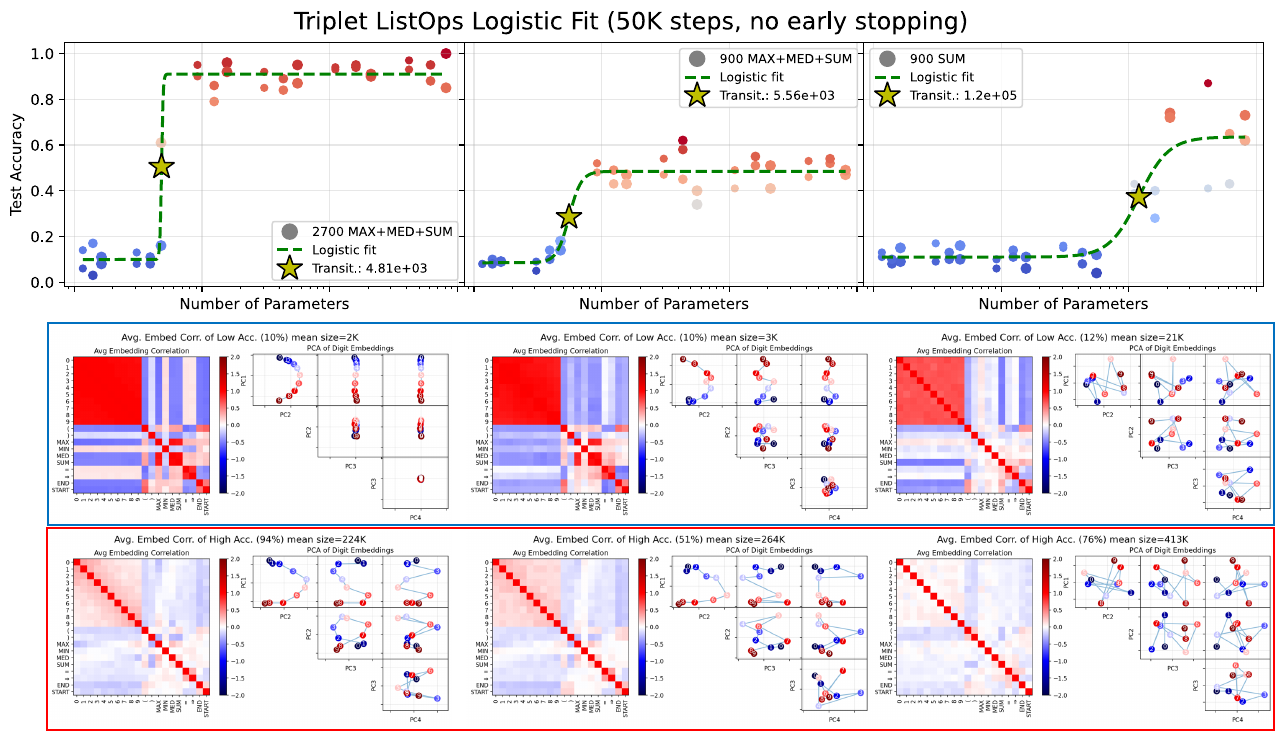}
    \caption{
    \textbf{Training only on triplet ListOps data.}
    The first row shows the same group of small transformer models trained on different variants of the triplet dataset.
    Each training set is constructed from 900 unique triplets. The 2700 MAX+MED+SUM dataset includes all 900 triplets, each labeled with three different operations. The 900 MAX+MED+SUM dataset contains 300 examples per operation, randomly sampled from the 2700 set. The 900 SUM dataset contains only the 900 unique SUM triplets.
    In all cases, the test set comprises the same held-out 100 SUM triplets.
    Red dots indicate models that exceed 50\% test accuracy, while blue dots denote models that fall below this threshold. The dashed green line represents a logistic fit, and the yellow star marks the transition point at 50\% accuracy. The x-axis shows the model size (in number of parameters), and subplots are ordered by increasing transition threshold. 
    The red and blue boxes beneath the main plot display the average PCA embedding representations at selected regions; red corresponds to high-accuracy models, while blue indicates low-accuracy ones. Models trained on the combined MAX+MED+SUM task develop a discernible structure during training, unlike models trained solely on SUM. Notably, augmenting MAX and MED data with only 300 SUM samples allows the model to achieve 50\% accuracy on a held-out 100 SUM sample test set, using a model 10× smaller than those trained on 900 SUM samples alone, which only reach $\approx80\%$ accuracy.
    }
    \label{fig:triplet-result}
\end{figure}

To increase control in our experimental setup, we train models only on triplet inputs \ref{fig:triplet-result}. 
We construct 1000 unique triplet samples, splitting them into 900 for training and 100 for testing.

For joint training on the MAX, MED, and SUM operations, we use a combined training set of 2700 triplets (900 per operation) and evaluate performance on the same 100 excluded SUM samples. 
For a fair comparison, we also create a balanced training set of 900 samples by selecting 300 examples from each of MAX, MED, and SUM.

With early stopping, models trained on the full 2700-sample mixed dataset achieve 100\% accuracy on the SUM test set, while those trained on only 900 SUM samples reach approximately 50\% accuracy. 
When trained for 50k steps without early stopping, models eventually learn the task, but SUM requires significantly more parameters—consistent with earlier observations. 
These results highlight that learning the SUM operation in isolation is more challenging and slower (Fig. \ref{fig:triplet-result}). 

\clearpage

\clearpage
\section{Observations from the norm of Attention and Feedforward outputs}
\label{ap:attn-ffwd}


To investigate the internal dynamics of our models, we focused on the final layer of a 3-layer transformer network, featuring a single attention head and an embedding dimension of 128. 
Our analysis centered on comparing the behavior of models trained on ''ALL3'' operations (MAX+MED+SUM) versus those trained solely on the ''SUM'' operation.

We introduced a novel metric to quantify the impact of different components within the network: 
the ratio of output to input norms for both the self-attention (SA) and feedforward (FFN) sublayers. Specifically, we computed:
\begin{enumerate}
    \item Attention ratio: $r_{attn} = \frac{\|SA(LN_1(x))\|}{\|x\|}$
    \item Feedforward ratio: $r_{ffwd} = \frac{\|FFN(LN_2(x_1))\|}{\|x_1\|}$
\end{enumerate}
where $LN_1$ and $LN_2$ are layer normalization operations, and $x_1$ is the output of the self-attention sublayer. 
These ratios provide insight into how much each component modifies its input, serving as a proxy for the component's impact on the overall computation.

We analyzed the distribution of these ratios across a test set consisting of sum operations for both the 'ALL3' and 'SUM' models. 
Kernel Density Estimation (KDE) plots were used to visualize the distributions, and we employed several statistical measures to quantify the differences.

\paragraph{Attention Sublayer}

The attention sublayer showed moderate but statistically significant differences between the 'ALL3' and 'SUM' models:
\begin{itemize}
    \item Kolmogorov-Smirnov test: statistic = 0.1592, p-value < 0.0001
    \item Jensen-Shannon divergence: 0.1591
    \item Wasserstein distance: 0.0696
    \item Effect size (Cohen's d): 0.1761
    \item 95\% CI for mean difference: (0.0466, 0.0860)
\end{itemize}

The KDE plot revealed that the 'ALL3' model's attention ratio distribution was more concentrated and peaked higher than the 'SUM' model's distribution. 
The positive effect size and confidence interval indicate that the 'ALL3' model generally had higher attention ratios.

\paragraph{Feedforward Sublayer}
The feedforward sublayer exhibited more pronounced differences:
\begin{itemize}
    \item Kolmogorov-Smirnov test: statistic = 0.2461, p-value < 0.0001
    \item Jensen-Shannon divergence: 0.1617
    \item Wasserstein distance: 0.2830
    \item Effect size (Cohen's d): -0.3379
    \item 95\% CI for mean difference: (-0.3042, -0.2226)
\end{itemize}

The KDE plot for the feedforward ratios showed a clear shift between the two distributions. 
The 'SUM' model's distribution was shifted towards higher values, as confirmed by the negative effect size and confidence interval.

\paragraph{Interpretation}
These results reveal distinct operational patterns between models trained on 'ALL3' operations versus those trained solely on 'SUM':
\begin{enumerate}
    \item In the attention sublayer, the 'ALL3' model shows slightly higher ratio values, \textbf{suggesting that attention mechanisms play a more pronounced role when the model is trained on diverse operations.} 
    This could indicate that the attention sublayer is capturing more complex patterns or relationships necessary for handling multiple operations.
    \item Conversely, in the feedforward sublayer, the 'SUM' model demonstrates significantly higher ratio values. 
    This \textbf{suggests that when trained on 'Sum' alone, the model relies more heavily on the feedforward network for computation.} 
    This could imply that the 'SUM' operation is being implemented more directly through feedforward transformations.
    \item The larger effect size in the feedforward layer (-0.3379) compared to the attention layer (0.1761) indicates that the difference in behavior is more pronounced in the feedforward component.
\end{enumerate}

These observations suggest a trade-off in how the network allocates its computational resources. 
The 'ALL3' model appears to leverage its attention mechanism more, potentially to handle the diversity of operations it was trained on. 
In contrast, the 'SUM' model seems to channel more of its computation through the feedforward network, possibly developing a more specialized but less flexible approach to solving the sum operation.

This analysis provides evidence that the internal dynamics of transformer models adapt significantly based on the diversity of tasks they are trained on, even when evaluated on the same type of operation ('SUM'). 
It highlights the importance of considering task diversity in understanding and optimizing neural network architectures.

\begin{figure}[!htbp]
    \centering
    \includegraphics[width= 0.49\linewidth]{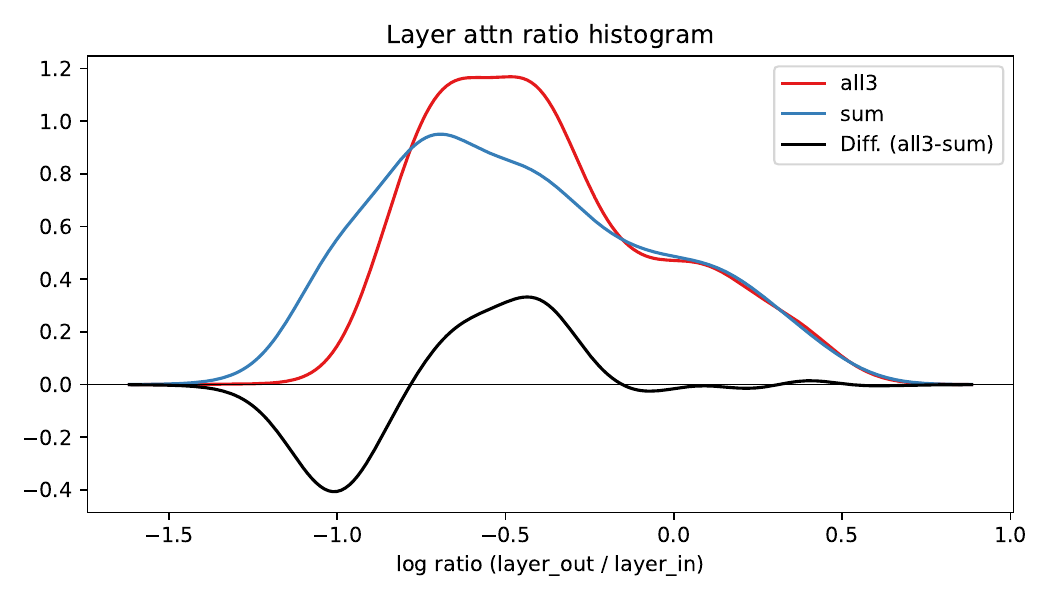}
    \includegraphics[width= 0.49\linewidth]{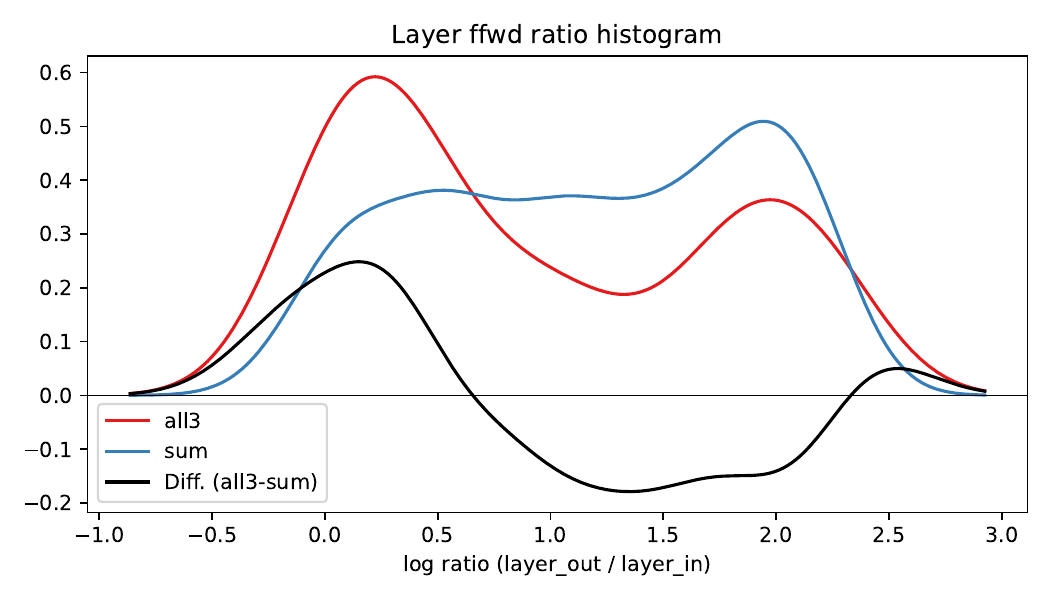}
    \caption{\textbf{Comparison of layer output/input ratio distributions} for models trained on all three operations (all3 - MAX+MED+SUM) versus sum operation alone (SUM). 
    \textbf{Left:} Attention layer ratio histogram. 
    \textbf{Right:} Feedforward layer ratio histogram. The x-axis represents the log ratio of layer output norm to input norm, while the y-axis shows the density. The black line represents the difference between the all3 and sum distributions (all3 - sum). These plots illustrate distinct operational patterns between the two models, with the attention layer showing increased activity in the all3 model and the feedforward layer demonstrating higher ratios in the sum model.}
    \label{fig:stats-layer3-all3-sum0}
\end{figure}

\begin{figure}
    \centering
    \includegraphics[width=\linewidth, 
    ]{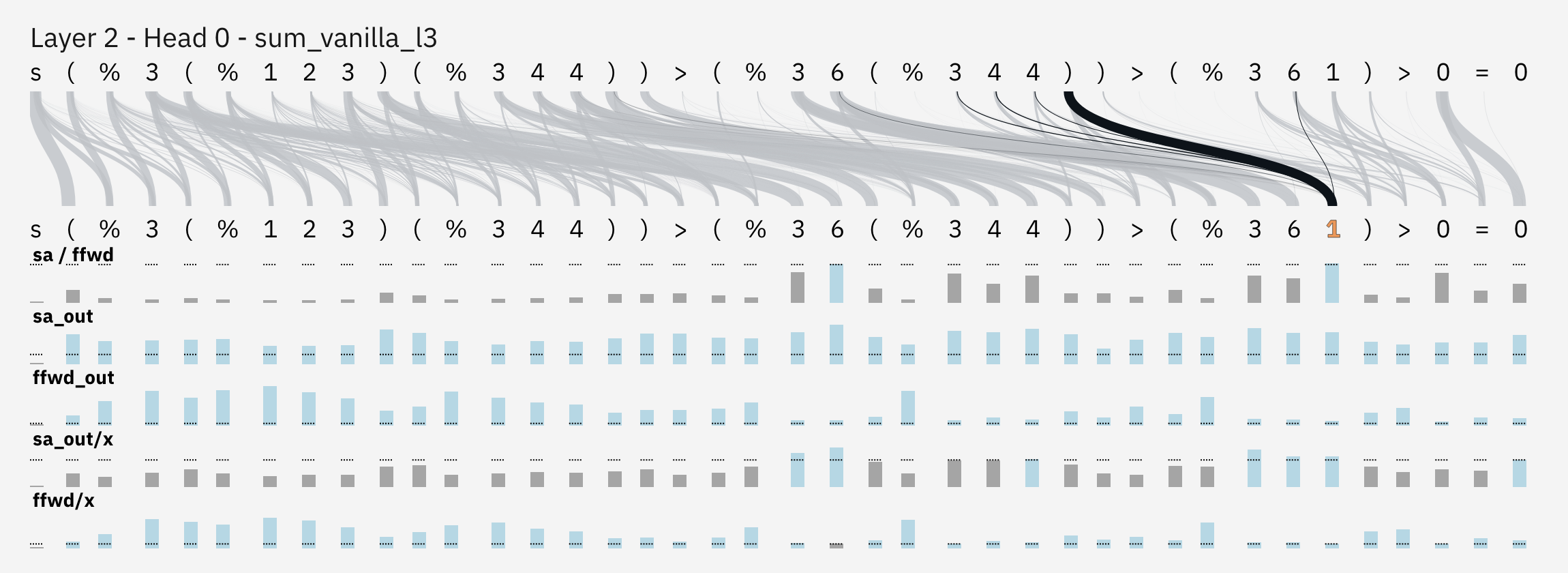}
    \includegraphics[width=\linewidth, 
    ]{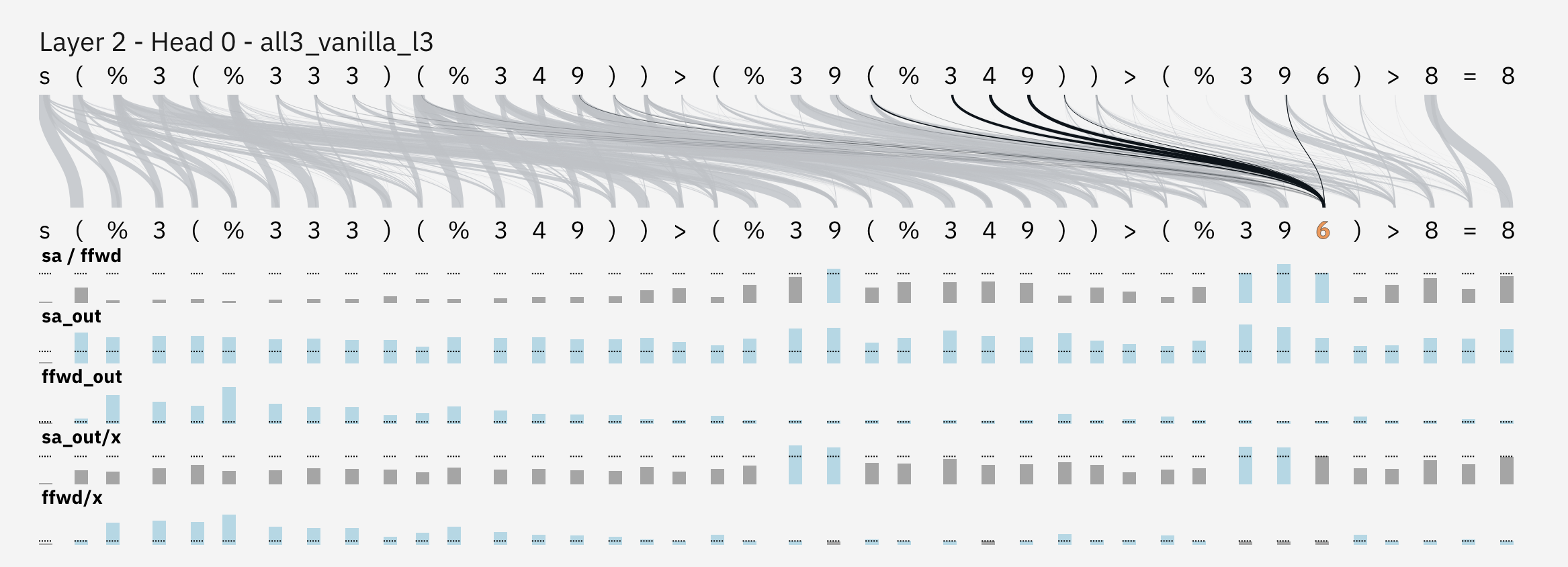}
    \caption{\textbf{Attention patterns and layer dynamics in SUM vs MAX+MED+SUM (all3) models.} 
    Each panel shows (CoT) solution to a SUM modulo 10 problem, where '>' indicates solution steps. 
    The first row shows the input sequence, with curved lines representing attention weights from Layer 2 in a 3-layer network. 
    Black lines highlight attention patterns for a specific digit (shown in orange). 
    Below are shown various layer metrics including the ratio of self-attention to feedforward norms (\texttt{sa/ffwd}), self-attention output norms (\texttt{sa\_out}), feedforward output norms (\texttt{ffwd\_out}), and ratios of layer outputs to inputs (\texttt{sa\_out/x, ffwd/x}). 
    \textbf{Top:} Model trained only on SUM operations shows attention primarily focused on parentheses and structural elements. 
    \textbf{Bottom:} Model trained on MAX+MED+SUM (all3) shows attention strongly connecting to digits being combined in each CoT step, suggesting direct involvement in numerical computation. 
    These distinct patterns suggest fundamentally different algorithms learned by each model.}
    \label{fig:attn-ffwd}
\end{figure}

    \label{fig:stats-layer3-all3-sum}

\begin{figure}[h]
    \centering
    \includegraphics[width=.9\linewidth]{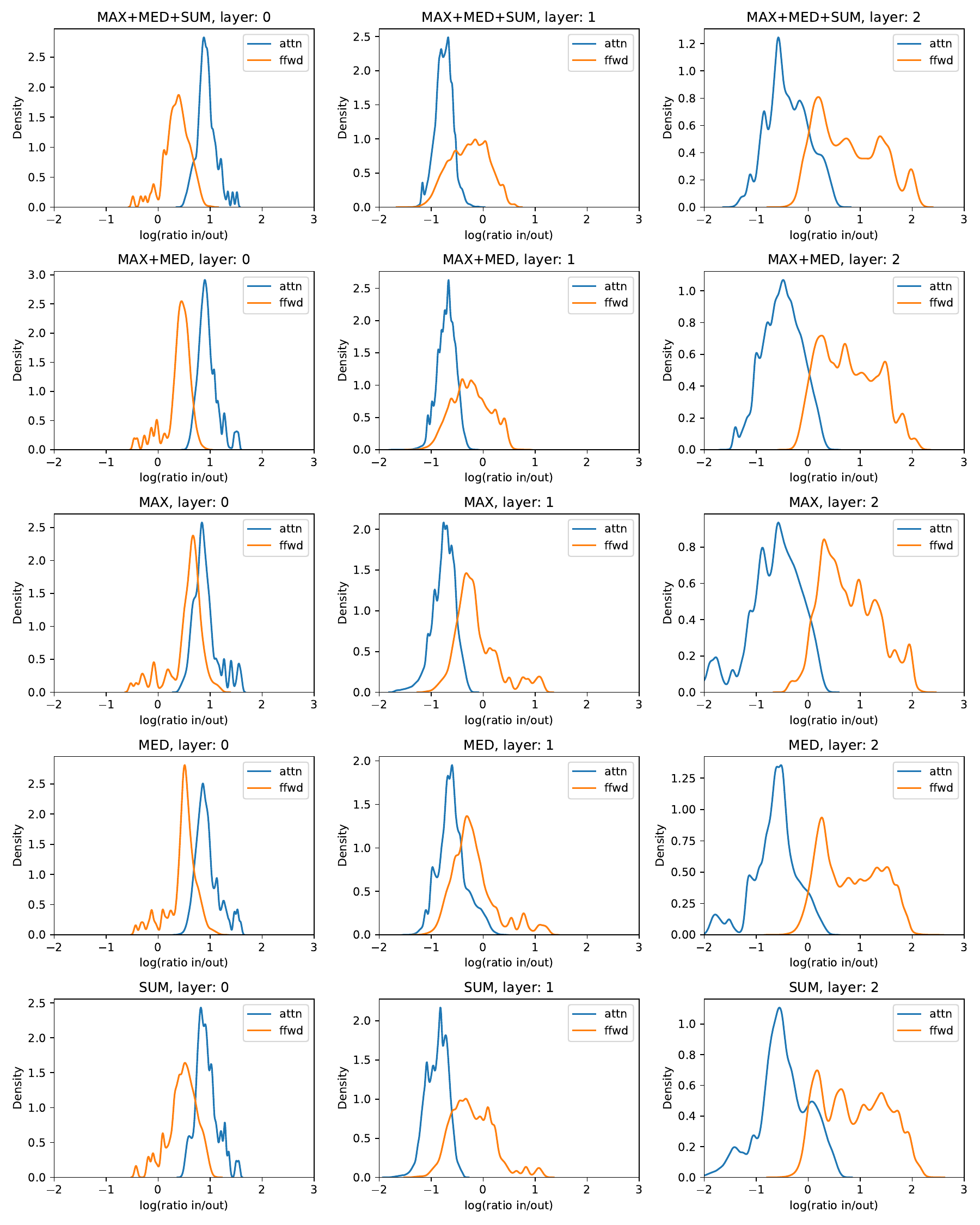}
    \caption{\textbf{Attention layer and Feedforward layer ratio histogram.} 
    Each row shows the attention and feedforward layer ratio histogram for models trained on MAX, MED,  SUM, MAX+MED,and MAX+MED+SUM. 
    Each column shows the ratio histogram in different attention blocks. The model had 128 embedding and 3 layers and all models reaches 99\% accuracy.
    }
    \label{fig:all_sa_ffwd}
\end{figure}

\begin{figure}[h]
    \centering
    \includegraphics[width=1\linewidth]{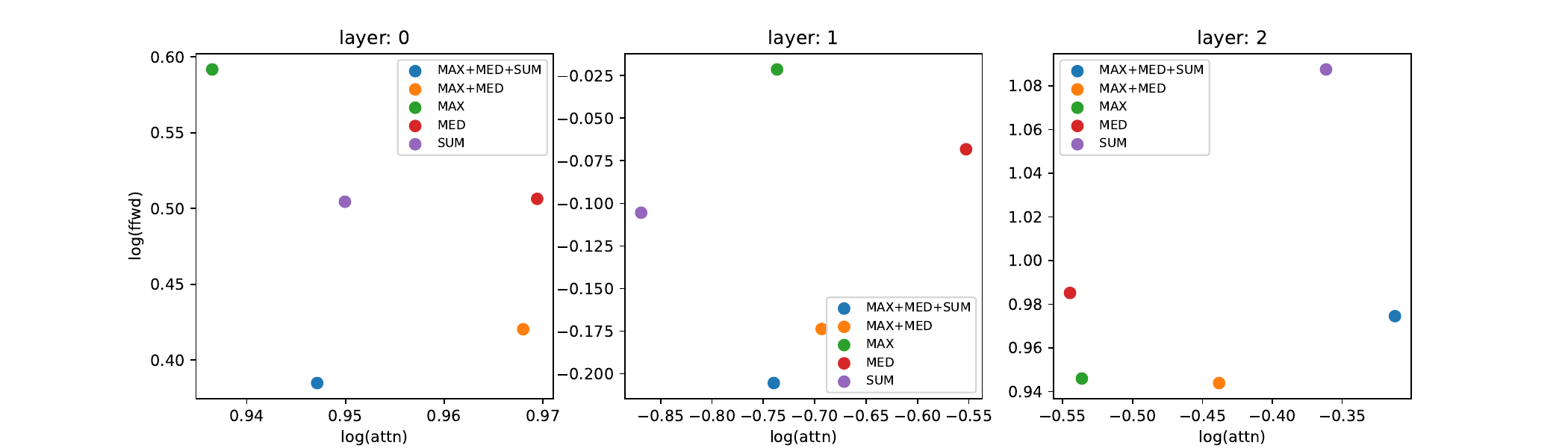}
    \caption{\textbf{The mean attention layer ratio vs the feedforward layer ratio.}  
    Each plots show the means in different layers. 
    }
    \label{fig:all_sa_ffwd_mean}
\end{figure}

\begin{figure}[h]
    \centering
    \includegraphics[width=1\linewidth]{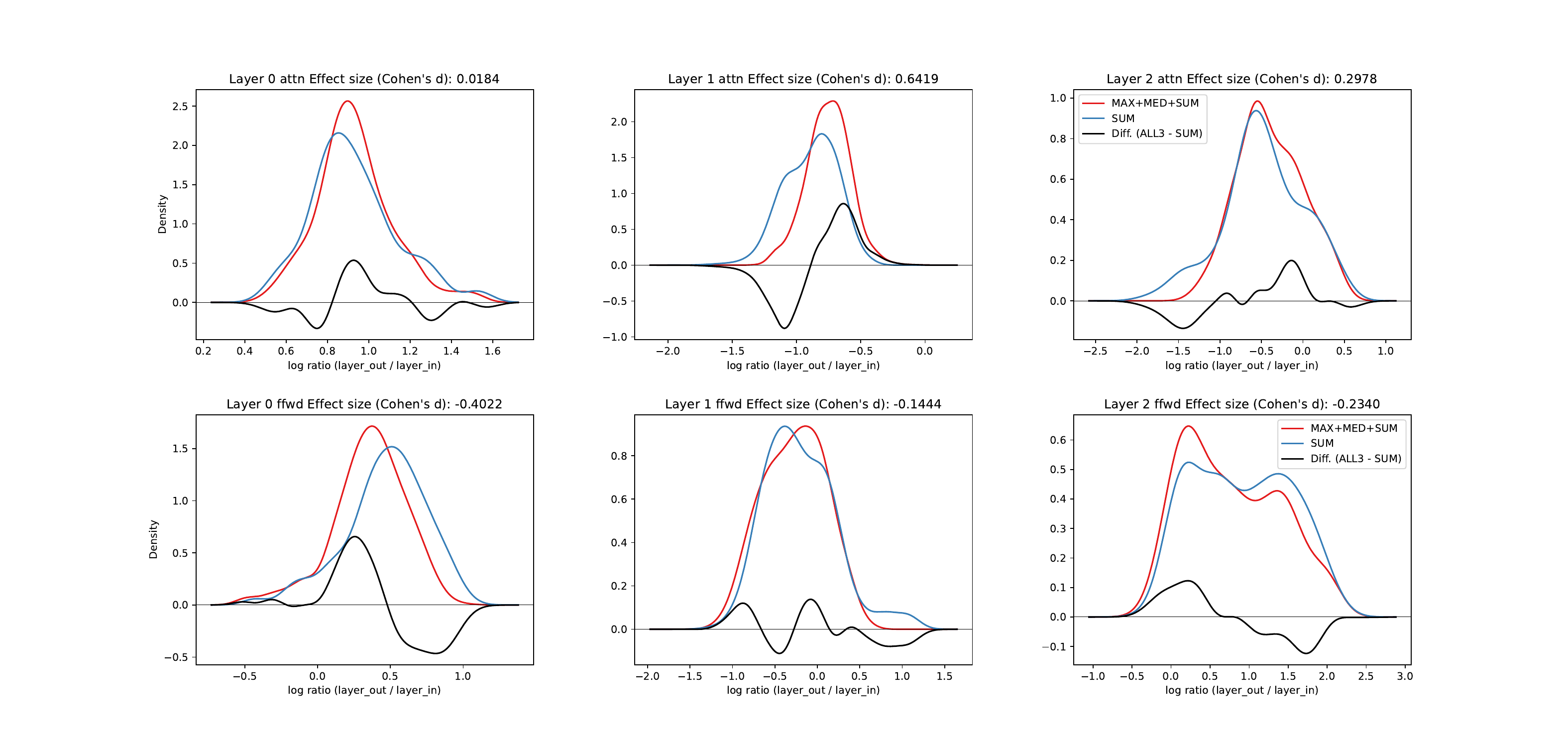}
    \caption{\textbf{Comparison of layer output/input ratio distributions} for models trained on all three operations (MAX+MED+SUM) versus sum operation alone (SUM). 
    The title of the figures contain the Effective size showing the difference between the model.
    }
    \label{fig:dif_all_sa_ffwd}
\end{figure}

\section{Ablation Studies}
\label{app:ablations}

\subsection{Number of Attention Heads}
Our main experiments use a single attention head for simplicity and interpretability.
To verify that our findings are not an artifact of this choice, we conducted a systematic sweep comparing 1 vs.\ 4 attention heads across the same range of embedding dimensions and task combinations.
We found that the qualitative pattern of results---including the relative ordering of task difficulty (ADD hardest, MAX/MIN easiest) and the effectiveness of specific task pairings (easy+hard helps, hard+hard does not)---remains unchanged across head configurations.
The transition points shift slightly but the relative benefits of joint training (2--7$\times$ reduction in model size for successful pairings) persist.
This suggests that our findings about task compatibility reflect properties of the tasks themselves rather than artifacts of the single-head architecture.

\subsection{Deep vs. Recurrent Architecture}
Our primary architecture uses a recurrent transformer: a single transformer block applied iteratively, feeding its output back as input for the next iteration.
We compared this to standard deep transformers with 2--4 stacked layers and found qualitatively similar results for the task difficulty hierarchy and joint training benefits (see Section~\ref{resutls_deep_transformer}).
The recurrent architecture was chosen for two reasons: (1) it converges faster during training, and (2) it produces more structured, interpretable embeddings that facilitate our PCA analysis.
The key finding---that joint training with compatible tasks reduces model size requirements---holds for both architectures.

\section{Additional Scaling Results}
\label{app:scaling}

\begin{figure}[h]
    \centering
    \includegraphics[width=\linewidth]{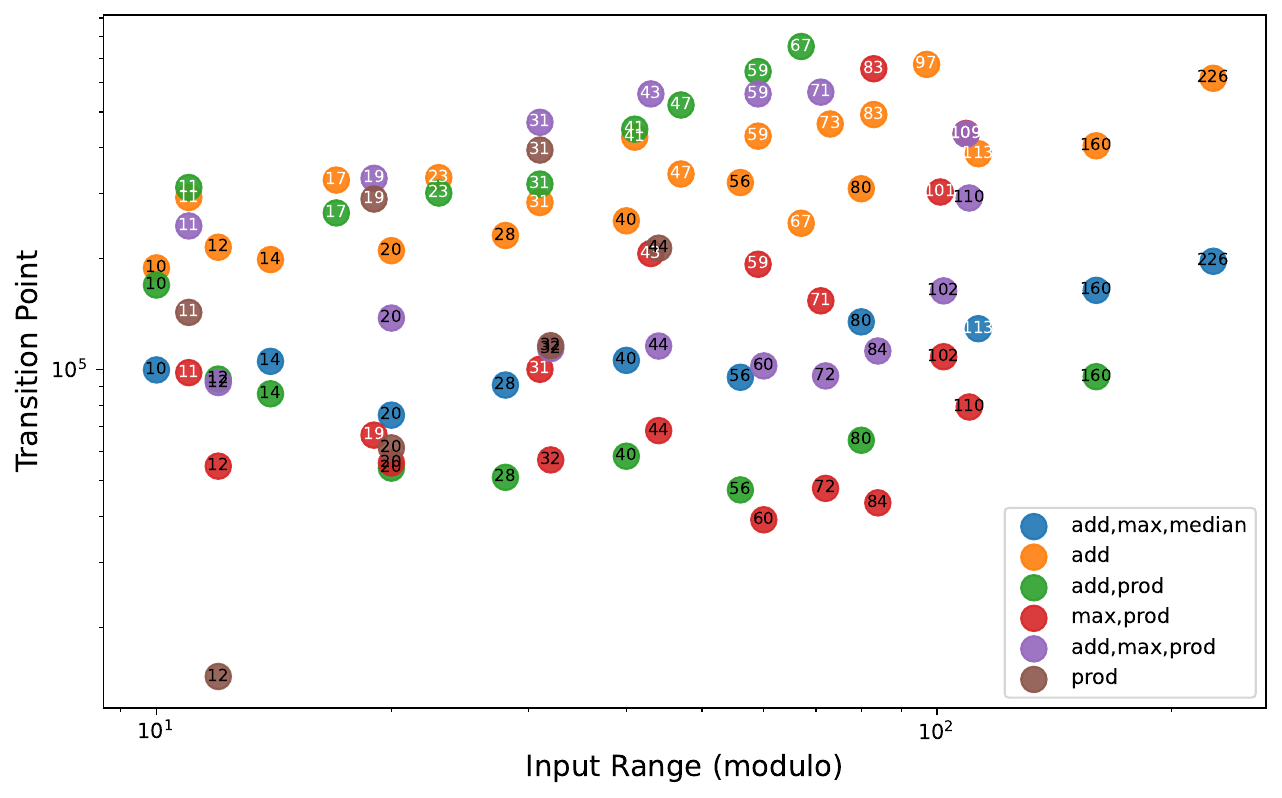}
    \caption{
    \textbf{Full scaling results across moduli.}
    Transition points (minimum model size to learn) as a function of modulo $n$ for various task combinations.
    Prime moduli are shown in white text.
    Key observations: (1) PROD-prime is consistently hard; (2) ADD+PROD helps ADD for non-prime moduli; (3) MAX+PROD helps PROD even for primes; (4) ADD+MAX+PROD at primes can be harder than individual tasks.
    }
    \label{fig:scaling-full}
\end{figure}

Figure~\ref{fig:scaling-full} shows the full scaling results across all task combinations tested, complementing the selected comparisons in the main text.

\section{Additional Curriculum Experiments}
\label{app:curriculum}

\begin{figure}[h]
    \centering
    \includegraphics[width=\linewidth]{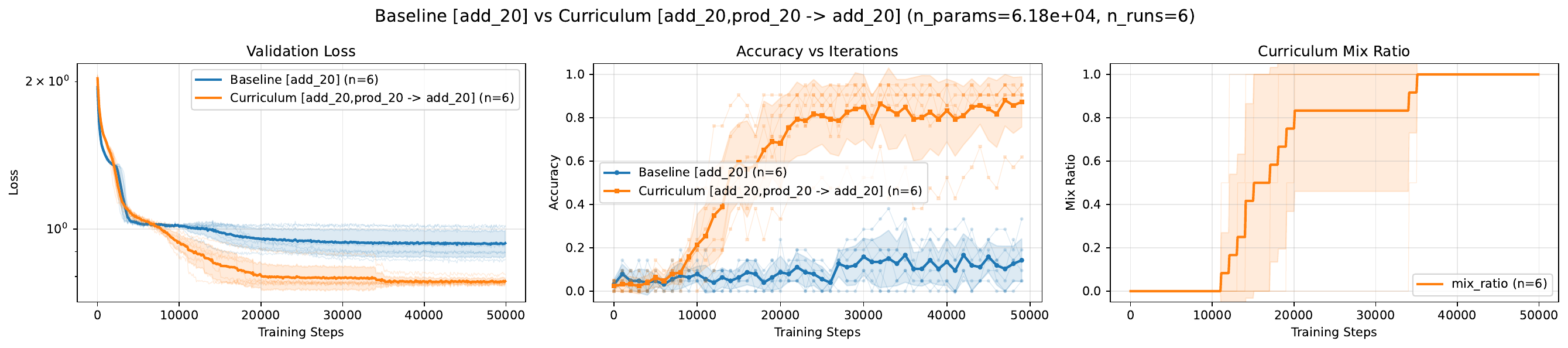}
    \caption{
    \textbf{Full curriculum comparison: ADD+PROD $\to$ ADD.}
    Left: Validation loss over training. Middle: Accuracy over training---baseline ADD (blue) stays at $\sim$15\%, curriculum (orange) reaches near-perfect accuracy. Right: Curriculum mix ratio showing transition from ADD+PROD to pure ADD.
    }
    \label{fig:curriculum-full}
\end{figure}

Figure~\ref{fig:curriculum-full} shows the full curriculum training results including loss, accuracy, and mix ratio curves comparing baseline ADD training with the ADD+PROD $\to$ ADD curriculum.

In addition, we tested transfer from MAX+MED to ADD.

\paragraph{MAX+MED $\to$ ADD transfer.}
We trained a small model ($n_{\text{embed}}=48$, below the pure ADD transition) on MAX+MED until proficiency, then gradually transitioned to pure ADD (never showing mixed MAX+MED+ADD expressions).
Key findings:
\begin{itemize}
    \item The model learns ADD immediately upon introduction, achieving high accuracy.
    \item The model forgets MAX and MED as they are phased out, but retains the structured embedding.
    \item Models as small as $n_{\text{embed}}=24$ can learn ADD this way---7$\times$ smaller than the pure ADD transition.
\end{itemize}

This confirms that the curriculum learning effect is robust across different task combinations: structured representations learned from easy tasks enable efficient learning of hard tasks.

\begin{figure}[h]
    \centering
    \includegraphics[width=.8\linewidth]{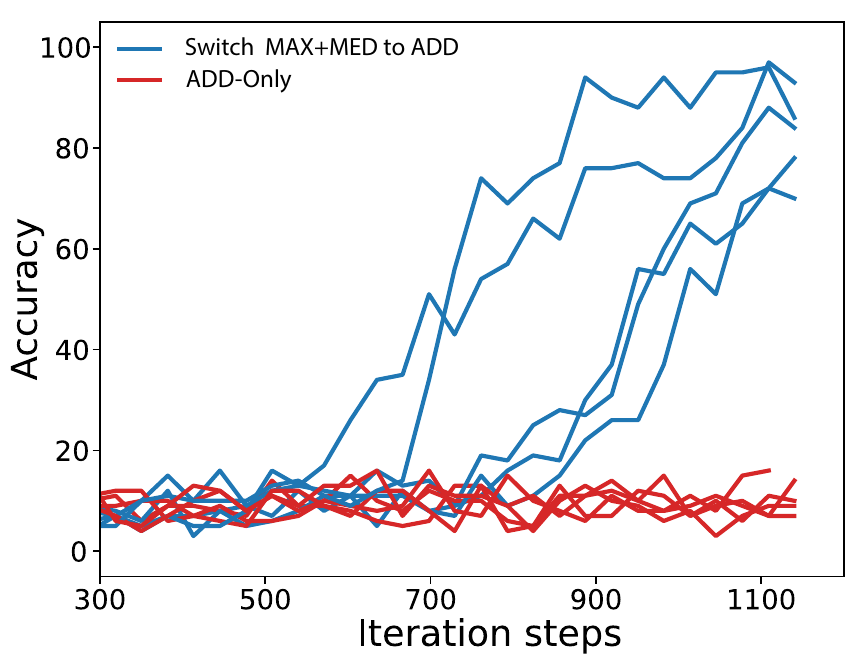}
    \caption{
    \textbf{Transfer from MAX+MED to ADD.}
    A model too small to learn pure ADD ($n_{\text{embed}}=48$) successfully learns ADD after pretraining on MAX+MED.
    Blue: curriculum (MAX+MED $\to$ ADD) reaches high accuracy.
    Red: baseline (pure ADD) fails to learn.
    }
    \label{fig:curriculum-maxmed}
\end{figure}





\clearpage
\section{Complexity: Number of operands and nesting level.}

\begin{figure}[!htbp]
    \centering
    \includegraphics[width=1\linewidth]{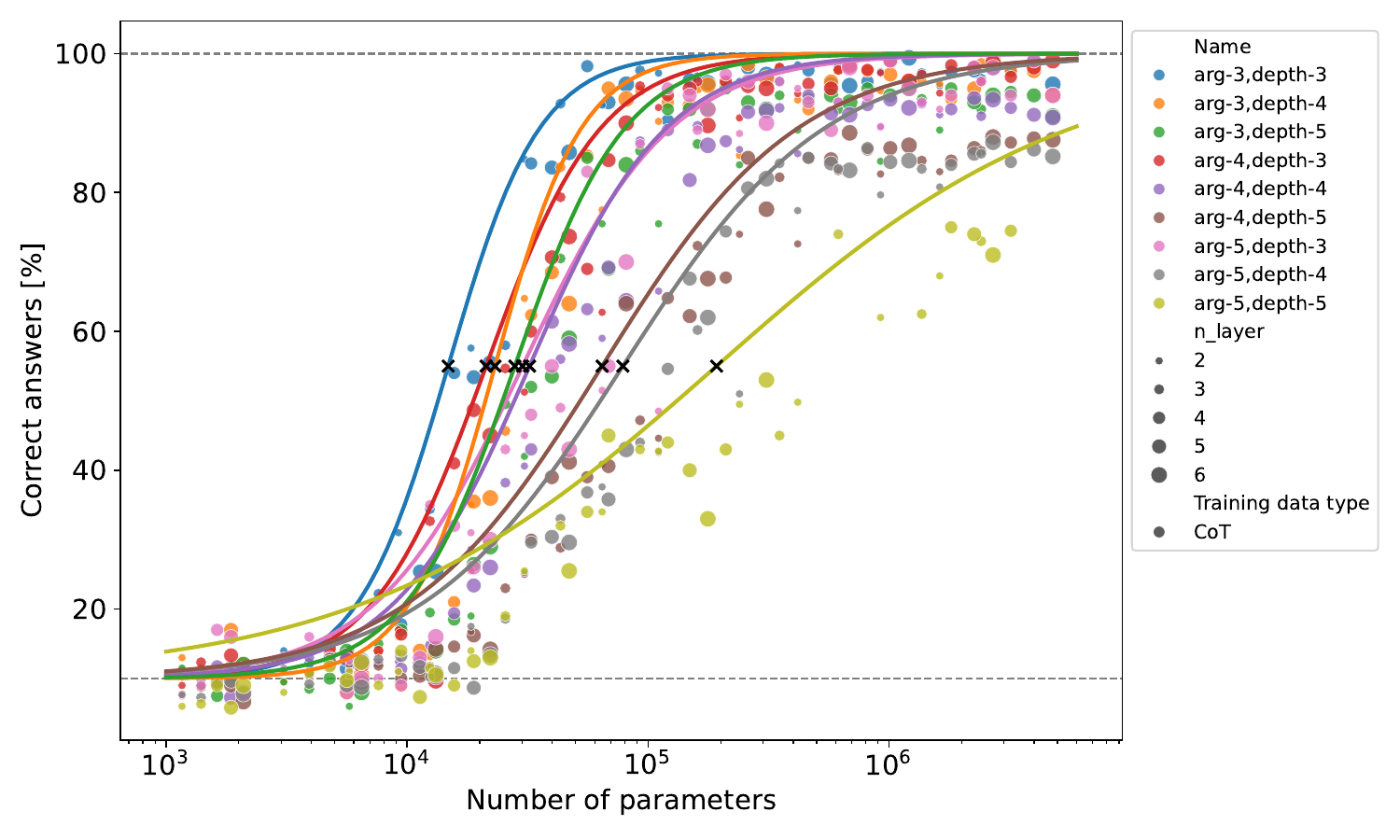}
    \caption{\textbf{Learning MAX+MED+SUM operations with varying numbers of operands and nesting levels.} The model requires more parameters as the number of operands and nesting levels increases. Higher nesting levels particularly demand larger model sizes to learn the task. We present the average of five simulations for each configuration and fit a sigmoid function, with the cross marking the middle value (transition point). The transition points reveal an interesting pattern: the sum of the number of operands and nesting levels groups together. For example, the transition points for arg-3,depth-4 (orange) and arg-4,depth-3 (red) are close to each other, as are those for arg-4,depth-5 (brown), and arg-5,depth-4 (grey). Early stopping criteria were
applied during training in all simulations.} 
    \label{fig:all3_arg_nest}
\end{figure}

\begin{figure}[htbp]
    \centering
    \begin{subfigure}[b]{0.45\textwidth}
        \centering
        \includegraphics[width=\textwidth]{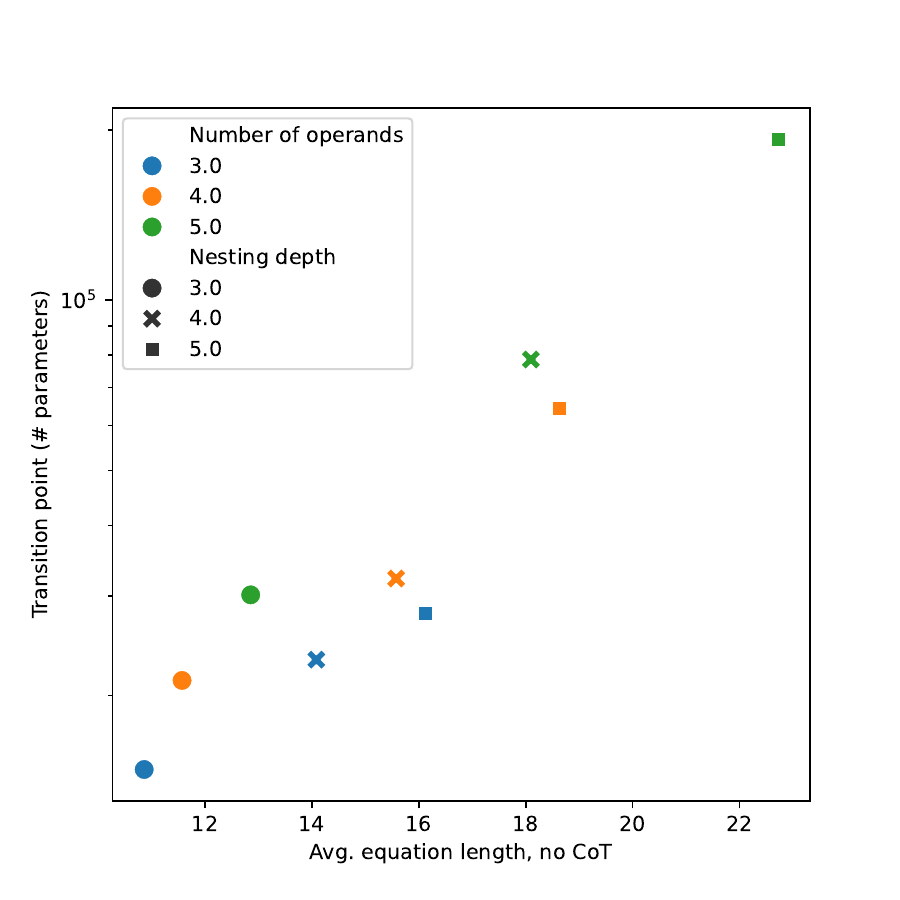}  
        \caption{\textbf{Average equation length vs. transition point.}}
        \label{fig:all3_avg_length}
    \end{subfigure}
    \hfill
    \begin{subfigure}[b]{0.45\textwidth}
        \centering
        \includegraphics[width=\textwidth]{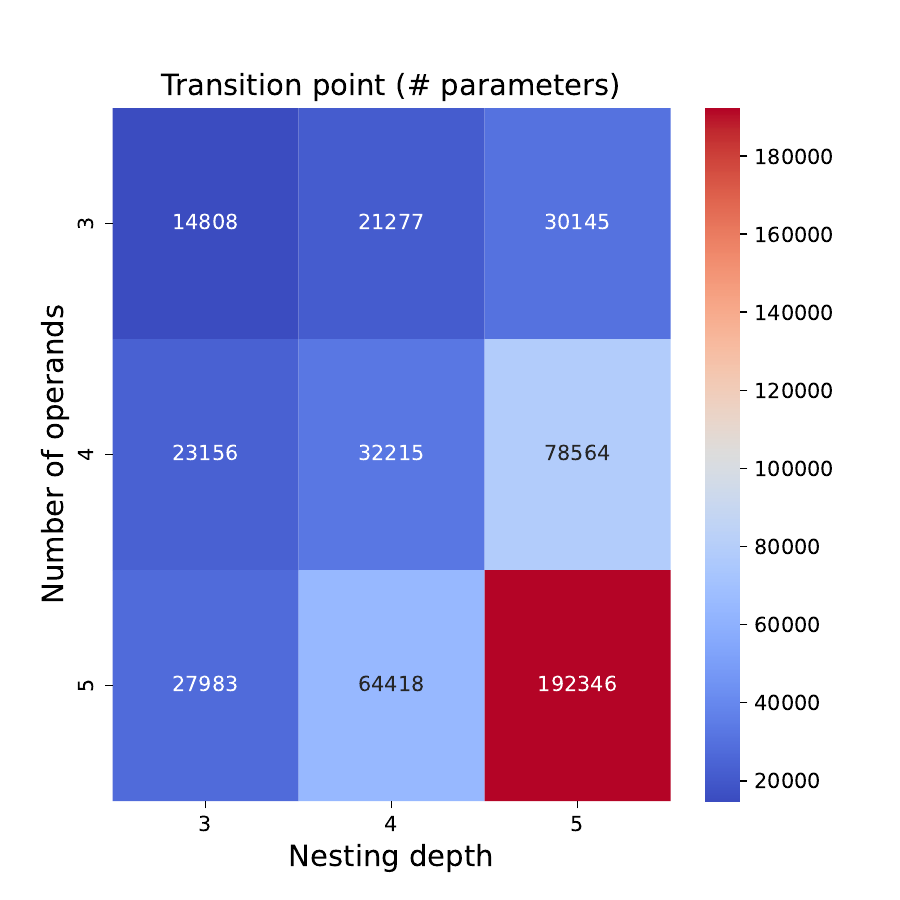}  
        \caption{\textbf{Transition points.}}
        \label{fig:all3_length}
    \end{subfigure}
    \caption{\textbf{Transition point vs. Equation Lenght and Nesting Depth.} (a) Transition point in function of the average equation length. (b) Heat plot of transition point in function number of operands and nesting depth. }
    \label{fig:transition_point_length}
\end{figure}

In previous studies, task complexity has been characterized through various measures, including the number of bits required to memorize the task, which corresponds to the length of the expression \cite{dave2024investigating}, the number of operands and nesting depth \cite{petruzzellis2024benchmarking}, and the structure of the computational graph \cite{dziri2023faith}, which captures the number of nestings. 
Here, we examine how the number of operands and nesting levels affect the learning ability of small language models. Focusing on the \verb|all3| task, which combines \verb|max|, \verb|med|, and \verb|sum| operations, we manipulate complexity by varying the number of operands (\verb|arg = {3, 4, 5}|) and nesting depth (\verb|depth = {3, 4, 5}|). 
The length of equations, measured by the number of characters, depends on both the number of operands and the nesting level. 
We find that nesting has a greater impact: increasing the nesting level from 3 to 4 results in longer equations than increasing the number of operands from 3 to 5 at a fixed nesting level. 
We train the GPT model on the \verb|all3| dataset with all combinations of \verb|arg| (number of operands) and \verb|depth| (nesting levels), finding that the model's performance correlates with the sum of operands and nesting levels (\verb|arg + depth|). 
Notably, transition points tend to group together for configurations with the same sum (Fig. \ref{fig:all3_arg_nest}, \ref{fig:transition_point_length}b). 
While Fig. \ref{fig:transition_point_length}a demonstrates that the model requires more parameters to solve longer equations, it also indicates that \verb|arg + depth| serves as a reliable predictor of the transition point. 

\clearpage
\section{Clustering in the embedding space}
\label{sec:clustering_PC}

\begin{figure}[htbp]
    \centering
    \includegraphics[width=1\linewidth]{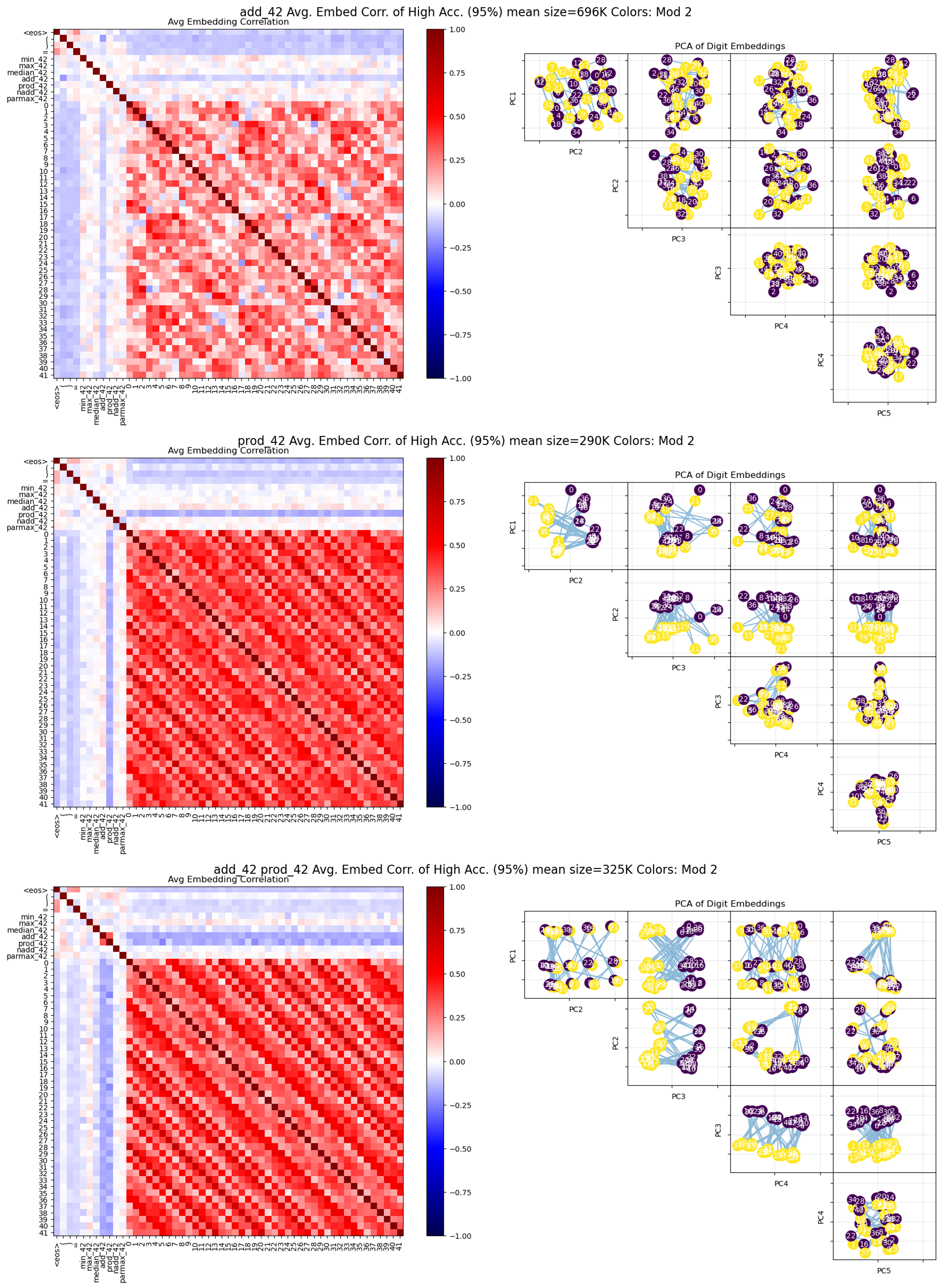}
    \caption{\textbf{ADD, PROD and ADD+PRD in MOD 42 PC embeddings}} 
    \label{fig:42_emb}
\end{figure}

\begin{figure}[htbp]
    \centering
    \includegraphics[width=1\linewidth]{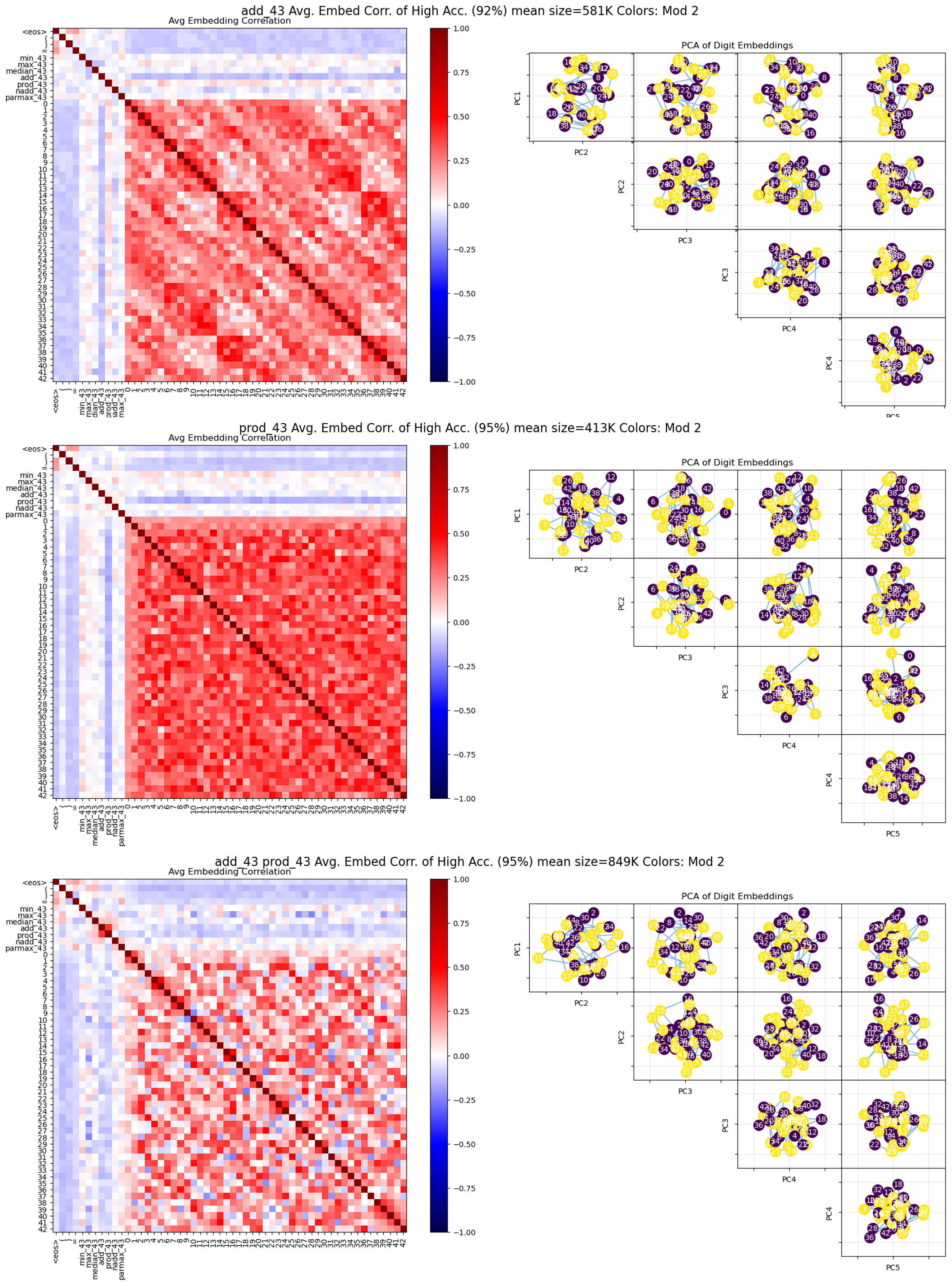}
    \caption{\textbf{ADD, PROD and ADD+PRD in MOD 43 PC embeddings}} 
    \label{fig:43_emb}
\end{figure}

\begin{figure}[htbp]
    \centering
    \includegraphics[width=1\linewidth]{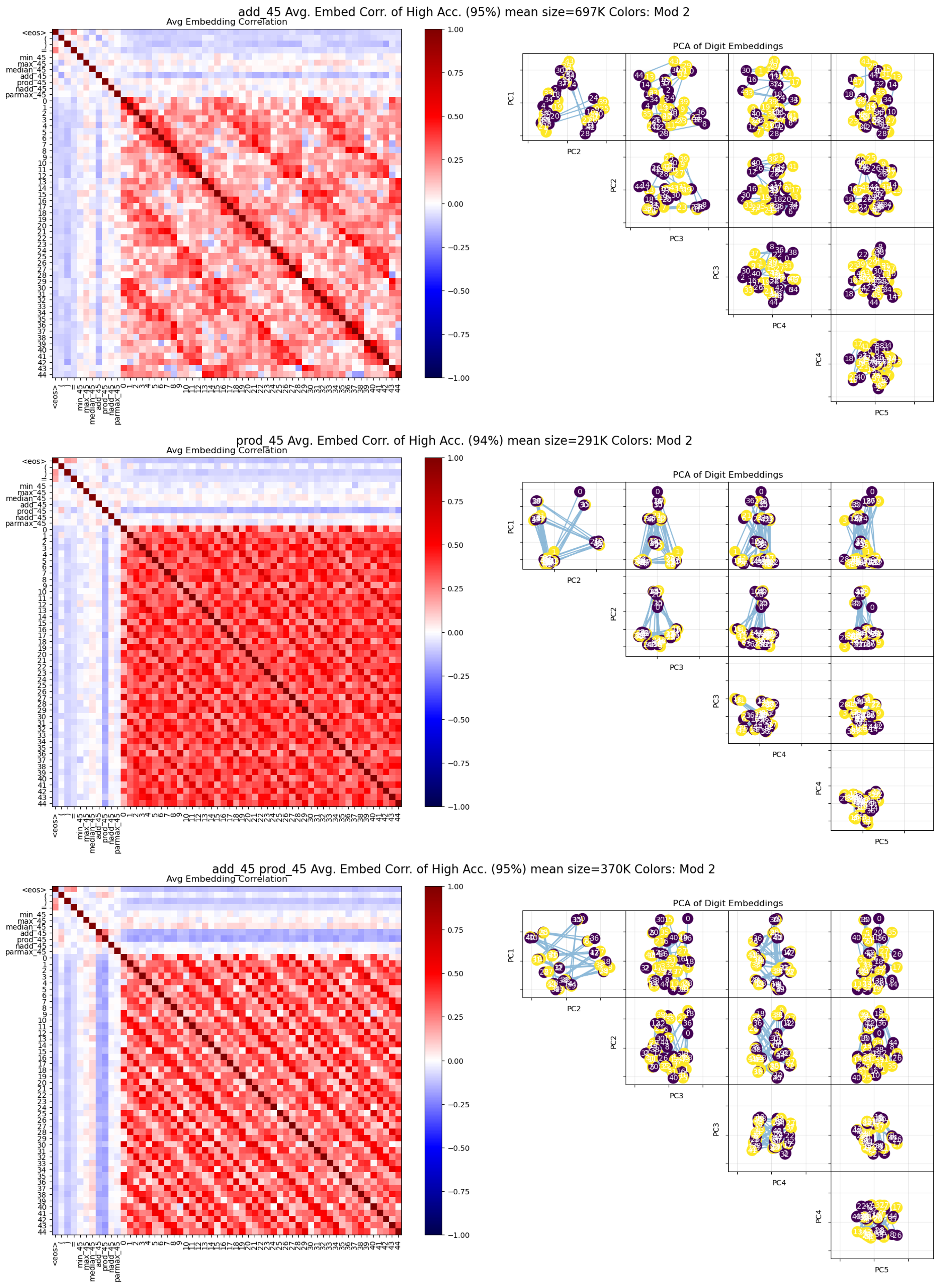}
    \caption{\textbf{ADD, PROD and ADD+PRD in MOD 45 PC embeddings}} 
    \label{fig:45_emb}
\end{figure}

\begin{figure}[htbp]
    \centering
    \includegraphics[width=1\linewidth]{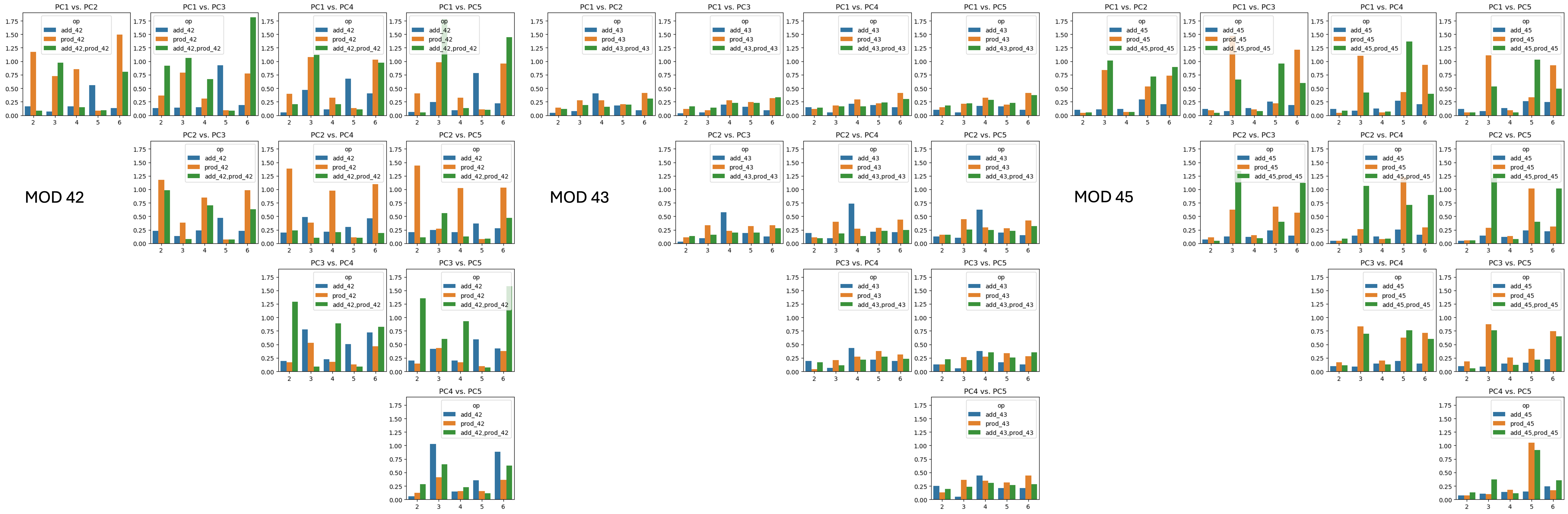}
    \caption{\textbf{Clustering in the PC space.} 
    We measure how well modulo-2, 3, 4, 5, and 6 divisor clusters separate in the PC space for the ADD, PROD, and ADD+PROD models on MOD 42, 43, and 45 (for runs with accuracy $>80\%$). 
    Cluster quality is computed from the mean position and standard deviation of each modulo group, and the pairwise distances between them.
    For MOD 42, strong clustering appears for mod-2, 3, 6, and sometimes 4—precisely the divisors of 42. Strong PROD signals almost always produce strong ADD+PROD signals, whereas strong ADD signals without PROD support do not transfer. 
    For MOD 43, no model shows strong clustering; both ADD and PROD struggle and do not help each other.
    For MOD 45, strong clustering occurs for mod-3 and mod-5, and occasionally mod-6 (via divisor 3). 
    We also observe that the mixed model only exhibits strong clustering when PROD does as well, indicating that the PROD embedding drives the shared structure and helps ADD.
    } 
    \label{fig:grouping_mod}
\end{figure}

\begin{figure}[htbp]
    \centering
    \includegraphics[width=1\linewidth]{figs/2025-09/average_clustering.png}
    \caption{\textbf{Clustering in the PC space by averaging over the first 5 PCs.} 
    We measure how well modulo-2, 3, 4, 5, and 6 divisor clusters separate in the PC space for the ADD, PROD, and ADD+PROD models on MOD 42, 43, and 45 (for runs with accuracy $>80\%$). 
    Cluster quality is computed from the mean position and standard deviation of each modulo group, and the pairwise distances between them, averaged over the first five PCs. 
    For MOD 42, strong clustering appears for mod-2, 3, and 4 (could be an artifact of 2) — precisely the divisors of 42. For MOD 43, we find no strong clustering compared to MOD 42 and MOD 45. For MOD 45, strong clustering occurs for mod-3 and mod-5, as expected from the divisors of 45. We also observe that the mixed model exhibits strong clustering only when PROD does as well, indicating that the PROD embedding drives the shared structure and supports ADD.
    } 
    \label{fig:avg_groping}
\end{figure}

\newpage
\newpage

~\newpage







\end{document}